\documentclass[10pt,twocolumn,letterpaper]{article}

\usepackage{iccv}
\usepackage{times}
\usepackage{epsfig}
\usepackage{graphicx}
\usepackage{amsmath}
\usepackage{amssymb}
\usepackage{subcaption}
\usepackage{wrapfig}

\usepackage[pagebackref=true,breaklinks=true,letterpaper=true,colorlinks,bookmarks=false]{hyperref}

\iccvfinalcopy 

\def\iccvPaperID{1140} 

\ificcvfinal\fi
\begin{document}

\title{Localization-Aware Active Learning for Object Detection}

\author{
Chieh-Chi Kao\\
University of California, Santa Barbara\\
{\tt\small chiehchi.kao@gmail.com}
\and
Teng-Yok Lee\\
Mitsubishi Electric Research Laboratories\\
{\tt\small tlee@merl.com}
\and
Pradeep Sen\\
University of California, Santa Barbara\\
{\tt\small psen@ece.ucsb.edu}
\and
Ming-Yu Liu\\
Mitsubishi Electric Research Laboratories\\
{\tt\small seanmingyuliu@gmail.com}
}

\maketitle

\begin{abstract}
Active learning---a class of algorithms that iteratively searches for the most informative samples to include in a training dataset---has been shown to be effective at annotating data for image classification. However, the use of active learning for object detection is still largely unexplored as determining informativeness of an object-location hypothesis is more difficult. In this paper, we address this issue and present two metrics for measuring the informativeness of an object hypothesis, which allow us to leverage active learning to reduce the amount of annotated data needed to achieve a target object detection performance. Our first metric measures ``localization tightness'' of an object hypothesis, which is based on the overlapping ratio between the region proposal and the final prediction. Our second metric measures ``localization stability'' of an object hypothesis, which is based on the variation of predicted object locations when input images are corrupted by noise. Our experimental results show that by augmenting a conventional active-learning algorithm designed for classification with the proposed metrics, the amount of labeled training data required can be reduced up to 25\%. Moreover, on PASCAL 2007 and 2012 datasets our localization-stability method has an average relative improvement of 96.5\% and 81.9\% over the baseline method using classification only.

\end{abstract}

\section{Introduction}



Prior works have shown that with a large amount of annotated data, convolutional neural networks (CNNs) can be trained to achieve a super-human performance for various visual recognition tasks.  As tremendous efforts are dedicated into the discovery of effective network architectures and training methods for further advancing the performance, we argue it is also important to investigate into effective approaches for data annotation as data annotation is essential but expensive.


Data annotation is especially expensive for the object-detection task. Compared to annotating image class, which can be done via a multiple-choice question, annotating object location requires a human annotator to specify a bounding box for an object. Simply dragging a tight bounding box to enclose an object can cost 10-times more time than answering a multiple-choice question~\cite{Crowdsourcing_Su2012,HumanVerification_Papadopoulos2016}.
Consequently, a higher pay rate has to be paid to a human labeler for annotating images for an object detection task. In addition to the cost, it is more difficult to monitor and control the annotation quality.


\emph{Active learning}~\cite{settles2010active} is a machine learning procedure that is useful in reducing the amount of annotated data required to achieve a target performance. It has been applied to various computer-vision problems including object classification~\cite{AL_GP_Kapoor2007,AL_EMPC_Freytag2014}, image segmentation~\cite{Konyushkova_2015,Jain_2016}, and activity recognition~\cite{ActivityRecog_Hasan2014,ActivityRecog_Hasan2015}.
Active learning starts by training a baseline model with a small, labeled dataset, and then applying the baseline model to the unlabeled data. For each unlabeled sample, it estimates whether this sample contains critical information that has not been learned by the baseline model. Once the samples that bring the most critical information are identified and labeled by human annotators, they can be added to the initial training dataset to train a new model, which is expected to perform better. Compared to \emph{passive learning}, which randomly selects samples from the unlabeled dataset to be labeled, active learning can achieve the same accuracies with fewer but more informative labeled samples.

Multiple metrics for measuring how informative a sample is have been proposed for the classification task, including maximum uncertainty, expected model change, density weighted, and so on~\cite{settles2010active}. The concept behind several of them is to evaluate how uncertain the current model is for an unlabeled sample. If the model could not assign a high probability to a class for a sample, then it implies the model is uncertain about the class of the sample. In other words, the class of the sample would be very informative to the model. This sample would require human to clarify.

Since an object-detection problem can be considered as an object-classification problem once the object is located, 
existing active learning approaches for object detection ~\cite{ALsatellite_Bietti2012, AL_detection_Sivaraman2014} mainly measure the information in the classification part. 
Nevertheless, in addition to classification, the accuracy of an object detector also relies on its localization ability. Because of the importance of localization, in this paper we present an active learning algorithm tailored for object detection, which considers the localization of detected objects. Given a baseline object detector which detects bounding boxes of objects, our algorithm evaluates the uncertainty of both the classification and localization.

Our algorithm is based on two quantitative metrics of the localization uncertainty.

\begin{enumerate}
  \item \emph{Localization Tightness (LT)}: 
The first metric is based on how tight the detected bounding boxes can enclose true objects. The tighter the bounding box, the more certain the localization. While it sounds impossible to compute the localization tightness for non-annotated images because the true object locations are unknown, for object detectors that follow the propose-then-classify pipeline~\cite{Fast_RCNN_Girshick15,Faster_RCNN_Ren2015}, we estimate the localization tightness of a bounding box based on its changes from the intermediate proposal (a box contains any kind of foreground objects) to the final class-specific bounding box.

  \item \emph{Localization Stability (LS)}: 
The second metric is based on whether the detected bounding boxes are sensitive to changes in the input image. To evaluate the localization stability, our algorithm adds different amounts of Gaussian noise to pixel values of the image, and measures how the detected regions vary with respect to the noise. This one can be applied to all kinds of object detectors, especially those that do not have an explicit proposal stage~\cite{YOLO_Redmon2016,SSD_Liu2016}.

\end{enumerate}



The contributions of this paper are two-fold:

\begin{enumerate}

\item 
We present different metrics to quantitatively evaluate the localization uncertainty of an object detector. Our metrics consider different aspects of object detection in spite that the ground truth of object locations is unknown, making our metrics suited for active learning.

\item 
We demonstrate that to apply active learning for object detection, both the localization and the classification of a detector should be considered when sampling informative images. 
Our experiments on benchmark datasets show that considering both the localization and classification uncertainty outperforms the existing active-learning algorithm works on the classification only and passive learning.


\end{enumerate}

\section{Related Works}

We now review active learning approaches used for image classification. For more detail of active learning, Settles's survey~\cite{settles2010active} provides a comprehensive review. In this paper, we use the maximum uncertainty method in the classification as the baseline method for comparison. The uncertainty based method is used for CAPTCHA recognition~\cite{CAPTCHA_Stark2015}, image classification~\cite{BALD_Islam2016}, and automated and manual video annotation~\cite{Karasev_2014}.
It also has been applied to different learning models including decision trees~\cite{HeterogeneousUncertainty_Lewis94}, SVMs~\cite{UncertaintySVM_Tong2002}, and Gaussian processes ~\cite{GP_Kapoor_2010}. 
We choose uncertainty-based method since it is efficient to compute.


Active learning is also applied for object detection tasks in various specific applications, such as satellite images~\cite{ALsatellite_Bietti2012} and vehicle images~\cite{AL_detection_Sivaraman2014}.
Vijayanarasimhan \textit{et al.}~\cite{AL_crowds_Vijayanarasimhan2014} propose an approach to actively crawl images from the web to train part-based linear SVM detector. Note that these methods only consider information from the classifier, while our methods aim to consider the localization part as well.

Current state-of-the-art object detectors are based on deep-learning. They can be classified into two categories. Given an input image, the first category explicitly generates region proposals, following by feature extraction, category classification, and fine-tuning of the proposal geometry ~\cite{Fast_RCNN_Girshick15,Faster_RCNN_Ren2015}.
The other category directly outputs the object location and class without the intermediate proposal stage, such as YOLO~\cite{YOLO_Redmon2016} and SSD~\cite{SSD_Liu2016}.
This inspires us to consider localization stability, which can be applied to both categories.

Besides active learning, there are other research directions to reduce the cost for annotation.
Temporal coherence of the video frames are used to reduce the annotation effort for training detectors~\cite{TemporalConsistency_Prest2012}.
Domain adaptation~\cite{LSDA_Hoffman2014} is used to transfer the knowledge from an image classifier to an object detector without the annotation of bounding boxes.
Papadopoulos \textit{et al.}~\cite{HumanVerification_Papadopoulos2016} suggest to simplify the annotation process from drawing a bounding box to simply answering a Yes/No question whether a bounding box tightly encloses an object.
Russakovsky \textit{et al.}~\cite{Russakovsky_2015} integrate multiple inputs from both computer vision and humans to label objects.

\begin{figure}[t]
\begin{center}
   \includegraphics[width=0.9\linewidth]{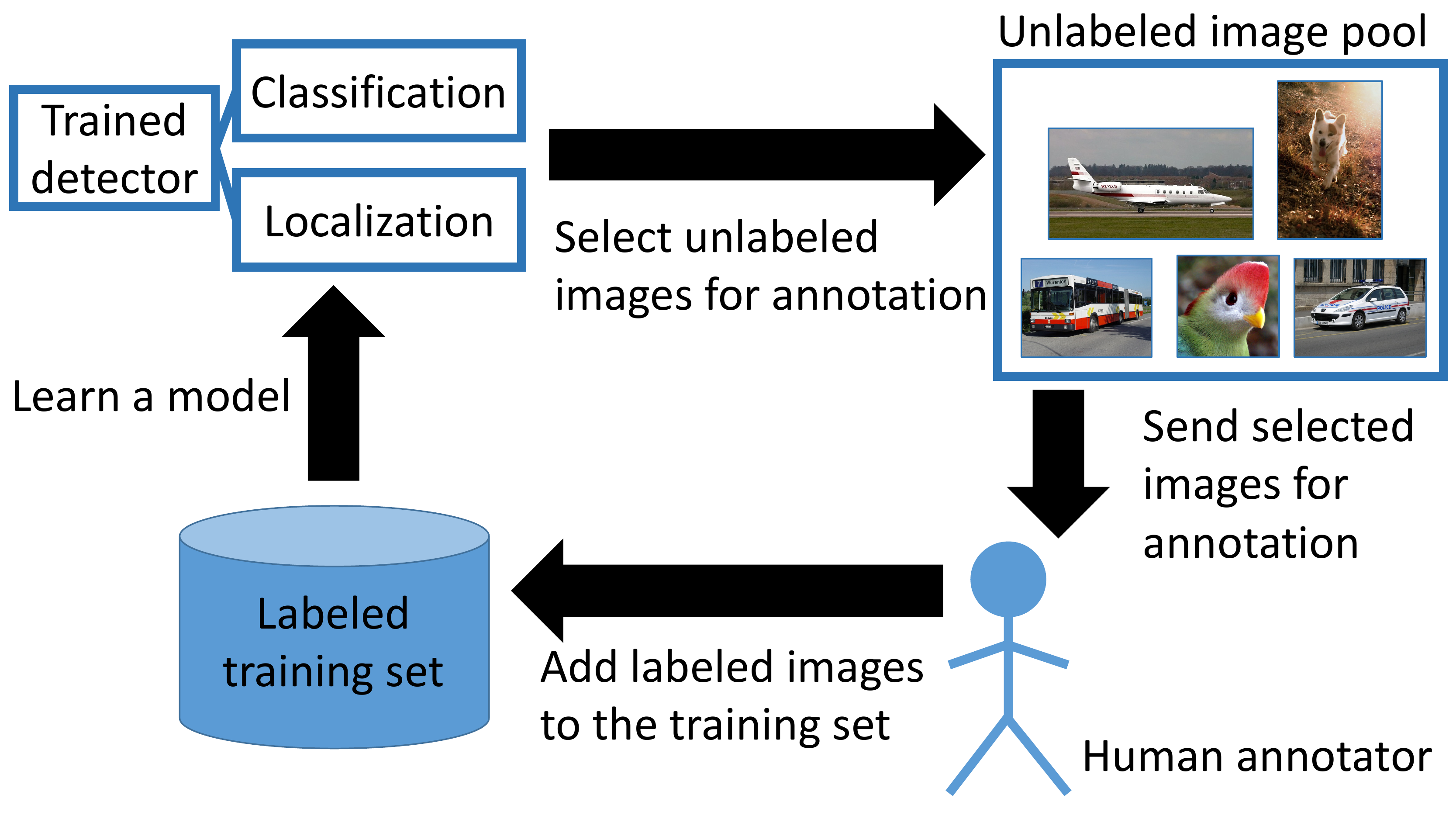} 
\end{center}
   \caption{A round of active learning for object detection.}
\label{fig:AL}
\end{figure}

\section{Active Learning for Object Detection}

The goal of our algorithm is to train an object detector that takes an image as input and outputs a set of rectangular bounding boxes. Each bounding box has the location and the scale of its shape, and a probability mass function of all classes. To train such an object detector, the training and validation images of the detector are annotated with an bounding box per object and its category. Such an annotation is commonly seen in public datasets including PASCAL VOC \cite{PASCAL} and MS COCO \cite{COCO}.

We first review the basic active learning framework for object detection in Sec. \ref{subsec:uncertainty}. It also reviews the measurement of classification uncertainty, which is the major measurement for object detection in previous active learning algorithms for object detection \cite{settles2010active, ALsatellite_Bietti2012, AL_detection_Sivaraman2014}. Based on this framework, we extend the uncertainty measurement to also consider the localization result of a detector, as described in Sec. \ref{sec:AL_tightness} and \ref{sec:AL_stability}. 



\subsection{Active Learning with Classification Uncertainty}
\label{subsec:uncertainty}

Fig.~\ref{fig:AL} overviews our active learning algorithm. Our algorithm starts with a small training set of annotated images to train a baseline object detector. In order to improve the detector by training with more images, we continue to collect images to annotate. Other than annotating all newly collected images, based on different characteristics of the current detector, we select a subset of them for human annotators to label. Once being annotated, these selected images are added to the training set to train a new detector. The entire process continues to collect more images, select a subset with respect to the new detector, annotate the selected ones with humans, re-train the detector and so on. Hereafter we call such a cycle of data collection, selection, annotation, and training as a \emph{round}.

A key component of active learning is the selection of images. 
Our selection is based on the uncertainty of both the classification and localization. The classification uncertainty of a bounding box is the same as the existing active learning approaches \cite{settles2010active, ALsatellite_Bietti2012, AL_detection_Sivaraman2014}. Given a bounding box $B$, its classification uncertainty $U_B(B)$ is defined as $U_B(B) = 1 - P_{max}(B)$ where $P_{max}(B)$ is highest probability out of all classes for this box. If the probability on a single class is close to 1.0, meaning that the probabilities for other classes are low, the detector is highly certain about its class. To the contrast, when multiple classes have similar probabilities, each probability will be low because the sum of probabilities of all classes must be one.

Based on the classification uncertainty per box, given the $i$-th image to evaluate, say $I_i$, its classification uncertainty is denoted as $U_C(I_i)$, which is calculated by the maximum uncertainty out of all detected boxes within.




\begin{figure}[t]
\begin{center}
   \includegraphics[width=0.9\linewidth]{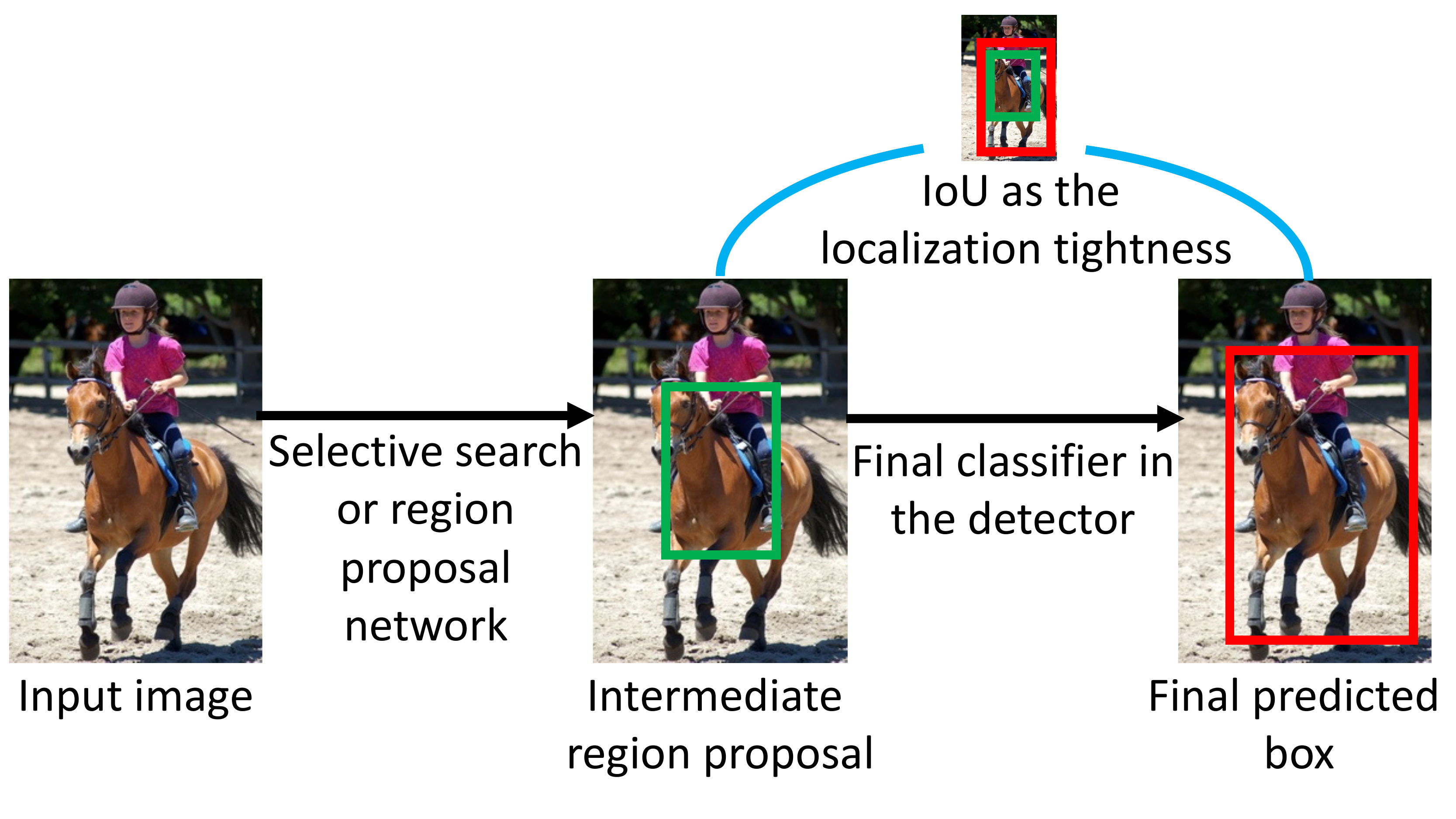}
\end{center}
   \caption{The process of calculating the tightness of each predicted box. Given an intermediate region proposal, the detector refines it to a final predicted box. The IoU calculated by the final predicted box and its corresponding region proposal is defined as the localization tightness of that box.}
\label{fig:tightness_fig}
\end{figure}

\subsection{Localization Tightness}
\label{sec:AL_tightness}

Our first metric of the localization uncertainty is based on the \emph{Localization Tightness} (LT) of a bounding box. The localization tightness measures how tight a predicted bounding box can enclose true foreground objects. Ideally, if the ground-truth locations of the foreground objects are known, the tightness can be simply computed as the IoU (Intersection over Union) between the predicted bounding box and the ground truth. Given two boxes $B^{1}$ and $B^{2}$, their IoU is defined as: $IoU(B^{1}, B^{2})=\frac{B^{1}\cap{B^{2}}}{B^{1}\cup{B^{2}}}$.

Because the ground truth is unknown for an image without annotation, an estimate for the localization tightness is needed. Here we design an estimate for object detectors that involves the adjustment from intermediate region proposals to the final bounding boxes. Region proposals are the bounding boxes that might contain any foreground objects, which can be obtained via the selective search~\cite{SelectiveSearch_Uijlings2013} or a region proposal network~\cite{Faster_RCNN_Ren2015}. Besides classifying the region proposals into specific classes, the final stage of these object detectors can even adjust the location and scale of region proposals based on the classified object classes. Fig.~\ref{fig:tightness_fig} illustrates the typical pipeline of these detectors where the region proposal (green) in the middle is adjusted to the red box in the right.

As the region proposal is trained to predict the location of foreground objects, the refinement process in the final stage is actually related to how well the region proposal predicts. If the region proposal locates the foreground object perfectly, there is no need to refine it. Based on this observation, we use the IoU value between the region proposal and the refined bounding box to estimate the localization tightness between an adjusted bounding box and the unknown ground truth. The estimated tightness $T$ of $j$-th predicted box $B_{0}^{j}$ can be formulated as following: $T(B_{0}^j)= IoU(B_{0}^{j}, R_{0}^j)$, where $R_{0}^j$ is the corresponding region proposal fed into the final classifier that generates $B_{0}^{j}$.

\begin{figure}[tb!]
\begin{tabular}{cc}
\includegraphics[width=0.44\linewidth]{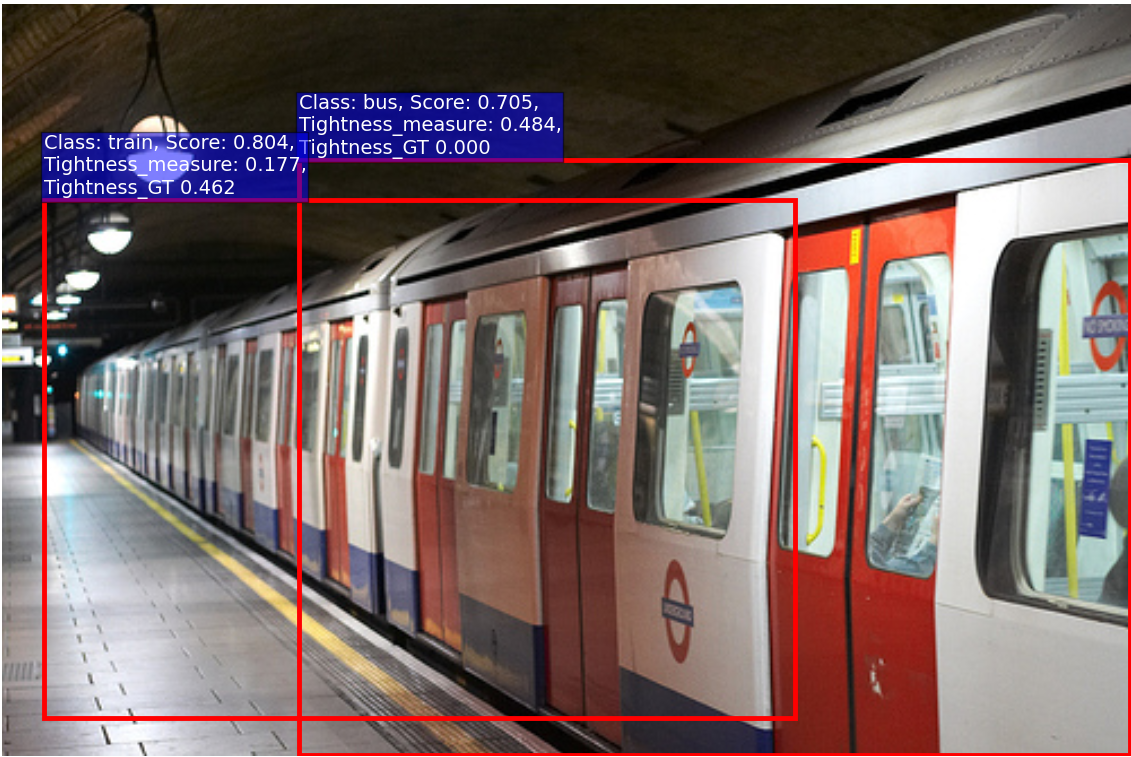} &
\includegraphics[width=0.44\linewidth]{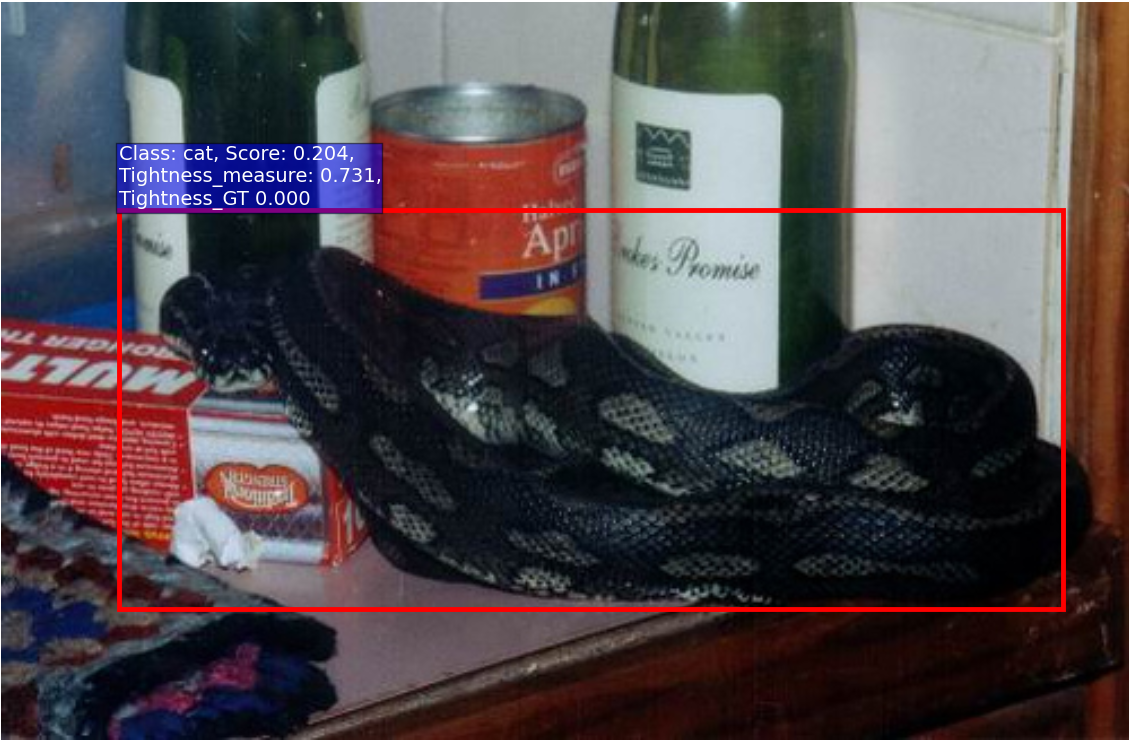} \\
(a) & (b)
\end{tabular}
\caption{Images preferred by $LT/C$. Top rows show two figures are two cases that will be selected by $LT/C$, which are images with certain category but loose bounding box (a) or images with tight bounding box but uncertain about the category (b).
}
\label{fig:LT_and_C}
\end{figure}

Once the tightness of all predicted boxes are estimated, we can extend the selection process to consider not only the classification uncertainty but also the tightness. Namely, we want to select images with inconsistency between the classification and the localization, as following:

\begin{itemize}
  \item Given a predicted box that is absolutely certain about its classification result ($P_{max}=1$), but it cannot tightly enclose a true object ($T = 0$). An example is shown in Figure \ref{fig:LT_and_C} (a).

  \item Reversely, if the predicted box can tightly enclose a true object ($T = 1$) but the classification result is uncertain (low $P_{max}$). An example is shown in Figure \ref{fig:LT_and_C} (b).
\end{itemize}

The score of a box is denoted as $J$, which is computed per Equ. \ref{equ:box_score}. Both conditions above can get value close to zero.


\begin{equation}
\label{equ:box_score}
J(B_{0}^{j}) = |T(B_{0}^{j}) + P_{max}(B_{0}^j) - 1|
\end{equation}




As each image can have multiple predicted boxes, we calculate the score per image as: $T_I(I_{i}) = min_j J(B_{0}^{j})$. Unlabeled images with low score will be selected to annotate in active learning. Since both the localization tightness and classification outputs are used in this metric, later we use $LT/C$ to denotes methods with this score.


\begin{figure}[t]
\begin{center}
   \includegraphics[width=0.9\linewidth]{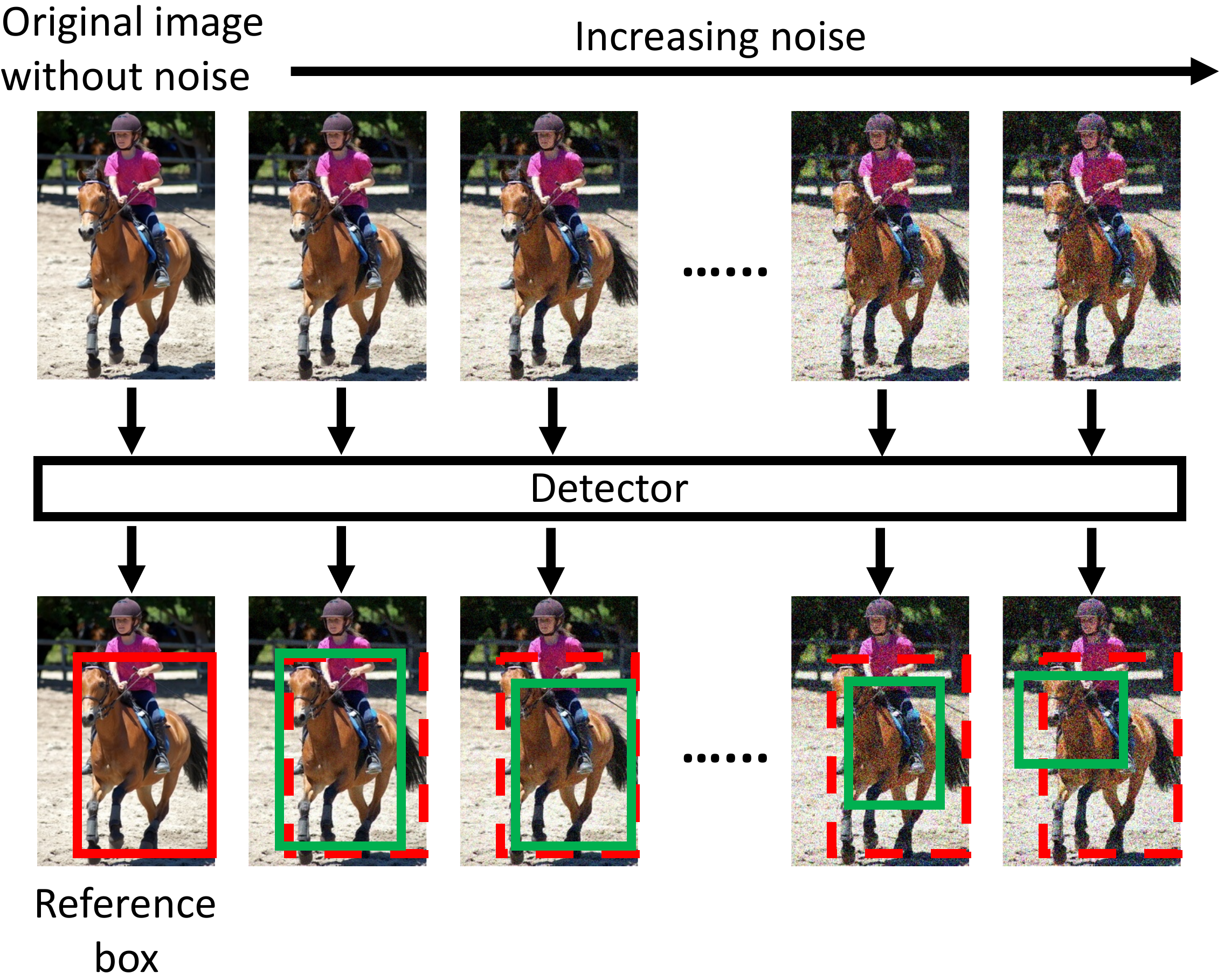}
\end{center}
   \caption{The process of calculating the localization stability of each predicted box. Given one input image, a reference box (red) is predicted by the detector. The change in predicted boxes (green) from noisy images is measured by the IoU of predicted boxes (green) and the corrsponding reference box (dashed red).}
\label{fig:noise}
\end{figure}

\subsection{Localization Stability}
\label{sec:AL_stability}

The concept behind the localization stability is that, if the current model is stable to noise, meaning that the detection result does not dramatically change even if the input unlabeled image is corrupted by noise, the current model already understands this unlabeled image well so there is no need to annotate this unlabeled image. In other words, we would like to select images that have large variation in the localization prediction of bounding boxes when the noise is added into the image.

Fig.~\ref{fig:noise} overviews the idea to calculate the localization stability of an unlabeled image. We first detect bounding boxes in the original image with the current model. These bounding boxes when noise is absent are called reference boxes. The $j$-th reference box is denoted as $B_{0}^j$. For each noise level $n$, a noise is added to each pixel of the image. We use Gaussian noise where the standard deviation is proportional to the level $n$; namely, the pixel value can be changed more for higher level. After detecting boxes in the image with noise level $n$, for each reference box (the red box in Fig.~\ref{fig:noise}), we find a corresponding box (green) in the noisy image to calculate how the reference box varies. The corresponding box is denoted as $C_{n}(B_{0}^j)$, which has the highest IoU value among all bounding boxes that overlap $B_{0}^j$.

Once all the corresponding boxes from different noise levels are detected, we can tell that the model is stable to noise on this reference box if the box does not significantly change across the noise levels. Therefore, the localization stability of each reference box $B_{0}^j$ can be defined as the average of IoU between the reference box and corresponding boxes across all noise levels. Given $N$ noise levels, it is calculated per Equ. \ref{equ:box_stability}:

\begin{equation}
\label{equ:box_stability}
S_B(B_{0}^j)= \frac{\sum_{n=1}^N IoU(B_{0}^{j}, C_{n}(B_{0}^j))}{N},
\end{equation}


With the localization stability of all reference boxes, the localization stability of this unlabeled image, says $I_i$, is defined based on their weighted sum per Equ. \ref{equ:image_localization_stability} where $M$ is the number of reference boxes. The weight of each reference box is its highest class probability in order to prefer boxes with high probability as foreground objects but high uncertainty to their locations.

\begin{equation}
\label{equ:image_localization_stability}
S_I(I_{i})= \frac{\sum_{j=1}^M P_{max}(B_{0}^j)S_B(B_{0}^j)}{\sum_{j=1}^M P_{max}(B_{0}^j)}.
\end{equation}


%
%

\section{Experimental Results}

\textbf{Reference Methods:}
Since no prior work does active learning for deep learning based object detectors, we designate two informative baselines that show the impact of proposed methods.

\begin{itemize}
\item \textbf{Random (R):} Randomly choose samples from the unlabeled set, label them, and put them into labeled training set. 

\item \textbf{Classification only (C):} Select images only based on the classification uncertainty $U_c$ in Sec. \ref{subsec:uncertainty}.

\end{itemize}

Our algorithm with two different metrics for the localization uncertainty are tested. First, the localization stability (Section~\ref{sec:AL_stability}) is combined with the classification information (\textbf{LS+C}). 
As images with high classification uncertainty and low localization stability should be selected for annotation, the score of the $i$-th image ($I_i$) image is defined as follows:
$U_C(I_i) - \lambda S_I(I_i)$
,where $\lambda$ is the weight to combine both, which is set to 1 across all the experiments in this paper.
Second, the localization tightness of predicted boxes is combined with the classification information (\textbf{LT/C}) as defined in Section~\ref{sec:AL_tightness}.

We also test three variants of our algorithm.
One uses the localization stability only (\textbf{LS}).
Another is the localization tightness of predicted boxes combined with the classification information but using the localization tightness calculated from ground-truth boxes (\textbf{LT/C(GT)}) instead of the estimate used in LT/C.
The other is combining all 3 cues together (\textbf{3in1}). 

For the easiness of reading, data for LS and 3in1 are shown in the supplementary result. Our supplementary result also includes the mAP curves with error bars that indicate the minimum and maximum average precision (AP) out of multiple trials of all methods. Furthermore, experiments with different designs of LT/C are included in the supplementary result.

%

\textbf{Datasets:}
We validated our algorithm on three datasets (PASCAL 2012, PASCAL 2007, MS COCO \cite{PASCAL,COCO}). For each dataset, we started from a small subset of the training set to train the baseline model, and selected from the remained training images for active learning. Since objects in training images from these datasets have been annotated with bounding boxes, our experiments used these bounding boxes as annotation without asking human annotators.

\textbf{Detectors:}
The object detector for all datasets is the Faster-RCNN (FRCNN) ~\cite{Faster_RCNN_Ren2015}, which contains the intermediate stage to generate region proposals. We also tested our algorithm with the Single Shot multibox Detector (SSD)~\cite{SSD_Liu2016} on the PASCAL 2007 dataset. Because the SSD does not contain a region proposal stage, the tests for localization tightness were skipped. Both FRCNN and SSD used VGG16~\cite{VGG} as the pre-trained network in the experiments shown in this paper.


\subsection{FRCNN on PASCAL 2012}
\label{sec:FRCNN_PASCAL2012}

\textbf{Experimental Setup:}
We evaluate all the methods with the FRCNN model~\cite{Faster_RCNN_Ren2015} using the RoI warping layer~\cite{InstanceSeg_Dai2016} on the PASCAL 2012 object-detection dataset~\cite{PASCAL} that consists of of 20 classes. Its training set (5,717 images) is used to mimic a pool of unlabeled images, and the validation set (5,823 images) is used for testing. Input images are resized to have 600 pixels on the shortest side for all FRCNN models in this paper.

The numbers shown in following sections on PASCAL datasets are averages over 5 trails for each method. All trials start from the same baseline object detectors, which are trained with 500 images selected from the unlabeled image pool. After then, each active learning algorithm is executed in 15 rounds. In each round, we select 200 images, add these images to the existing training set, and train a new model. Each model is trained with 20 epoches.

Our experiments used Gaussian noise as the noise source for the localization stability. We set the number of noise level $N$ to 6. The standard deviations of these levels are \{8, 16, 24, 32, 40, 48\} where the pixels range from [0, 255].

%
%
%

\begin{figure}[t]
    \centering
    \begin{subfigure}[t]{0.24\textwidth}
        \centering
        \includegraphics[width=\linewidth]{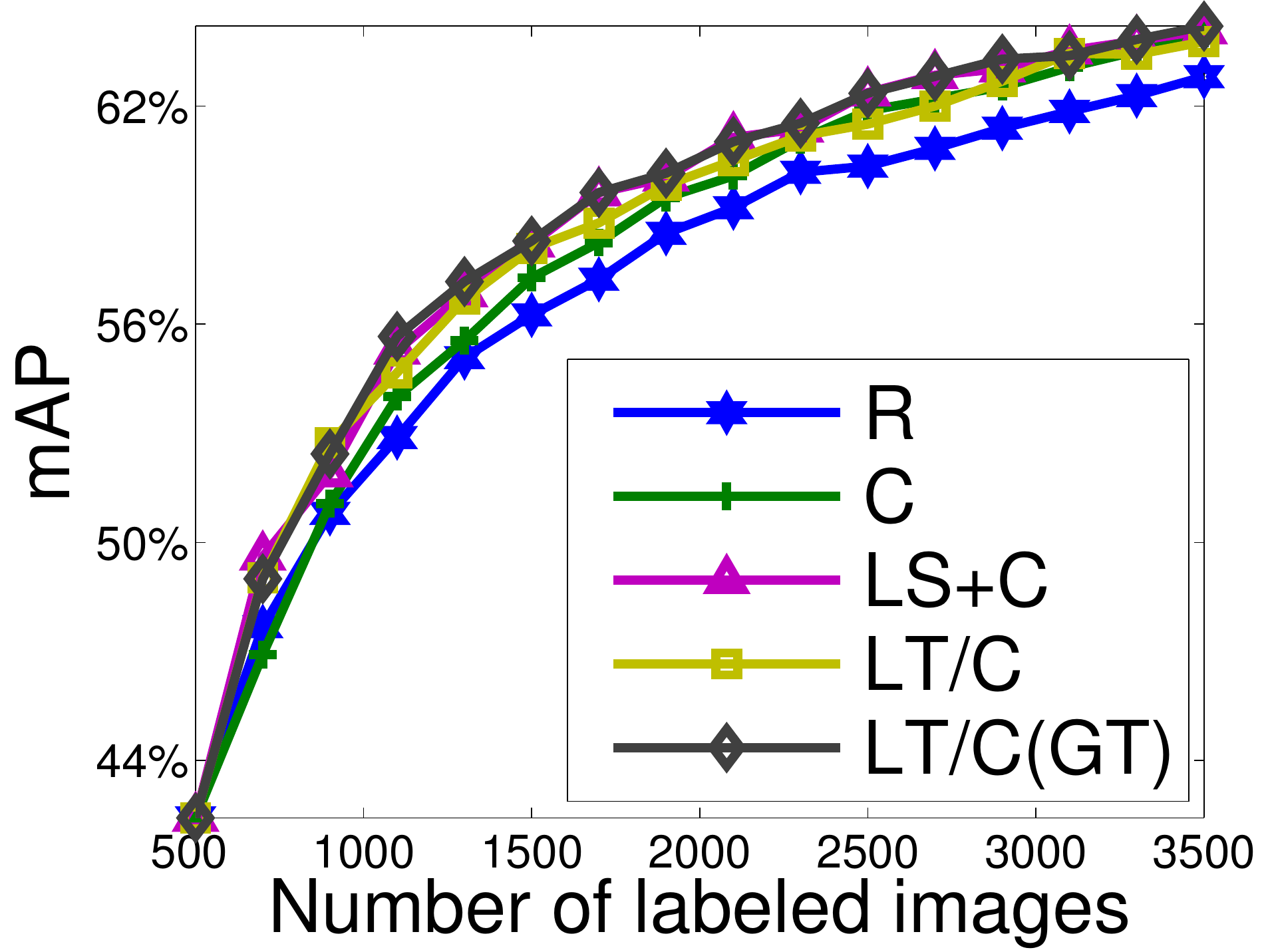}
        \caption{mAP}
        \label{fig:PASCAL2012_mAP}
    \end{subfigure}%
    ~
    \begin{subfigure}[t]{0.24\textwidth}
        \centering
        \includegraphics[width=\linewidth]{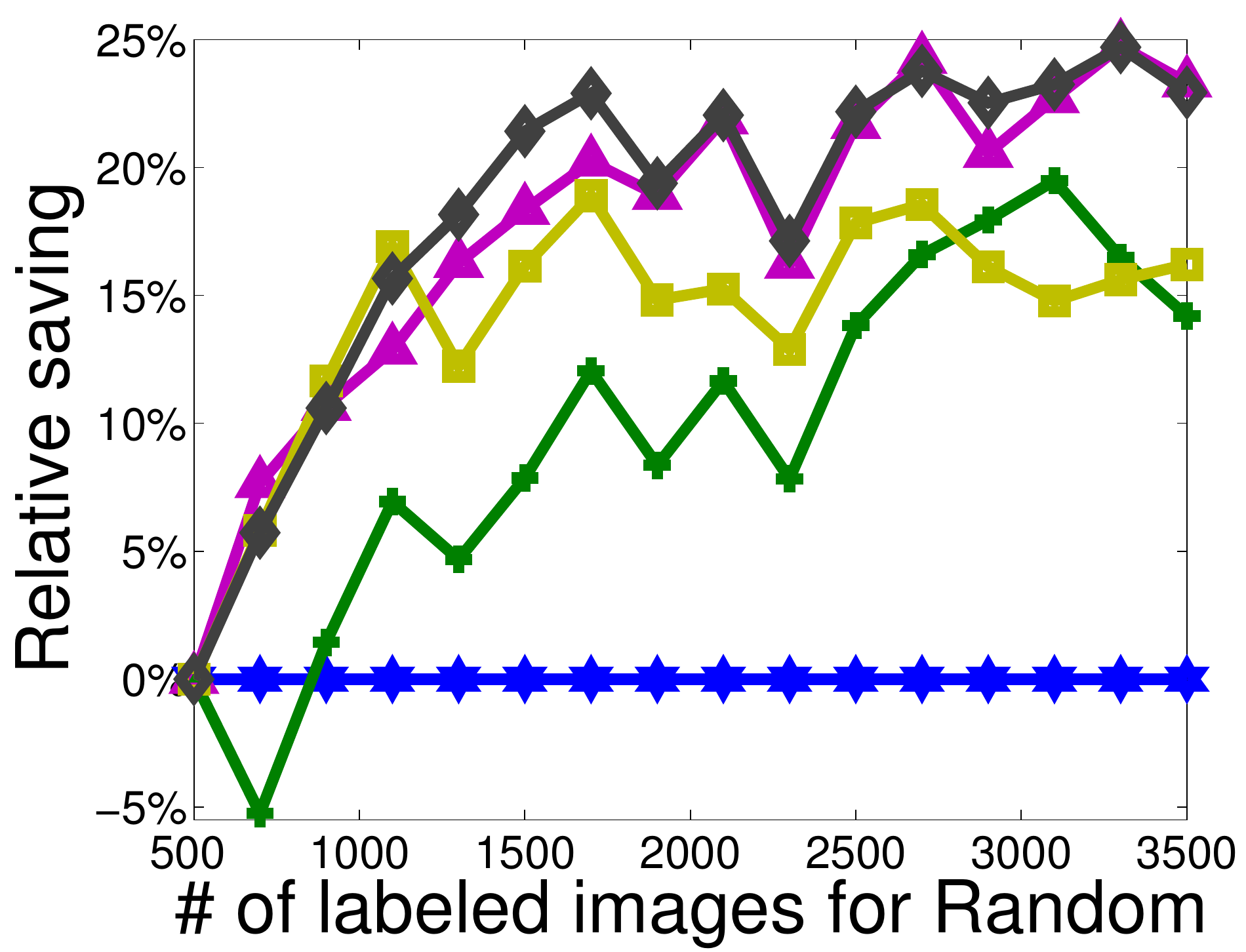}
        \caption{Saving}
        \label{fig:PASCAL2012_saving}
    \end{subfigure}
    \caption{
(a) Mean average precision curve of different active learning methods on PASCAL 2012 detection dataset. Each point in the plot is an average of 5 trials.
(b) Relative saving of labeled images for different methods.
}
\label{fig:PASCAL2012}
\end{figure}


\textbf{Results:}
Fig.~\ref{fig:PASCAL2012_mAP} and Fig.~\ref{fig:PASCAL2012_saving} show the mAP curve and the relative saving of labeled images, respectively, for different active learning methods. We have three major observations from the results on the PASCAL 2012 dataset. First, LT/C(GT) outperforms all other methods in most of the cases as shown in Fig.~\ref{fig:PASCAL2012_saving}.
This is not surprising since LT/C(GT) is based on the ground-truth annotations.
In the region that achieves the same performance as passive learning with a dataset of 500 to 1,100 labeled images, the performance of the proposed LT/C is similar to LT/C(GT), which represents the full potential of LT/C.
This implies that LT/C using the estimate of tightness of predicted boxes (Section~\ref{sec:AL_tightness}) can achieve results close to its upper bound.


Second, in most of the cases, active learning approaches work better than random sampling.
The localization stability with the classfication uncertainty (LS+C) has the best performance among all methods other than LT/C(GT).
In terms of average saving, LS+C and LT/C have 96.5\% and 36.3\% relative improvement over the baseline method C.


\begin{figure}[t]
    \centering
    \begin{subfigure}[t]{0.24\textwidth} 
        \centering
        \includegraphics[width=\linewidth]{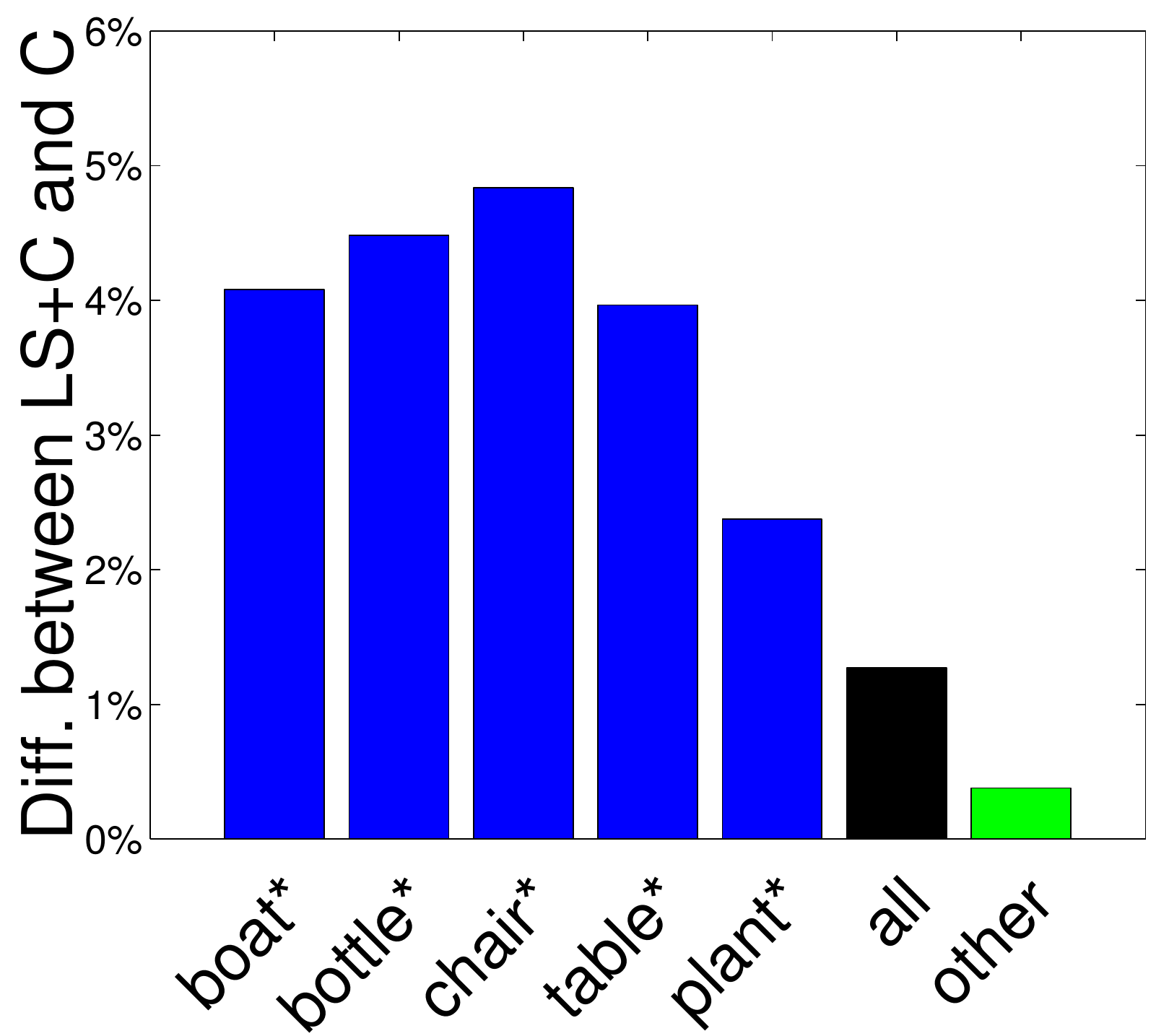}
        \caption{PASCAL 2012}
        \label{fig:PASCAL2012_C_CS_diff}
    \end{subfigure}%
    ~
    \begin{subfigure}[t]{0.24\textwidth} 
        \centering
        \includegraphics[width=\linewidth]{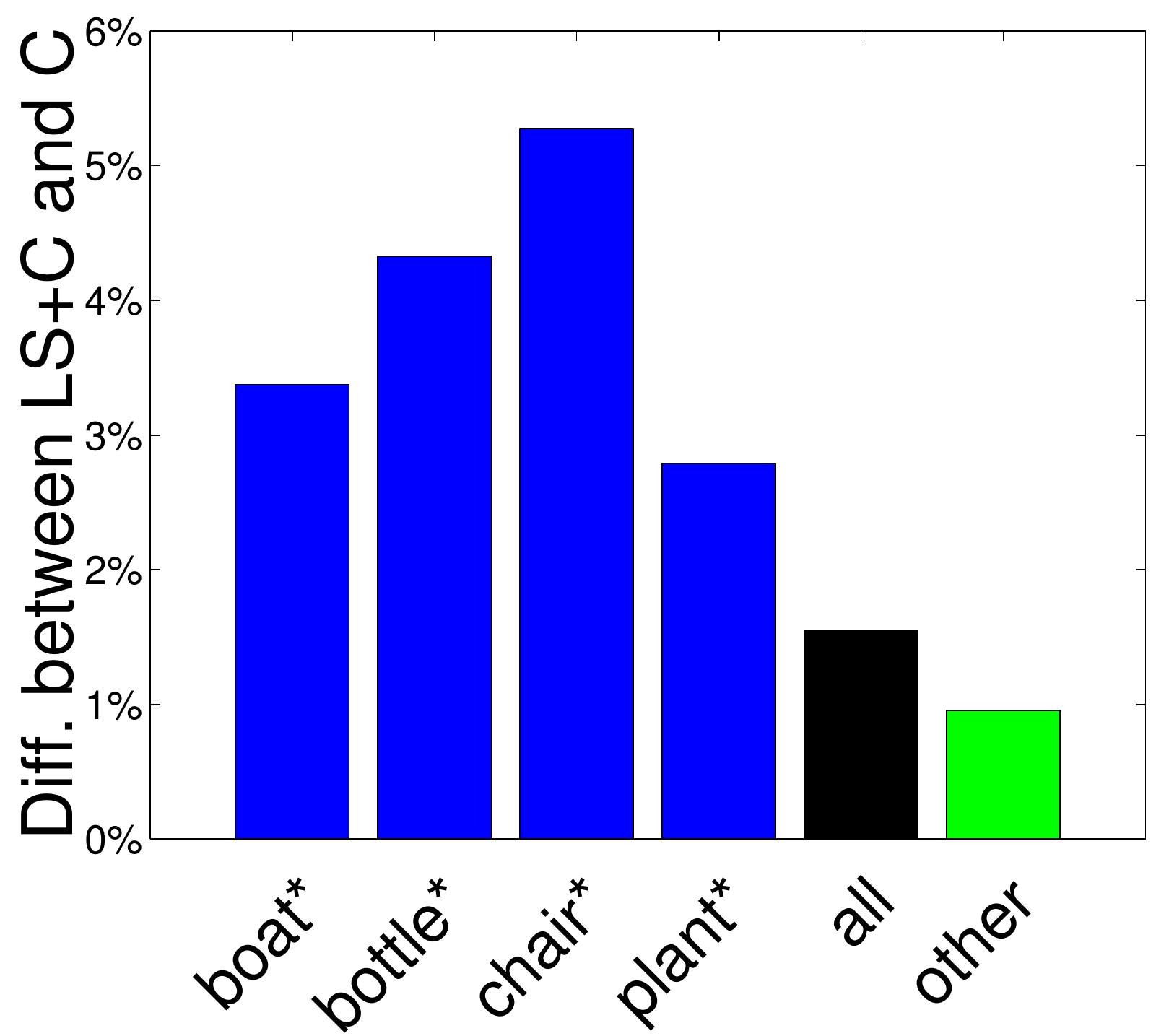}
        \caption{PASCAL 2007}
        \label{fig:PASCAL2007_C_CS_diff}
    \end{subfigure}
    \caption{The difference in difficult classes (blue bars) between the proposed method (LS+C) and the baseline method (C) in average precision on (a) PASCAL 2012 dataset (b) PASCAL 2007 dataset. Black and green bars are the average improvements of LS+C over C for all classes and non-difficult classes.
    }
\label{fig:PASCAL_C_CS_diff}
\end{figure}

Last, we also note that the proposed LS+C method has more improvements in the difficult categories.
We further analyze the performance of each method by inspecting the AP per category.
Table~\ref{tab:PASCAL2012_1100imgs} shows the average precision for each method on the PASCAL 2012 validation set after 3 rounds of active learning, meaning that every model is trained on a dataset with 1,100 labeled images. 
For categories with AP lower than 40\% in passive learning (R), we treat them as difficult categories, which have a asterisk next to their name. For these difficult categories (blue bars) in Fig.~\ref{fig:PASCAL2012_C_CS_diff}, we notice that the improvement of LS+C over C is large.
For those 5 difficult categories the average improvement of LS+C over C is 3.95\%, while the average improvement is only 0.38\% (the green bar in Fig.~\ref{fig:PASCAL2012_C_CS_diff}) for the rest 15 non-difficult categories. This 10$\times$ difference shows that adding the localization information into active learning for object detection can greatly help the learning for difficult categories.
It is also noteworthy that for those 5 difficult categories, the baseline method C performs slightly worse than random sampling by 0.50\% in average.
It indicates that C focuses on non-difficult categories to get an overall improvement in mAP. 

\begin{table*}[htbp]
\center
\begin{scriptsize}
\tabcolsep=0.11cm\begin{tabular*}{0.93\linewidth}{l|*{20}{c}|c}
method &aero&bike&bird&boat*&bottle*&bus&car&cat&chair*&cow&table*&dog&horse&mbike&persn&plant*&sheep&sofa&train&tv&mAP\\
\hline

R&\underline{71.1} &61.5 &\underline{54.7} &28.4 &32.0 &\underline{68.1} &57.9 &75.4 &25.8 &44.2 &36.4 &73.0 &61.9 &67.3 &68.1 &21.6 &51.9 &41.0 &\textbf{65.5} &51.7 &52.9 \\
C&70.7 &62.9 &54.7 &25.5 &30.8 &66.1 &56.2 &\textbf{78.1} &26.4 &\textbf{54.5} &36.7 &\textbf{76.9} &\textbf{68.3} &\underline{67.7} &67.4 &22.5 &\underline{57.7} &40.8 &63.6 &52.5 &54.0 \\
LS+C&\textbf{73.9} &\underline{63.7} &\textbf{56.9} &\textbf{29.6} &\textbf{35.2} &66.5 &\underline{58.5} &\underline{77.9} &\textbf{31.3} &\underline{50.8} &\underline{40.7} &\underline{73.8} &\underline{65.4} &66.9 &\underline{68.4} &\textbf{24.8} &\textbf{58.0} &\textbf{44.9} &64.2 &\underline{53.9} &\textbf{55.3} \\
LT/C&69.8 &\textbf{64.6} &54.6 &\underline{29.5} &\underline{33.8} &\textbf{70.3} &\textbf{59.7} &75.5 &\underline{29.5} &46.3 &\textbf{41.8} &73.0 &62.5 &\textbf{69.0} &\textbf{70.8} &\underline{23.2} &56.5 &\underline{42.8} &\underline{64.3} &\textbf{55.9} &\underline{54.7} \\

\end{tabular*}
\end{scriptsize}
\caption{Average precision for each method on PASCAL 2012 validation set after 3 rounds of active learning (number of labeled images in the training set is 1,100). Each number shown in the table is an average of 5 trials and displayed in percentage. Numbers in bold are the best results per column, and underlined numbers are the second best results. Catergories with AP lower than 40\% in passive learning (R) are defined as difficult categories and marked by asterisk.}\label{tab:PASCAL2012_1100imgs}
\end{table*}

\subsection{FRCNN on PASCAL 2007}
\label{sec:FRCNN_PASCAL2007}
\textbf{Experimental Setup:}
We evaluate all the methods with the FRCNN model~\cite{Faster_RCNN_Ren2015} using the RoI warping layer~\cite{InstanceSeg_Dai2016} on the PASCAL VOC 2007 object-detection dataset~\cite{PASCAL} that consists of 20 classes.
Both training and validation sets (total 5,011 images) are used as the unlabeled image pool, and the test set (4,952 images) is used for testing.
All the experimental settings are the same as the experiments on the PASCAL 2012 dataset as mentioned Section~\ref{sec:FRCNN_PASCAL2012}.

%
%

\begin{figure}[t]
    \centering
    \begin{subfigure}[t]{0.24\textwidth}
        \centering
        \includegraphics[width=\linewidth]{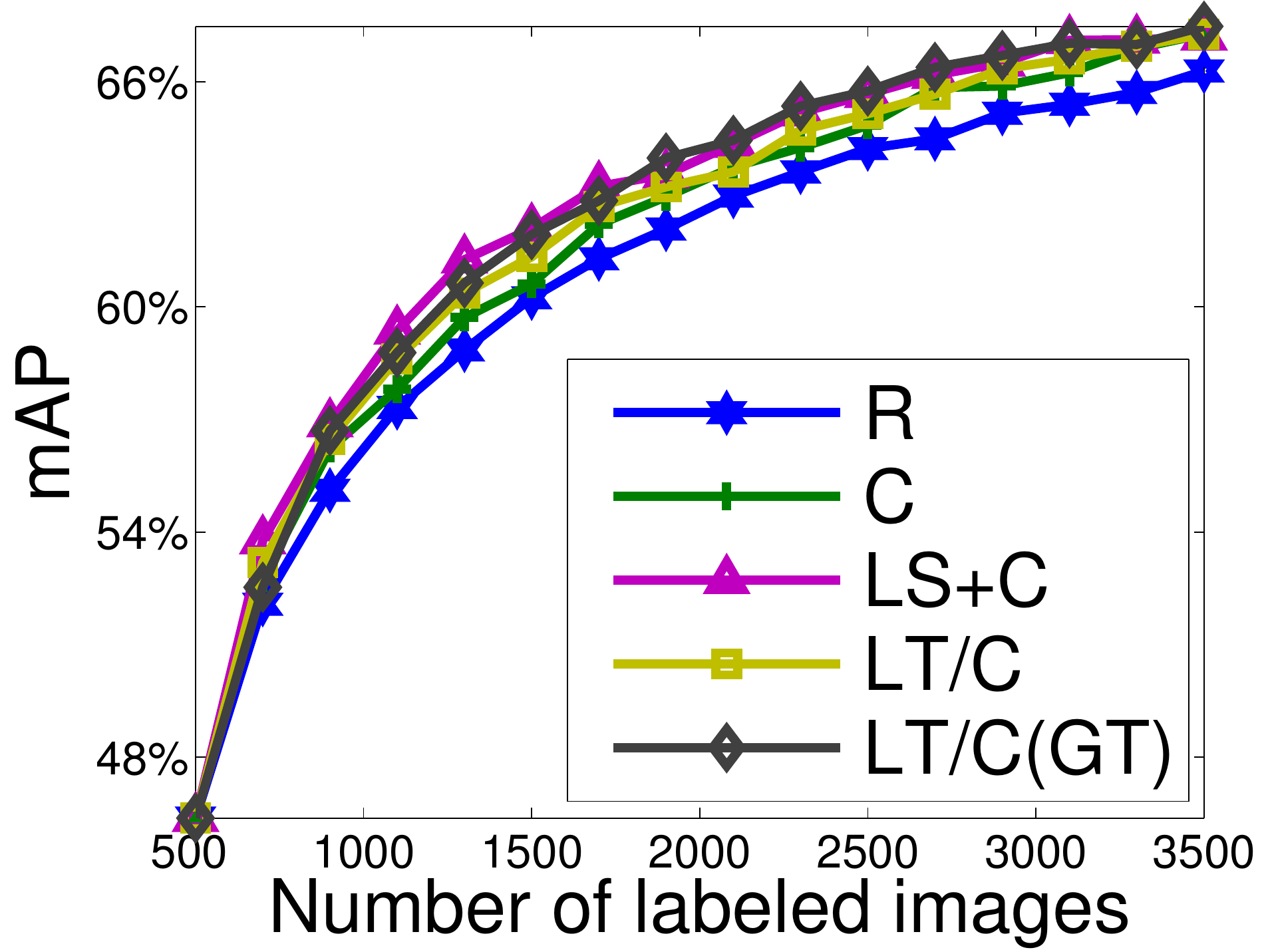}
        \caption{mAP}
        \label{fig:PASCAL2007_mAP}
    \end{subfigure}%
    ~
    \begin{subfigure}[t]{0.24\textwidth}
        \centering
        \includegraphics[width=\linewidth]{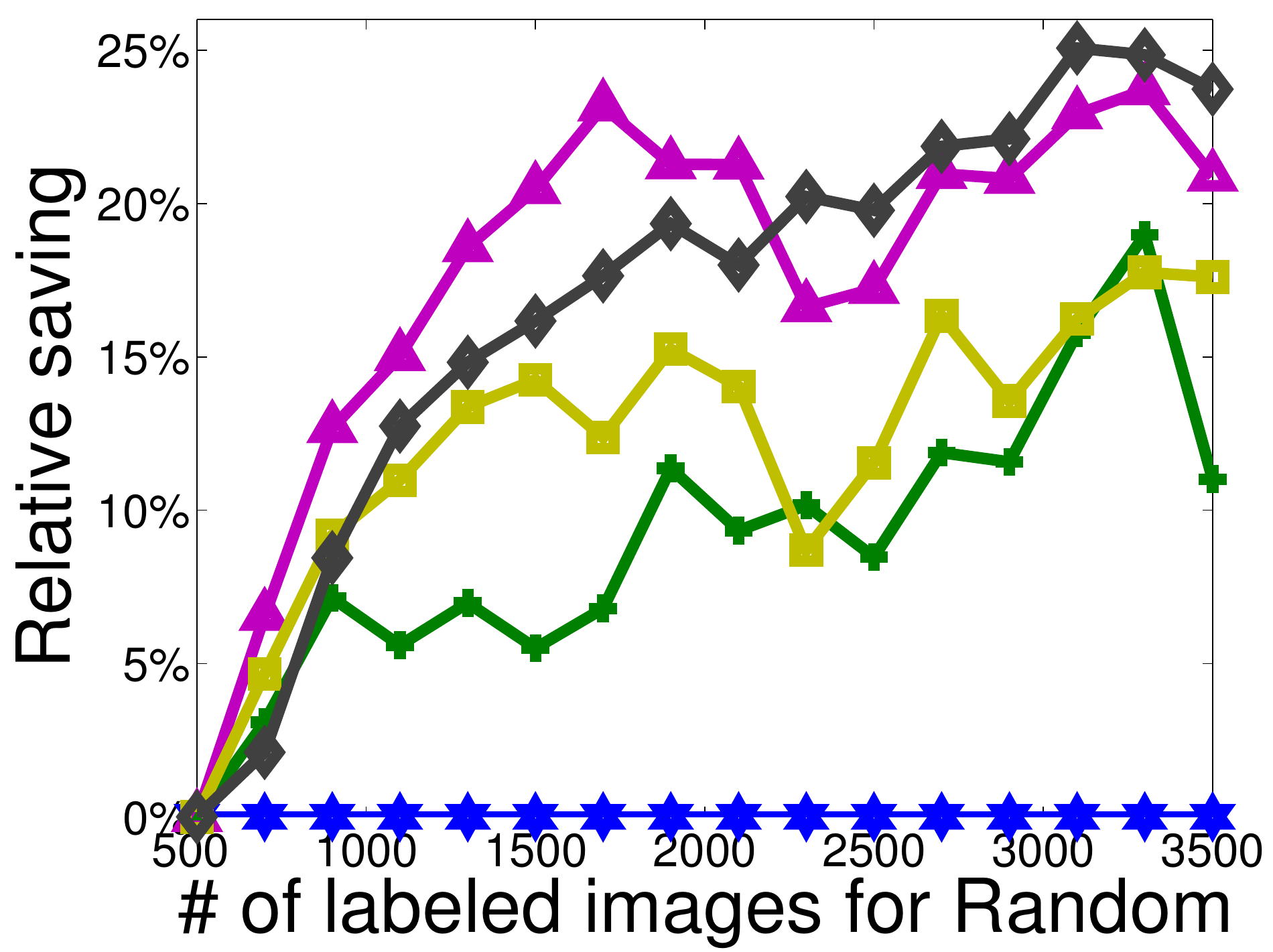}
        \caption{Saving}
        \label{fig:PASCAL2007_saving}
    \end{subfigure}
    \caption{
(a) Mean average precision curve of different active learning methods on PASCAL 2007 detection dataset. Each point in the plot is an average of 5 trials.
(b) Relative saving of labeled images for different methods.
}
\label{fig:PASCAL2007}
\end{figure}



\textbf{Results:}
Fig.~\ref{fig:PASCAL2007_mAP} and Fig.~\ref{fig:PASCAL2007_saving} show the mAP curve and relative saving of labeled images for different active learning methods.
In terms of average saving, LS+C and LT/C have 81.9\% and 45.2\% relative improvement over the baseline method C.
Table~\ref{tab:PASCAL2007_1100imgs} shows the AP for each method on the PASCAL 2007 test set after 3 rounds of active learning.
The proposed LS+C and LT/C are better than the baseline classification-only method (C) in terms of mAP.

It is interesting to see that LS+C method has the same behavior as shown in the experiments on the PASCAL 2012 dataset. Namely, LS+C also outperforms the baseline model C on difficult categories.
As the setting in experiments on the PASCAL 2012 dataset, categories with AP lower than 40\% in passive learning (R) are considered as difficult categories.
For those 4 difficult categories, the average improvement in AP of LS+C over C is 3.94\%, while the average improvement is only 0.95\% (the green bar in Fig.~\ref{fig:PASCAL2007_C_CS_diff}) for the other 16 categories.

\begin{table*}[htbp]
\center
\begin{scriptsize}
\tabcolsep=0.11cm\begin{tabular*}{0.93\linewidth}{l|*{20}{c}|c}
method &aero&bike&bird&boat*&bottle*&bus&car&cat&chair*&cow&table&dog&horse&mbike&persn&plant*&sheep&sofa&train&tv&mAP\\
\hline

R&\textbf{61.6} &67.2 &54.1 &40.0 &33.6 &64.5 &73.0 &73.9 &34.5 &60.8 &52.2 &69.3 &\textbf{74.7} &\underline{66.6} &\underline{67.1} &25.9 &52.1 &54.2 &\textbf{66.1} &54.9 &57.3 \\
C&56.9 &\underline{68.0} &\underline{54.9} &36.8 &34.4 &\underline{68.1} &71.7 &\textbf{75.5} &34.0 &\textbf{68.6} &51.0 &\underline{71.4} &\underline{74.7} &65.2 &65.9 &24.9 &\underline{60.0} &53.9 &63.0 &\underline{57.4} &57.8 \\
LS+C&\underline{61.5} &64.4 &\textbf{55.8} &\underline{40.2} &\textbf{38.7} &66.3 &\underline{73.8} &\underline{74.7} &\textbf{39.6} &\underline{68.0} &\underline{56.3} &\textbf{71.5} &73.8 &\textbf{67.2} &66.7 &\underline{27.7} &\textbf{61.3} &\textbf{57.0} &\underline{65.6} &57.4 &\textbf{59.4} \\
LT/C&57.6 &\textbf{69.7} &52.9 &\textbf{41.1} &\underline{38.4} &\textbf{69.7} &\textbf{74.4} &71.8 &\underline{36.4} &61.2 &\textbf{58.1} &69.5 &74.3 &66.2 &\textbf{67.8} &\textbf{28.0} &55.5 &\underline{56.3} &65.5 &\textbf{58.2} &\underline{58.6} \\
\end{tabular*}
\end{scriptsize}
\caption{Average precision for each method on PASCAL 2007 test set after 3 rounds of active learning (number of labeled images in the training set is 1,100). The other experimental settings are the same as shown in Table~\ref{tab:PASCAL2012_1100imgs}.}
\label{tab:PASCAL2007_1100imgs}
\end{table*}

\begin{figure}[t]
    \centering
    \begin{subfigure}[t]{0.24\textwidth}
        \centering
        \includegraphics[width=\linewidth]{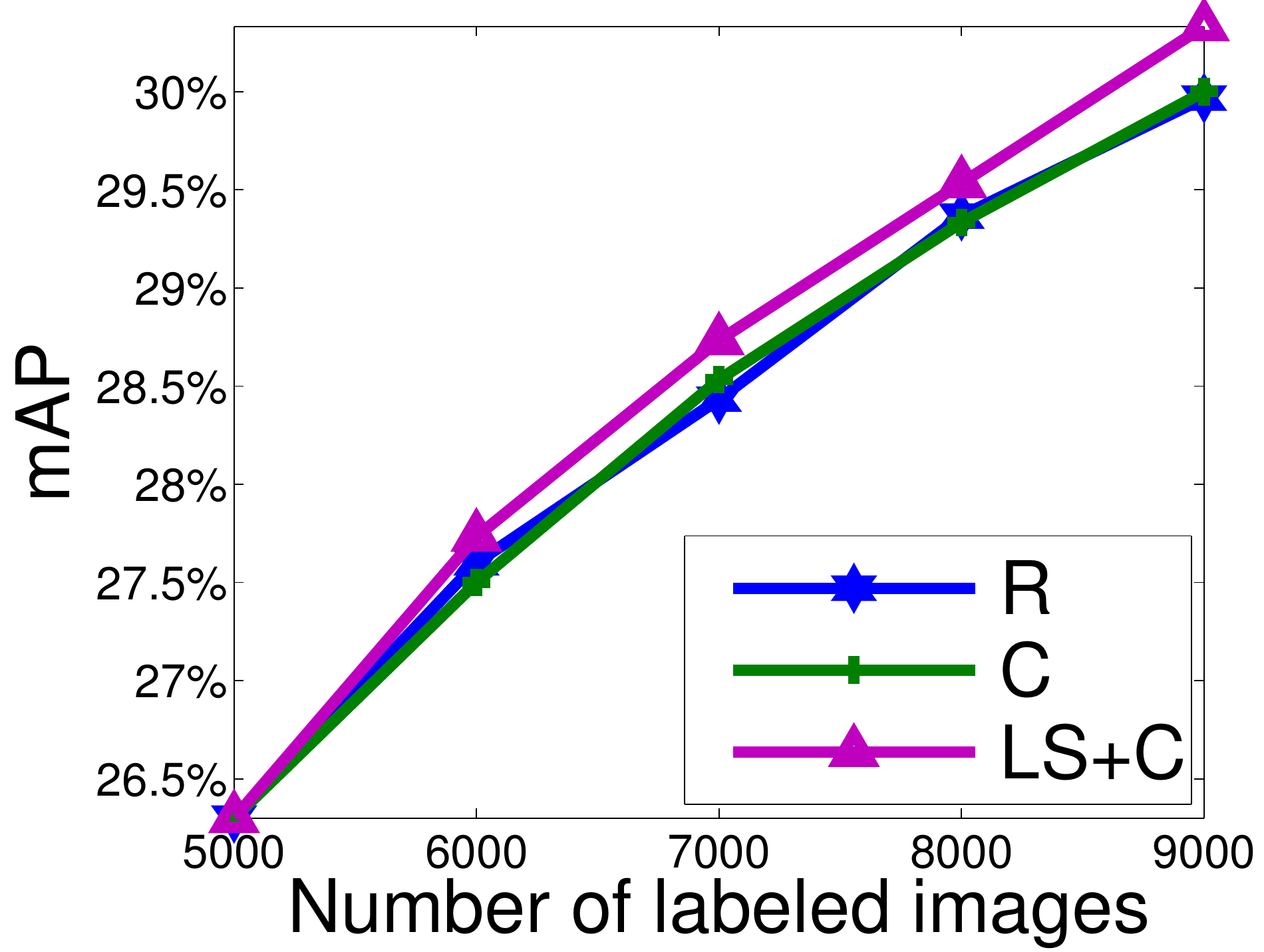}
        \caption{mAP}
        \label{fig:MSCOCO_mAP}
    \end{subfigure}%
    ~
    \begin{subfigure}[t]{0.24\textwidth}
        \centering
        \includegraphics[width=\linewidth]{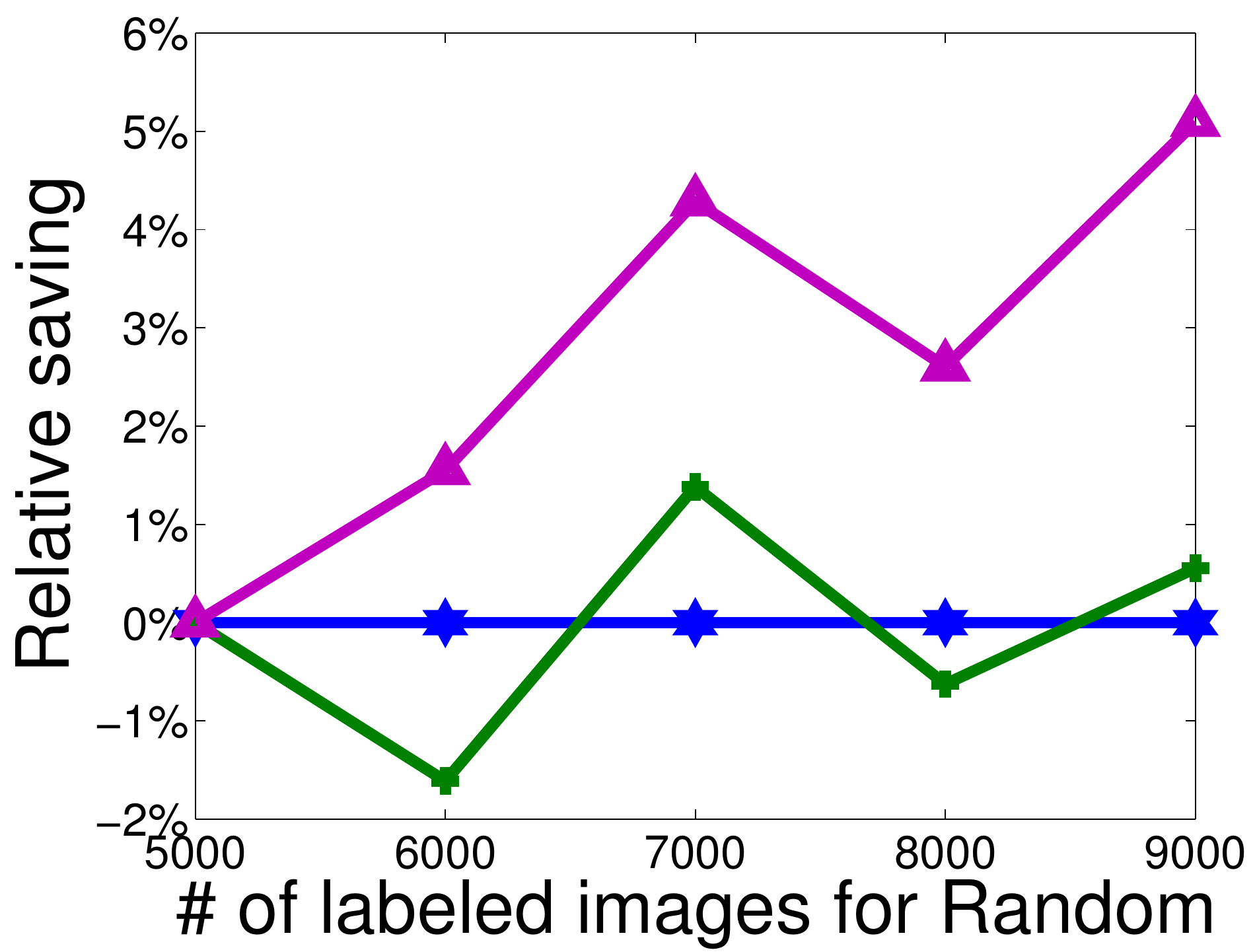}
        \caption{Saving}
        \label{fig:MSCOCO_saving}
    \end{subfigure}
    \caption{(a) Mean average precision curve (@IoU=0.5) of different active learning methods on MS COCO detection dataset. (b) Relative saving of labeled images for different methods. Each point in the plots is an average of 3 trials.  }
\label{fig:MSCOCO}
\end{figure}

\subsection{FRCNN on MS COCO}
\label{sec:FRCNN_MSCOCO}

\textbf{Experimental Setup:}
For the MS COCO object-detection dataset~\cite{COCO}, we evaluate three methods: passive learning (R), the baseline method using classification only (C), and the proposed LS+C. Our experiments still use the FRCNN model~\cite{Faster_RCNN_Ren2015} with the RoI warping layer~\cite{InstanceSeg_Dai2016}.
Compared to the PASCAL datasets, the MS COCO has more categories (80) and more images (80k for training and 40k for validation).
Our experiments use the training set as the unlabeled image pool, and the validation set for testing.

The numbers shown in this section are averages over 3 trails for each method. All trials start from the same baseline object detectors, which are trained with 5,000 images selected from the unlabeled image pool. After then, each active learning algorithm is executed in 4 rounds. In each round, we select 1,000 images, add these images to the existing training set, and train a new model. Each model is trained with 12 epoches.

\textbf{Results:}
Fig.~\ref{fig:MSCOCO_mAP} and Fig.~\ref{fig:MSCOCO_saving} show the mAP curve and the relative saving of labeled images for the testing methods.
Fig.~\ref{fig:MSCOCO_mAP} shows that classification-only method (C) does not have improvement over passive learning (R), which is not similar to the observations for the PASCAL 2012 in Section~\ref{sec:FRCNN_PASCAL2012} and the PASCAL 2007 in Section~\ref{sec:FRCNN_PASCAL2007}.
By incorporating the localization information, LS+C method can achieve 5\% relative saving in the amount of annotation compared with passive learning, as shown in Fig.~\ref{fig:MSCOCO_saving}.

\begin{figure}[t]
    \centering
    \begin{subfigure}[t]{0.24\textwidth}
        \centering
        \includegraphics[width=\linewidth]{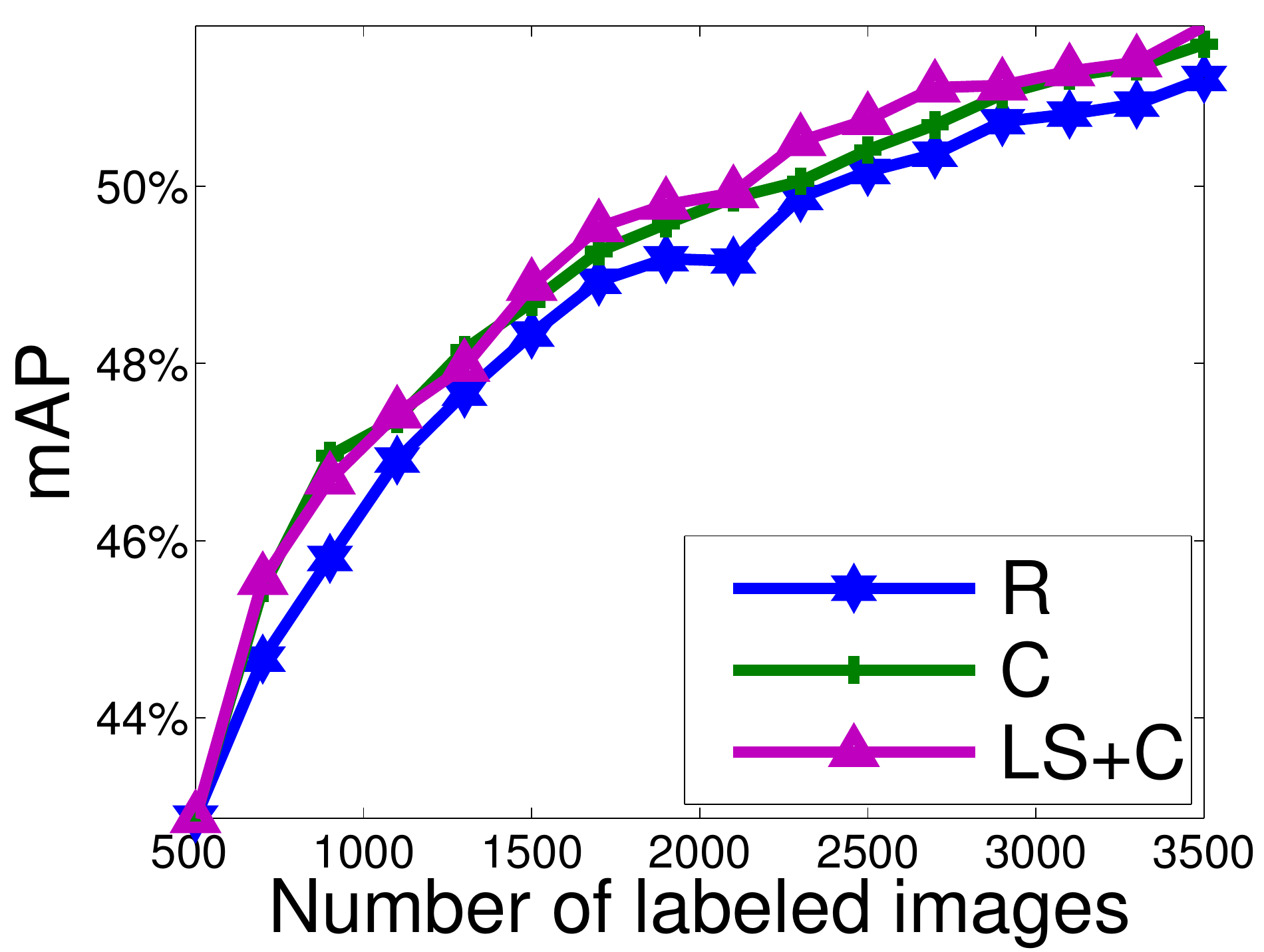}
        \caption{mAP}
        \label{fig:SSD_mAP}
    \end{subfigure}%
    ~
    \begin{subfigure}[t]{0.24\textwidth}
        \centering
        \includegraphics[width=\linewidth]{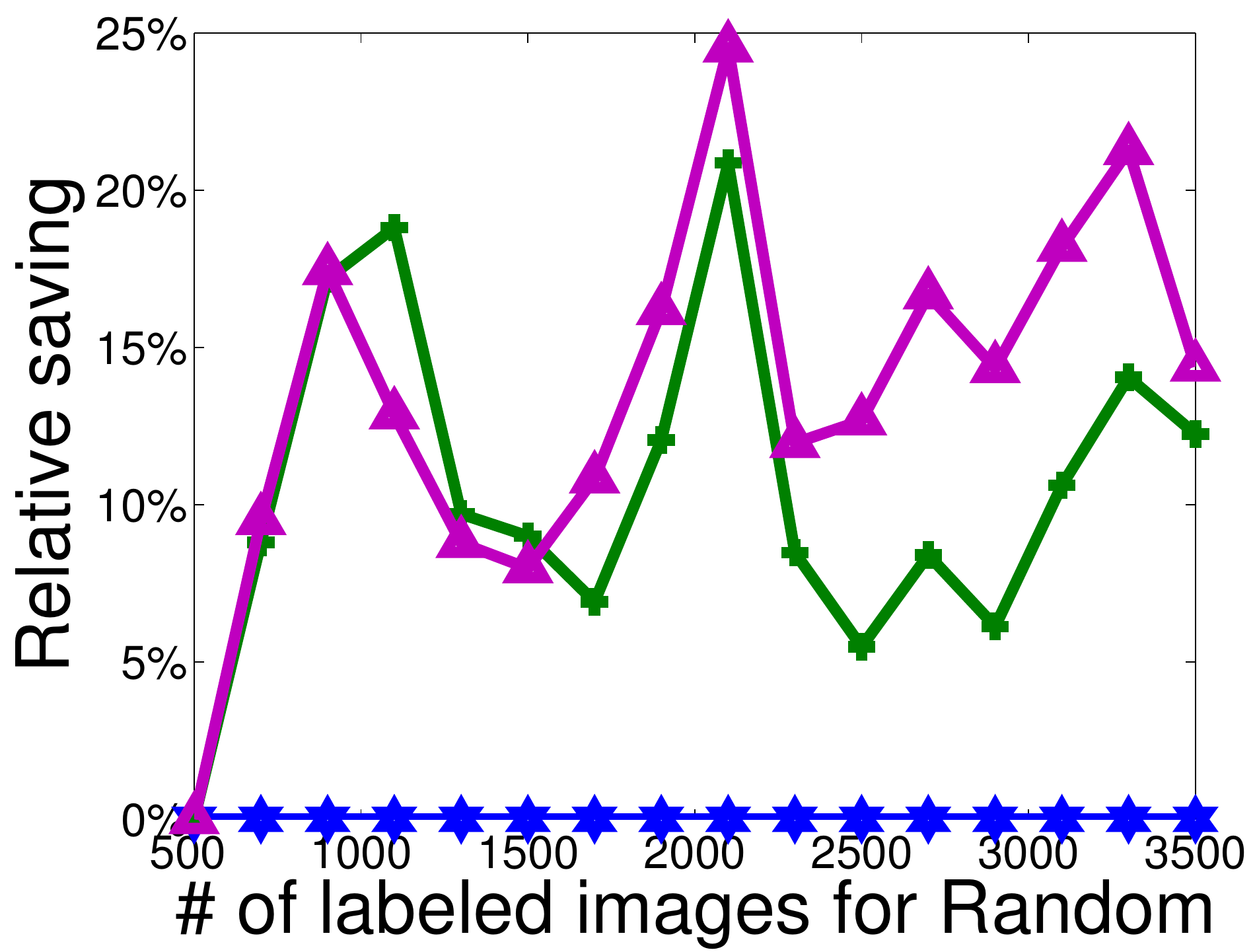}
        \caption{Saving}
        \label{fig:SSD_saving}
    \end{subfigure}
    \caption{(a) Mean average precision curve of SSD with different active learning methods on PASCAL 2007 detection dataset. (b) Relative saving of labeled images for different methods. Each point in the plots is an average of 5 trials.}
\label{fig:SSD}
\end{figure}

\subsection{SSD on PASCAL 2007}
\label{sec:SSD_PASCAL2007}

\textbf{Experimental Setup:}
Here we test our algorithm on a different object detector: the single shot multibox detector (SSD)~\cite{SSD_Liu2016}.
The SSD is a model without an intermediate region-proposal stage, which is not suitable for the localization-tightness based methods.
We test the SSD on the PASCAL 2007 dataset where the training and validation sets (total 5,011 images) are used as the unlabeled image pool, and the test set (4,952 images) is used for testing.
Input images are reiszed to 300$\times$300.

Similar to the experimental settings in Section~\ref{sec:FRCNN_PASCAL2012} and \ref{sec:FRCNN_PASCAL2007}, the numbers shown in this section are averages over 5 trails.
All trials start from the same baseline object detectors which are trained with 500 images selected from the unlabeled image pool. After then, each active learning algorithm is executed in 15 rounds.
A difference from previous experiments is that each model is trained with 40,000 iterations, not a fixed number of epochs. In our experiments, the SSD takes more iterations to converge. Consequently, when the number of labeled images in the training set is small, a fixed number of epochs means training with fewer number of iterations and the SSD cannot converge.

\textbf{Results:}
Fig.~\ref{fig:SSD_mAP} and Fig.~\ref{fig:SSD_saving} show the mAP curve and the relative saving of labeled images for the testing methods. Fig.~\ref{fig:SSD_mAP} shows that both active learning method (C and LS+C) have improvements over passive learning (R).
Fig.~\ref{fig:SSD_saving} shows that in order to achieve the same performance of passive learning with a training set consists of 2,300 to 3,500 labeled images, the proposed method (LS+C) can reduce the amount of image for annoation (12 - 22\%) more than the baseline active learning method (C) (6 - 15\%).
In terms of average saving, LS+C is 29.0\% better than the baseline method C.

\section{Discussion}

\textbf{Extreme Cases:}
There could be extreme cases that the proposed methods may not be helpful.
For instance, if perfect candidate windows are available (LT/C), or feature extractors are resilient to Gaussian noise (LS+C).

If we have very precise candidate windows, which means that we need only the classification part and it is not a detection problem anymore. While this might be possible for few special object classes (e.g. human faces), to our knowledge, there is no perfect region proposal algorithms that can work for all type of objects. As shown in our experiments, even state-of-the-art object detectors can still incorrectly localize objects. Furthermore, when perfect candidates are available, the localization tightness will always be 1, and our LT/C degenerates to classification uncertainty method (C), which can still work for active learning.

Also, we have tested the resiliency to Gaussian noise of state-of-the-art feature extractors (AlexNet, VGG16, ResNet101). Classification task on the validation set of ImageNet (ILSVRC2012) is used as the testbed. The results demonstrate that none of these state-of-the-art feature extractors is resilient to noise. Moreover, if the feature extractor is robust to noise, the localization stability will always be 1, and our LS+C degenerates to classification uncertainty method (C), which can still work for active learning. Please refer to the supplemental material for more details.

\textbf{Estimate of Localization Tightness:}
Our experiment shows that if the ground truth of bounding box is known, localization tightness can achieve best accuracies, 

but the benefit degrades when using the estimated tightness instead. 
To analyze the impact of the estimate, after we trained the FRCNN-based object detector with 500 images of PASCAL2012 training set, we collected the ground-truth-based tightness and the estimated values of all detected boxes in the 5,215 test images. 

\begin{wrapfigure}{r}{0.45\linewidth}
	\includegraphics[width=1.1\linewidth]{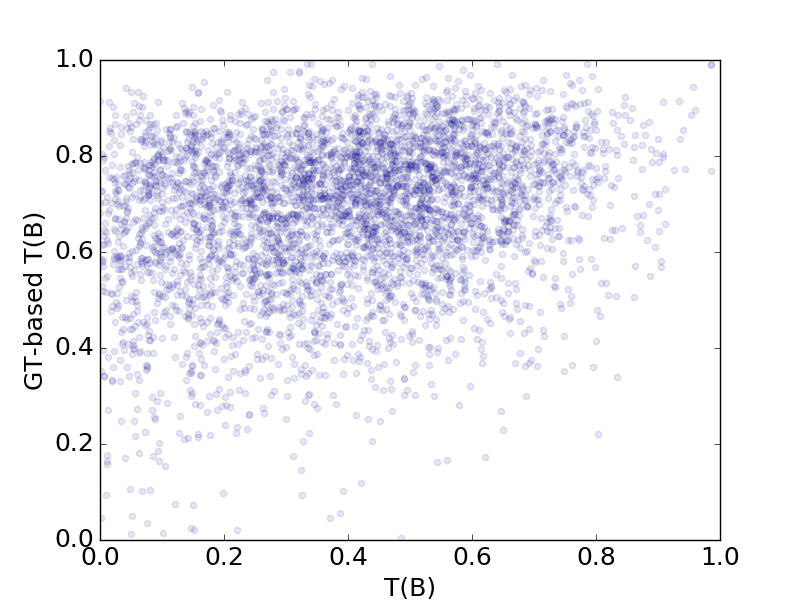}
\end{wrapfigure}

Here shows a scatter plot where the coordinates of each point represents the two scores of a detected box.
As this scatter plot shows an upper-triangular distribution, it implies that our estimate is most accurate when the proposals can tightly match the final detection boxes. 
Otherwise, it could be very different from the ground-truth value. This could partially explain why using the estimated cannot achieve the same performance as the ground-truth-based tightness. 

\textbf{Computation Speed:}
Regarding the speed of our approach, as all testing object detector are CNN-based, the main speed bottleneck lies in the forwarding propagation. In our experiment with FRCNN-based detectors, for instance, forwarding propagation used 137 milliseconds per image, which is 82.5\% of the total time when considering only classification uncertainty. The calculation of $T_I$ has similar speed as $U_C$. The calculation of localization stability $S_I$ needs to run the detector multiple times, and thus is slower than calculating other metrics.

Nevertheless, as these metrics are fully automatic to calculate, using our approach to reduce the number of images to annotate is still cost efficient. Considering that drawing a box from scratch can take 20 seconds in average \cite{Crowdsourcing_Su2012}, and checking whether a box tightly encloses an object can take 2 seconds \cite{HumanVerification_Papadopoulos2016}, the extra overhead to check images with our metrics is small, especially that we can reduce 20 - 25\% of images to annotate.

\section{Conclusion}

In this paper, we present an active learning algorithm for object detection. When selecting unlabeled images for annotation to train a new object detector, our algorithm considers both the classification and localization results of the unlabeled images while existing works mainly consider the classification part alone. We present two metrics to quantitatively evaluate the localization uncertainty, which are how tight the detected bounding boxes can enclose true objects, and how stable the bounding boxes are when adding noise to the image. For object detection, our experiments show that by considering the localization uncertainty, our active learning algorithm can improve the active learning algorithm using the classification outputs only. As a result, we can train object detectors to achieve the same accuracy with fewer annotated images.

\section*{Acknowledgments}

This work was conducted during the first author's internship in Mitsubishi Electric Research Laboratories. This work was sponsored in part by National Science Foundation grant \#13-21168.

{\small
\bibliographystyle{ieee}
\bibliography{bibfile}
}

\clearpage 

\section*{Supplementary Materials}
This document includes the data and analysis of the proposed methods that are not covered in the main paper due to the spcae limitation.
We first define the abbreviation for all methods as following:
\begin{center}
\begin{tabular}{ |l|l| }
  \hline
  Abbreviation & Method\\
  \hline
  R & Random \\
  \hline
  C & Classification \\
  \hline
  LS & Localization Stability \\
  \hline
  LS+C & Localization Stability and Classification \\
  \hline
  LT/C & Localization Tightness and Classification \\
  \hline
  LT/C(GT) &  Localization Tightness and Classification \\
     & with Ground Truth \\
  \hline
  3in1 &  Localization Stability, Localization \\
  			& Tightness, and Classification \\ 
  \hline
\end{tabular}
\end{center}
This abbreviation is used in all the text, figures, and tables in this document.

\section*{A. Design of Localization-Tightness Metric}
Given the measurement of localization tightness, we need to design a metric to utilize it for active learning.
The most intuitive way is to use the localization tightness alone to decide the score for each box.
However, in our experiments it does not help for selecting samples to annotate.
We further analyze it by showing the images selected by different methods as shown in Fig.~\ref{fig:LT_and_C}.
When using only localization tightness as the cue to calculate the score of each detected box for active learning, it tends to find images (Fig.~\ref{fig:LT_and_C}, first row) that have tiny objects (e.g., airplane, bird), which are not chosen that often by other methods  (Fig.~\ref{fig:LT_and_C}, second row).
However, these classes are easier ones that the detector already does well so that the overall performance of using localization tightness alone is worse than other metrics.  
\begin{figure}[t]
\begin{center}
   	\includegraphics[width=0.32\linewidth]{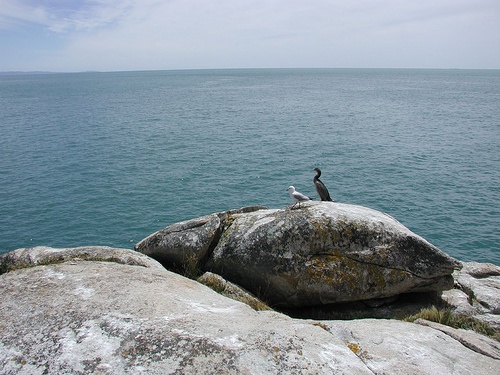}
	\includegraphics[width=0.32\linewidth]{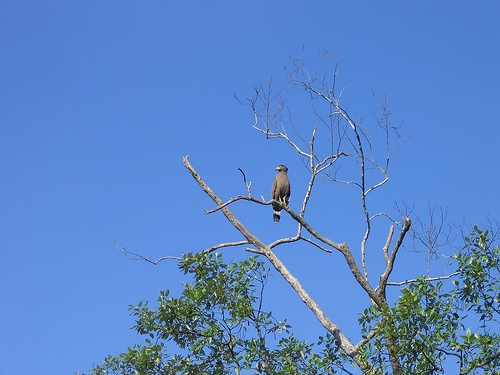}
	\includegraphics[width=0.32\linewidth]{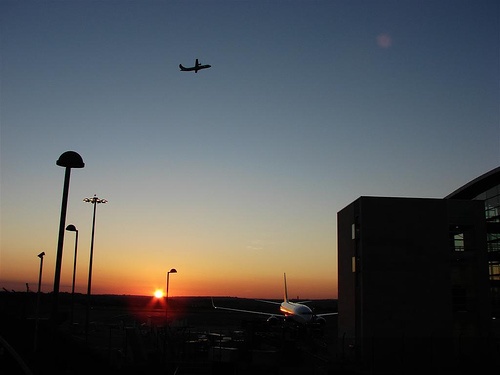}
	\\
   	\includegraphics[width=0.32\linewidth]{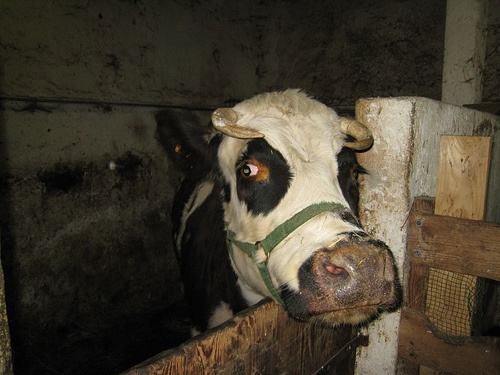}
	\includegraphics[width=0.32\linewidth]{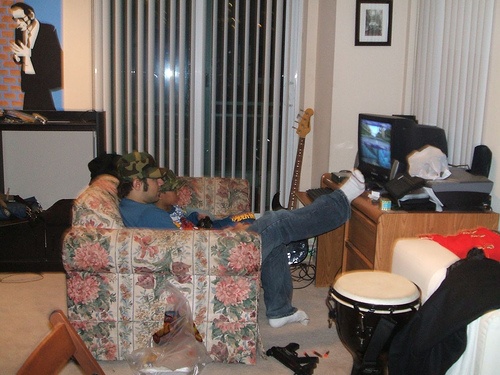}
	\includegraphics[width=0.32\linewidth]{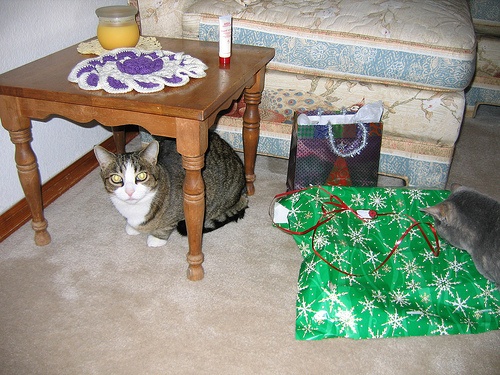}	
\end{center}
   \caption{\textbf{First row:} Example images selected for annotation by the method using information from localization only to evaluate the score of each box. \textbf{Second row:} Example images selected for annotation by the method using classification uncertainty only (C).}
\label{fig:LT_and_C}
\end{figure}

Based on the observations in Sec.~3.2 in the main paper, we would like to find images contain boxes that have disagreement in classification and localization results.
When designing a metric using localization tightness, there are two important qeustions: "How to define the score for an image with detected boxes?" and "How to define the score for a detected box?"
For the first question, two methods have been tested: using the lowest score of all boxes (min(.)), and using a weighted sum of all boxes (wsum(.)), where the weight is $P_{max}$ of each box.
For the second question, different metrics have been tested as following, where P ($P_{max}(B)$ in the main paper) is the highest probability out of $K$ categories of box $B$, and T ($T(B)$ in the main paper) is the localization tightness of box $B$.
For a set of unlabeled images, the following methods choose images with lower scores to annotate in active learning.

\begin{description}
\item[min(\textbar T+P-1\textbar)] This metric is the one (\textbf{LT/C}) we used in the main paper. It selects images with boxes that have disagreement between classification and localization results. It also picks images contain boxes that are not very certain in both classification and localization results.
\item[min(-\textbar P-T\textbar)] Different from LT/C, this metric only selects images with boxes that have disagreement between classification and localization results. It does not select boxes that are not very certain in both two outputs. 
\item[wsum(\textbar T+P-1\textbar)] This method uses the same metric as LT/C to evaluate the score of each box. However, instead of using the highest score out of all boxes as the score of an imgae, it uses a weighted sum across all boxes.
\item[wsum(T)] This method uses only the information from localization outputs when deciding the score of each box. Images with boxes that have low localization tightness will be chosen by this method.
\end{description}

\begin{figure}[t]
\begin{center}
   \includegraphics[width=0.9\linewidth]{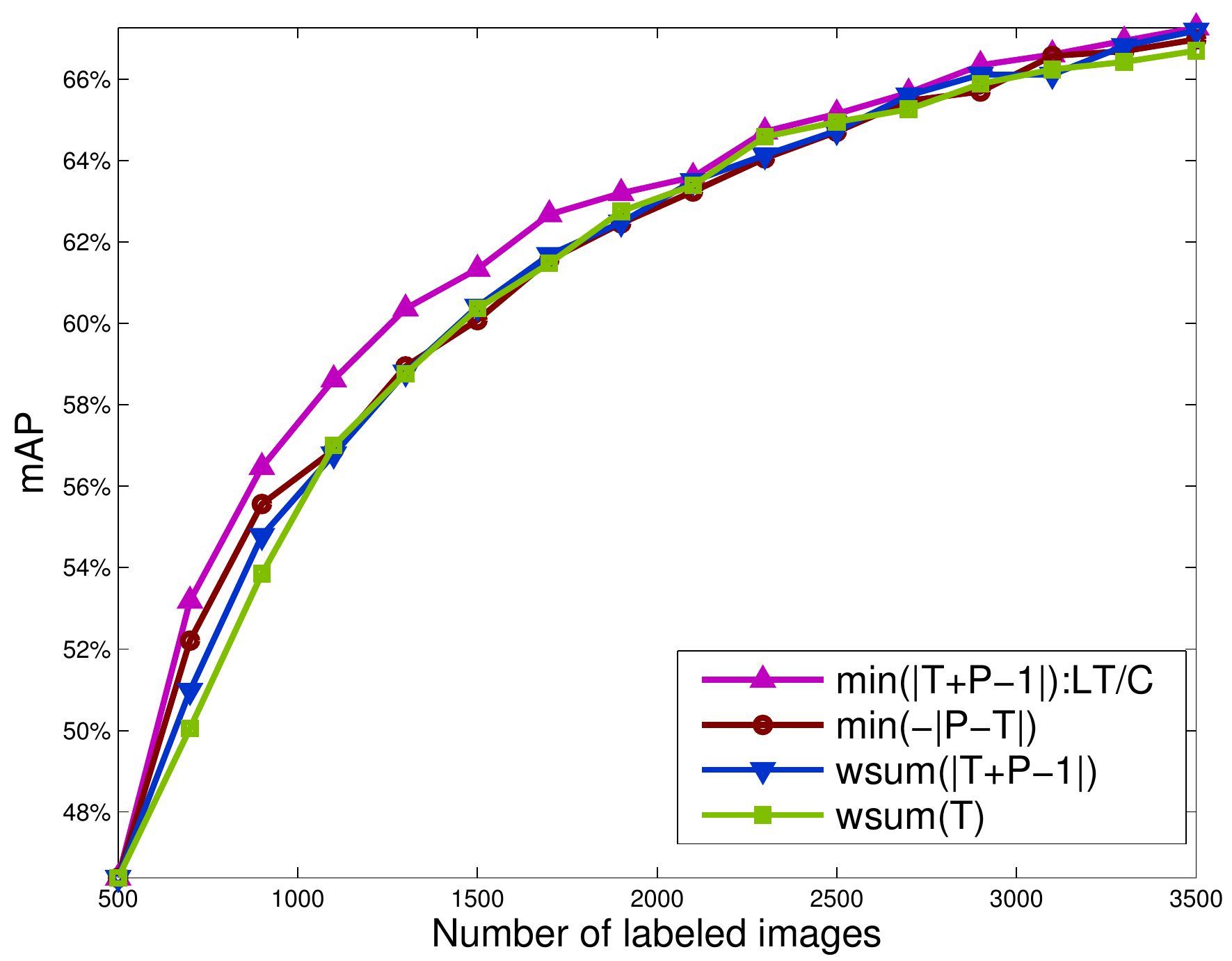}
\end{center}
   \caption{Mean average precision curve of different metrics of localization tightness on PASCAL 2007 detection dataset. Each point in the plot is an average of 5 trials.}
\label{fig:PASCAL2007_LT_mAP}
\end{figure}

Fig.~\ref{fig:PASCAL2007_LT_mAP} shows the mean average precision (mAP) curves of different metrics using localization tightness, and the experimental setup is the same as mentioned in Sec.~4.2 in the main paper.
The proposed LT/C outperforms the rest metrics clearly at the first half of the experiment.
Among the second half, LT/C is still the best among all metrics, but the gap between LT/C and the others becomes smaller.

The difference between LT+C and max(\textbar P-T\textbar) is selecting images with boxes that are both uncertain in classification and localization outputs.
We hypothesize that images with uncertainty in both outputs are more informative, which make LT/C better than max(\textbar P-T\textbar).
Also, given the same metric for calculating the score of a detected box, LT/C and wsum(\textbar T+P-1\textbar) use different strategy to define the score of an image.
The overlapping ratio of images sampled by these two methods is only 17.9\% (an average over 5 trials), which implies that how to define the score of an image greatly affects the sampling process.


\section*{B. Discussion of Extreme Cases}
\begin{figure}[t]
\begin{center}
   \includegraphics[width=0.99\linewidth]{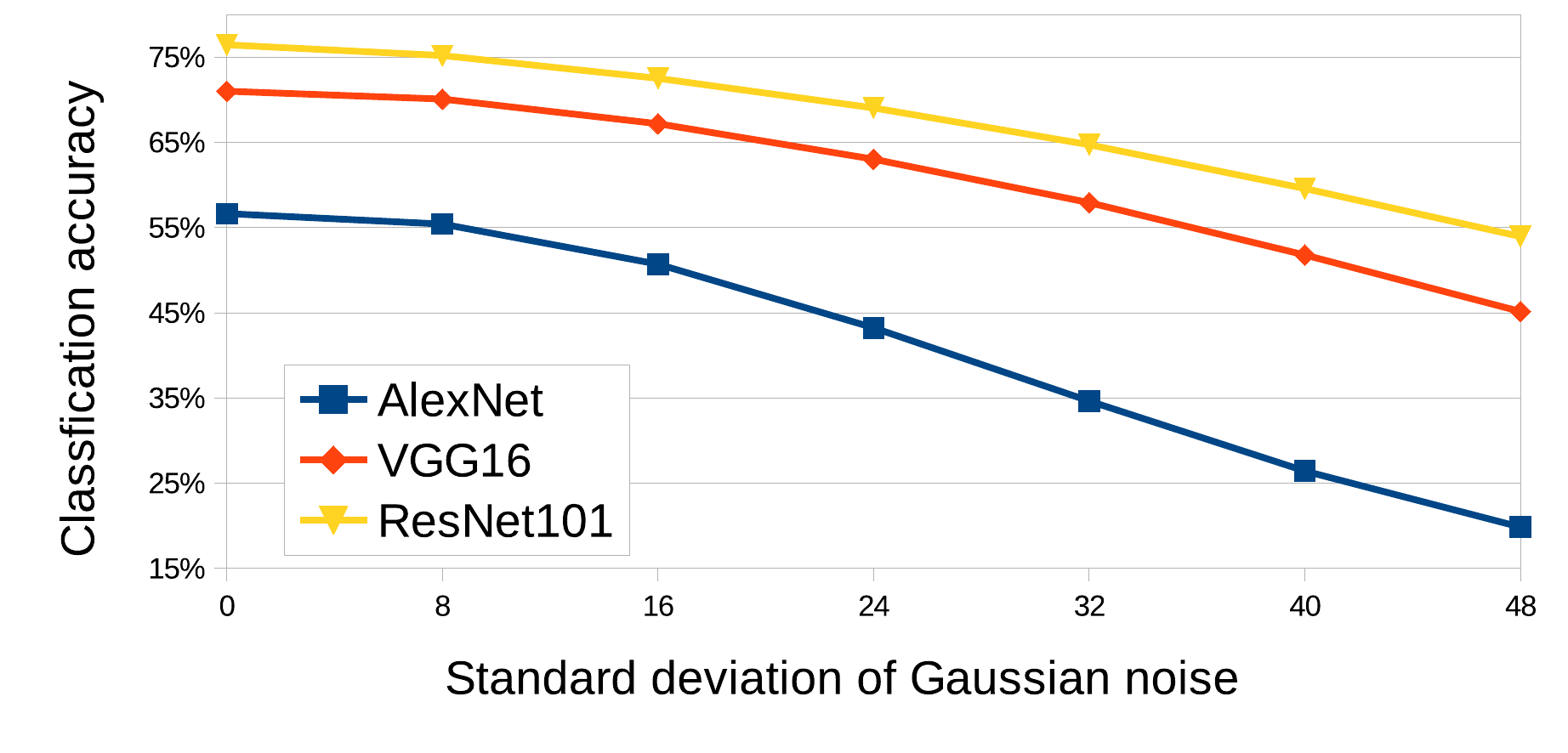}
\end{center}
   \caption{Top-1 classification accuracy of different neural network models when input images are corrupted by Gaussian noise on PASCAL 2012 validation dataset.}
\label{fig:PASCAL2012_NoiseTest}
\end{figure}
As mentioned in Sec. 5 in the main paper, there could be extreme cases that the proposed methods may not be helpful.
For instance, if perfect candidate windows are available (LT/C), or feature extractors are resilient to Gaussian noise (LS+C).

We had discussed the case of perfect candidate windows in the main paper. In the following, we discuss more about the case of feature extractors are resilient to noise.
We have tested the resiliency to Gaussian noise of state-of-the-art feature extractors (AlexNet, VGG16, ResNet101). Classification task on the validation set of ImageNet (ILSVRC2012) is used as the testbed. Pre-trained models are used as the classifier and input images are corrupted by Gaussian noise of different levels. Fig.~\ref{fig:PASCAL2012_NoiseTest} shows the top-1 classification accuracy under different standard deviation of Gaussian noise. With the largest standard deviation, the accuracy can drop 23-37\%. It demonstrates that none of these state-of-the-art feature extractors is resilient to noise.
Goodfellow et. al~\cite{Goodfellow_2014} also hypothesized that NNs with non-linear modules (e.g., sigmoid) mainly work in linear region, could be vulnerable to local perturbation such as Gaussian noise.


\begin{figure*}[t]
\begin{center}
   \includegraphics[width=0.95\linewidth]{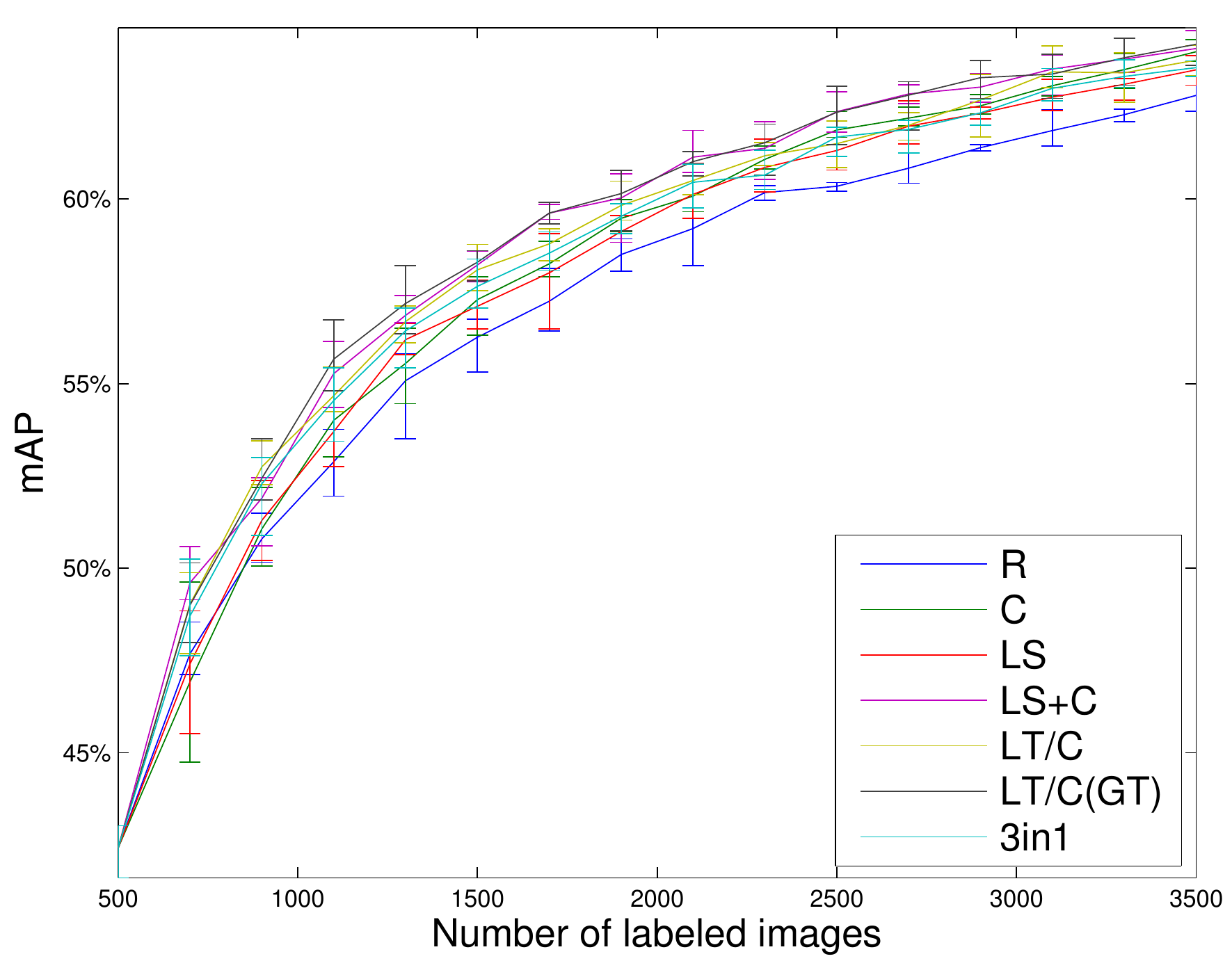}
\end{center}
   \caption{Mean average precision curve of different active learning methods on the PASCAL 2012 detection dataset. Each point in the plot is an average of 5 trials. The error bars represent the minimum and maximum values out of 5 trials at each point. This is a full version (LS and 3in1 added) of Fig.~5a in the main paper.}
\label{fig:PASCAL2012_mAP_full}
\end{figure*}

\begin{figure*}[t]
\begin{center}
   \includegraphics[width=0.95\linewidth]{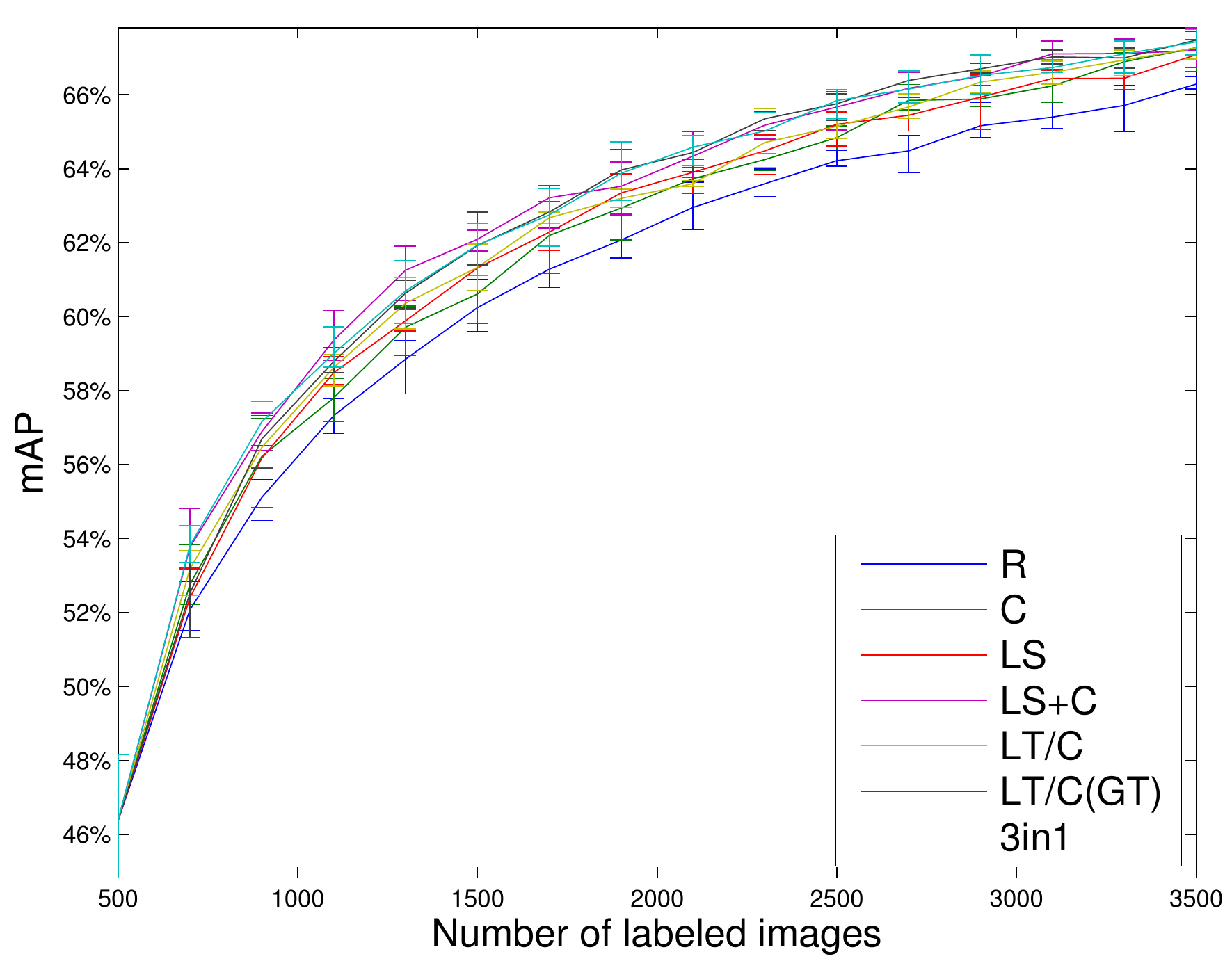}
\end{center}
   \caption{Mean average precision curve of different active learning methods on the PASCAL 2007 detection dataset. Each point in the plot is an average of 5 trials. The error bars represent the minimum and maximum values out of 5 trials at each point. This is a full version (LS and 3in1 added) of Fig.~7a in the main paper.}
\label{fig:PASCAL2007_mAP_full}
\end{figure*}

\section*{C. Full Experimental Results }
In this section, the full results from the experiments of active learning methods on the PASCAL and MS COCO datasets are presented. 
These results are not covered in the main paper due to the easiness of reading and space constraint.

\begin{figure*}[t]
    \centering
    \begin{subfigure}[t]{0.49\textwidth} 
        \centering
        \includegraphics[width=\linewidth]{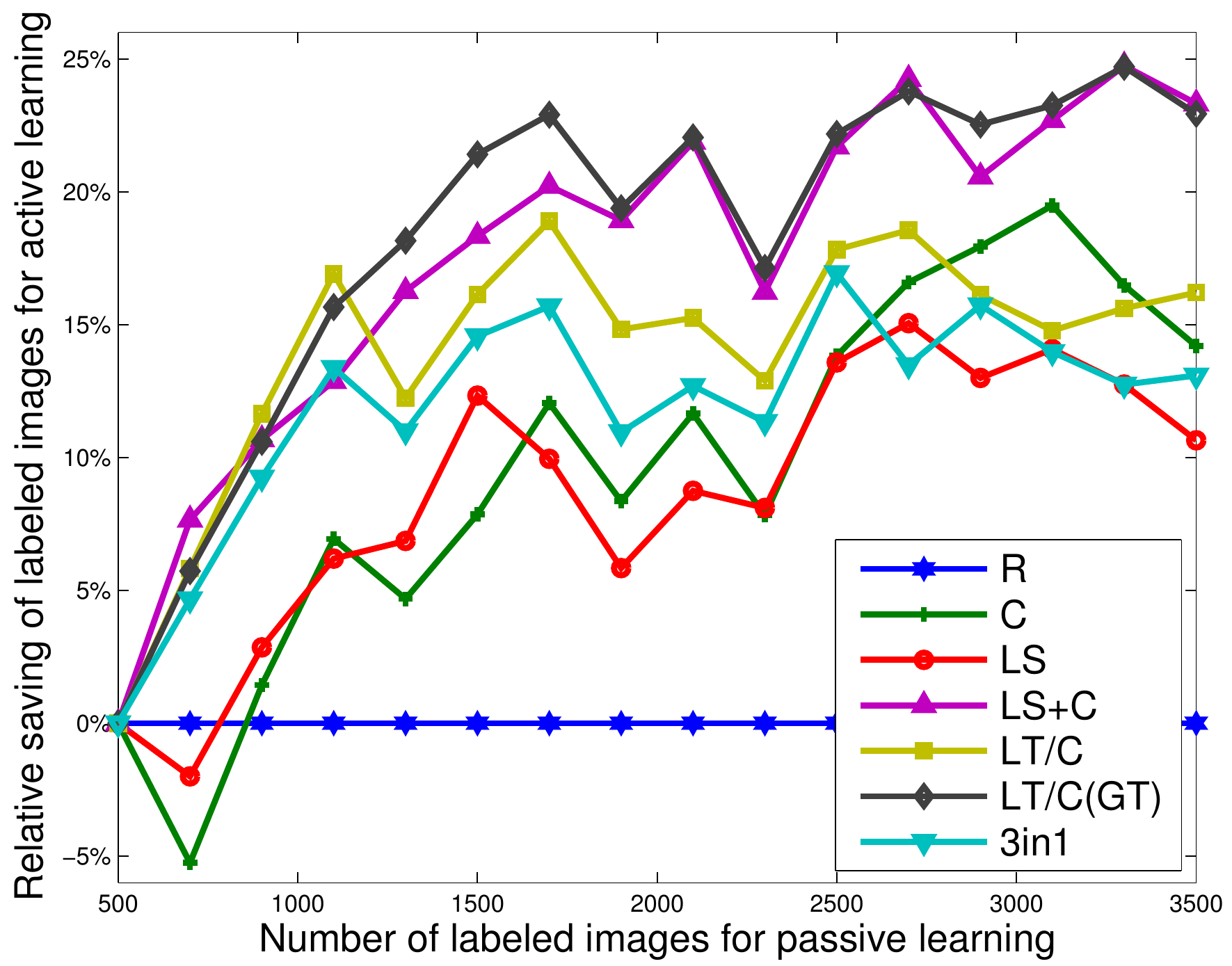}
        \caption{PASCAL 2012}
        \label{fig:PASCAL2012_saving_full}
    \end{subfigure}%
    ~
    \begin{subfigure}[t]{0.49\textwidth} 
        \centering
        \includegraphics[width=\linewidth]{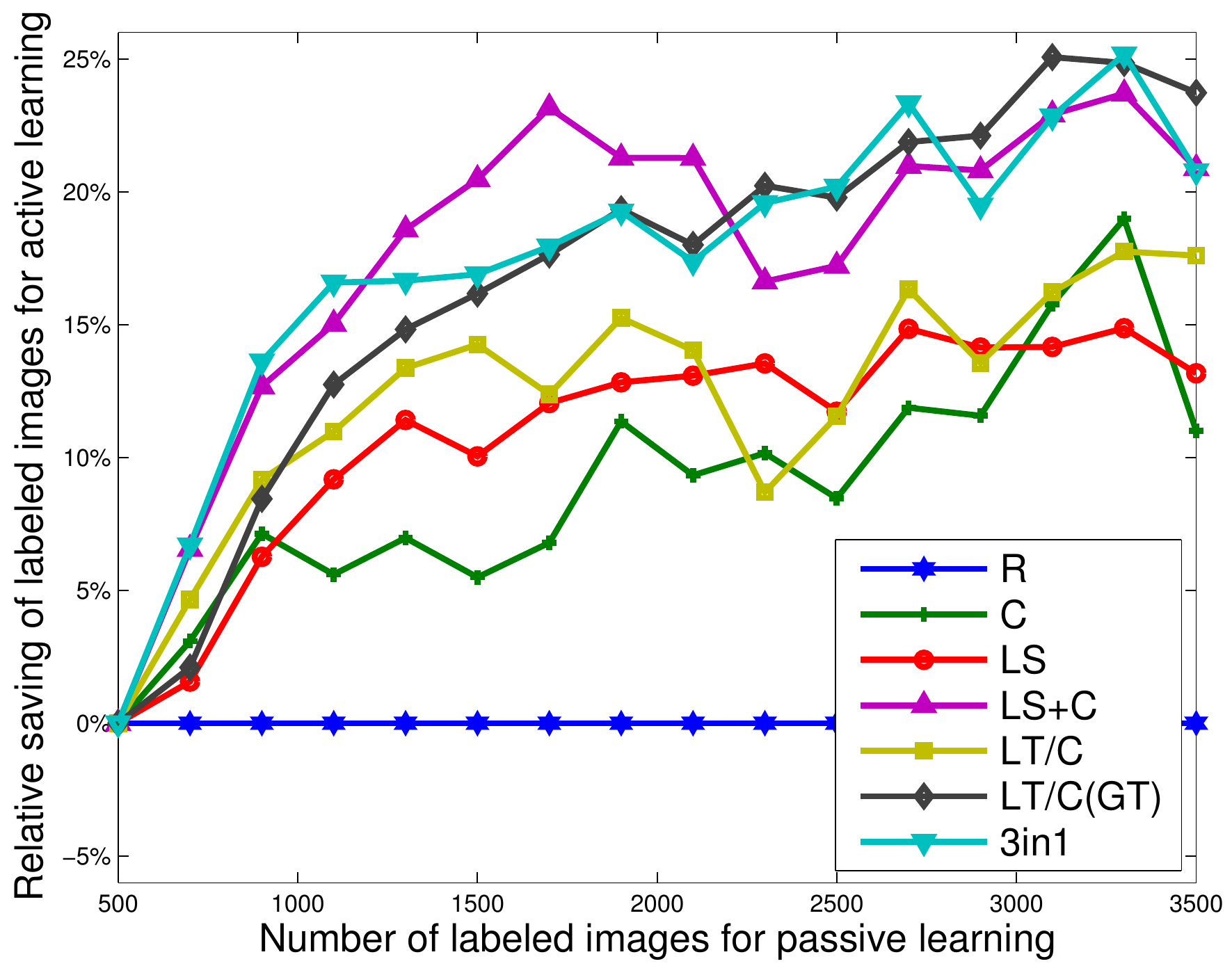}
        \caption{PASCAL 2007}
        \label{fig:PASCAL2007_saving_full}
    \end{subfigure}
    \caption{Relative saving of labeled images for different active learning methods on the (a) PASCAL 2012 validation dataset and (b) PASCAL 2007 testing set. (a) and (b) are full versions (LS and 3in1 added) of Fig.~5b and Fig.~7b in the main paper.}
\label{fig:PASCAL_saving_full}
\end{figure*}

\paragraph{Results of Using Localization Stability Only:}
As an ablation experiment, results for the method using localization stability only (LS) are added into the plot of mAP curves and the table of classwise APs.
Table~\ref{tab:PASCAL2012_1100imgs_full} and Table~\ref{tab:PASCAL2007_1100imgs_full} show the average precision for each method after 3 rounds of active learning on the PSACAL 2012 validation and PASCAL 2007 testing set.
Fig.~\ref{fig:PASCAL2012_mAP_full} and Fig.~\ref{fig:PASCAL2007_mAP_full} show the mAP curves of each active learning method on the PASCAL 2012 and 2007 datasets.
Each point in the plot is an average of 5 trials. 
Also, error bars that represent the minimum and maximum values out of 5 trials are added at each point to show the distribution of 5 trials.
Fig.~\ref{fig:PASCAL2012_saving_full} and Fig.~\ref{fig:PASCAL2007_saving_full} show the relative saving in labeled images of each active learning method on the PASCAL 2012 and 2007 datasets.
As shown in Fig.~\ref{fig:PASCAL2012_saving_full} and Fig.~\ref{fig:PASCAL2007_saving_full}, LS outperforms the random sampling for the most cases.
Also, combining the localization stability with the classification uncertainty (LS+C) works better than using either only the localization stability (LS) or classification uncertainty (C).

\begin{table*}[htbp]
\center
\begin{scriptsize}
\tabcolsep=0.11cm\begin{tabular*}{0.93\linewidth}{l|*{20}{c}|c}
method &aero&bike&bird&boat*&bottle*&bus&car&cat&chair*&cow&table*&dog&horse&mbike&persn&plant*&sheep&sofa&train&tv&mAP\\
\hline
R&71.1 &61.5 &54.7 &28.4 &32.0 &\underline{68.1} &57.9 &75.4 &25.8 &44.2 &36.4 &73.0 &61.9 &67.3 &68.1 &21.6 &51.9 &41.0 &\textbf{65.5} &51.7 &52.9 \\
C&70.7 &62.9 &54.7 &25.5 &30.8 &66.1 &56.2 &\textbf{78.1} &26.4 &\textbf{54.5} &36.7 &\textbf{76.9} &\textbf{68.3} &\underline{67.7} &67.4 &22.5 &\underline{57.7} &40.8 &63.6 &52.5 &54.0 \\
LS&\textbf{75.1} &61.3 &\textbf{57.6} &\textbf{34.7} &\underline{35.1} &65.1 &58.2 &75.4 &29.3 &43.9 &38.5 &70.7 &57.5 &66.1 &68.5 &23.0 &56.1 &40.3 &64.2 &53.6 &53.7 \\
LS+C&\underline{73.9} &63.7 &\underline{56.9} &\underline{29.6} &\textbf{35.2} &66.5 &\underline{58.5} &\underline{77.9} &\underline{31.3} &\underline{50.8} &40.7 &\underline{73.8} &\underline{65.4} &66.9 &68.4 &\textbf{24.8} &\textbf{58.0} &\textbf{44.9} &64.2 &53.9 &\textbf{55.3} \\
LT/C&69.8 &\textbf{64.6} &54.6 &29.5 &33.8 &\textbf{70.3} &\textbf{59.7} &75.5 &29.5 &46.3 &\textbf{41.8} &73.0 &62.5 &\textbf{69.0} &\textbf{70.8} &23.2 &56.5 &42.8 &\underline{64.3} &\underline{55.9} &\underline{54.7} \\
3in1&72.9 &\underline{63.8} &52.7 &29.5 &33.6 &66.4 &57.2 &76.0 &\textbf{31.5} &48.5 &\underline{41.6} &72.2 &62.6 &67.6 &\underline{68.8} &\underline{24.5} &57.6 &\underline{43.6} &63.0 &\textbf{57.1} &54.5 \\
\end{tabular*}
\end{scriptsize}
\caption{Average precision for each method on the PASCAL 2012 validation set after 3 rounds of active learning (the number of labeled images in the training set is 1,100). This is a full version (LS and 3in1 added) of Table~1 in the main paper. All the experimental settings are the same with Table~1 in the main paper.}
\label{tab:PASCAL2012_1100imgs_full}
\end{table*}

\begin{table*}[htbp]
\center
\begin{scriptsize}
\tabcolsep=0.11cm\begin{tabular*}{0.93\linewidth}{l|*{20}{c}|c}
method &aero&bike&bird&boat*&bottle*&bus&car&cat&chair*&cow&table*&dog&horse&mbike&persn&plant*&sheep&sofa&train&tv&mAP\\
\hline
R&\underline{61.6} &67.2 &54.1 &40.0 &33.6 &64.5 &73.0 &73.9 &34.5 &60.8 &52.2 &69.3 &\textbf{74.7} &66.6 &67.1 &25.9 &52.1 &54.2 &\underline{66.1} &54.9 &57.3 \\
C&56.9 &\underline{68.0} &54.9 &36.8 &34.4 &\underline{68.1} &71.7 &\textbf{75.5} &34.0 &\textbf{68.6} &51.0 &\underline{71.4} &\underline{74.7} &65.2 &65.9 &24.9 &\underline{60.0} &53.9 &63.0 &57.4 &57.8 \\
LS&\textbf{64.4} &63.9 &\textbf{56.3} &\textbf{45.1} &38.0 &65.5 &73.7 &71.2 &38.6 &62.7 &57.0 &67.6 &69.0 &64.6 &67.1 &\textbf{29.6} &56.2 &\underline{57.3} &\textbf{68.6} &53.6 &58.5 \\
LS+C&61.5 &64.4 &\underline{55.8} &40.2 &\underline{38.7} &66.3 &73.8 &\underline{74.7} &\underline{39.6} &\underline{68.0} &56.3 &\textbf{71.5} &73.8 &\underline{67.2} &66.7 &27.7 &\textbf{61.3} &57.0 &65.6 &57.4 &\textbf{59.4} \\
LT/C&57.6 &\textbf{69.7} &52.9 &\underline{41.1} &38.4 &\textbf{69.7} &\underline{74.4} &71.8 &36.4 &61.2 &\underline{58.1} &69.5 &74.3 &66.2 &\textbf{67.8} &\underline{28.0} &55.5 &56.3 &65.5 &\underline{58.2} &58.6 \\
3in1&57.6 &65.1 &53.3 &37.1 &\textbf{39.0} &68.0 &\textbf{74.6} &73.9 &\textbf{39.8} &64.9 &\textbf{58.5} &70.4 &73.7 &\textbf{67.3} &\underline{67.3} &27.4 &59.9 &\textbf{58.0} &65.1 &\textbf{59.2} &\underline{59.0} \\
\end{tabular*}
\end{scriptsize}
\caption{Average precision for each method on the PASCAL 2007 testing set after 3 rounds of active learning (the number of labeled images in the training set is 1,100). This is a full version (LS and 3in1 added) of Table~2 in the main paper. All the experimental settings are the same with Table~2 in the main paper.}
\label{tab:PASCAL2007_1100imgs_full}
\end{table*}

\paragraph{Results of Using 3 Cues:}
In order to see that if the localization-uncertainty measurements have complementary information, we further combine all cues for selecting informative images. As images with high classification uncertainty, low localization stability, and low localization tightness should be selected for annotation, the score of the $i$-th image ($I_i$) image is defined as follows:
$U_{C}(I_i) - \lambda_{ls} S_{I}(I_i) - \lambda_{lt} T_{I}(I_i)$
where $\lambda_{ls}$ and $\lambda_{lt}$ are set to 1 across all the experiments in this paper.

On PASCAL 2012, combining all cues together does not work better than either LS+C or LT/C (Fig.~\ref{fig:PASCAL2012_saving_full}). On PASCAL 2007, 3in1 is compatible with LS+C, and better than LT/C (Fig.~\ref{fig:PASCAL2007_saving_full}).
It seems that localization-uncertainty measurements do not have complementary information.
We further analyze the overlapping ratio between images chosen by different active learning methods in Table~\ref{tab:PASCAL2012_overlap} and Table~\ref{tab:PASCAL2007_overlap}.
When we compare the overlapping ratio between 3in1 and three other metrics (C, LS, LT/C), both C and LS have an overlapping ratio around 30\%, but LT/C has only about 10\%.
This implies that among the three cues, LT/C provides the least information in 3in1 method.
We notice that the images chosen by 3in1 method are highly overlapped with LS+C (over 60\%), but 3in1 does not outperform LS+C.
Our hypothesis is that the images (about one third of total images) chosen differently by 3in1 and LS+C make this difference in performance.

\begin{table}[htbp]
\center
\begin{scriptsize}
\begin{tabular*}{0.748\linewidth}{|c|c|c|c|c|c|}
\cline{1-2}
Method & R\\
\cline{1-3}
C & 3.5\% & C \\
\cline{1-4}
LS & 4.0\% & 2.7\%& LS \\
\cline{1-5}
LS+C& 4.4\% & 34.7\%& 34.6\% &LS+C\\
\cline{1-6}
LT/C& 5.0\% & 5.9\%& 2.4\% & 5.2\%&LT/C\\
\hline
3in1& 4.6\% & 30.4\%& 25.7\% & 62.4\%&8.8\%\\
\hline
\end{tabular*}
\end{scriptsize}
\caption{Overlapping ratio between 200 images chosen by different active learning methods on the PASCAL 2012 dataset after the first round of active learning. Each number shown in the table is an average over 5 trials.}
\label{tab:PASCAL2012_overlap}
\end{table} 

\begin{table}[htbp]
\center
\begin{scriptsize}
\begin{tabular*}{0.7628\linewidth}{|c|c|c|c|c|c|}
\cline{1-2}
Method & R\\
\cline{1-3}
C & 4.1\% & C \\
\cline{1-4}
LS & 4.2\% & 3.5\%& LS \\
\cline{1-5}
LS+C& 4.3\% & 34.0\%& 39.7\% &LS+C\\
\cline{1-6}
LT/C& 5.6\% & 5.9\%& 4.5\% & 5.7\%&LT/C\\
\hline
3in1& 3.9\% & 30.5\%& 32.0\% & 65.3\%&12.0\%\\
\hline
\end{tabular*}
\end{scriptsize}
\caption{Overlapping ratio between 200 images chosen by different active learning methods on the PASCAL 2007 dataset after the first round of active learning. Each number shown in the table is an average over 5 trials.}
\label{tab:PASCAL2007_overlap}
\end{table}

\begin{figure}[t]
\begin{center}
   \includegraphics[width=0.95\linewidth]{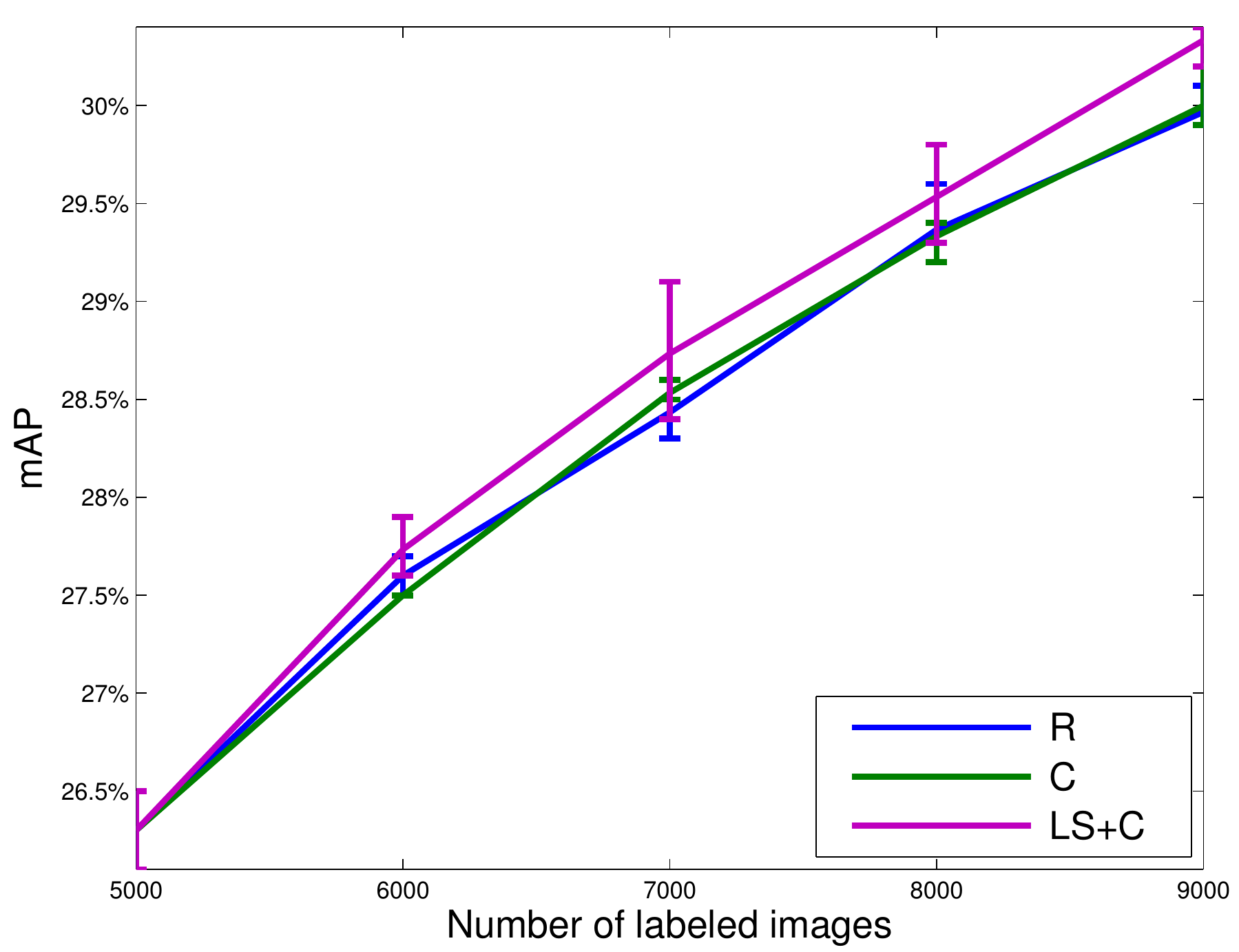}
\end{center}
   \caption{Mean average precision curve of different active learning methods on the MS COCO validation set. Each point in the plot is an average of 3 trials. The error bars represent the minimum and maximum values out of 3 trials at each point. This is a full version of Fig.~9a in the main paper.}
\label{fig:MSCOCO_mAP_full}
\end{figure}

\begin{figure}[t]
\begin{center}
   \includegraphics[width=0.95\linewidth]{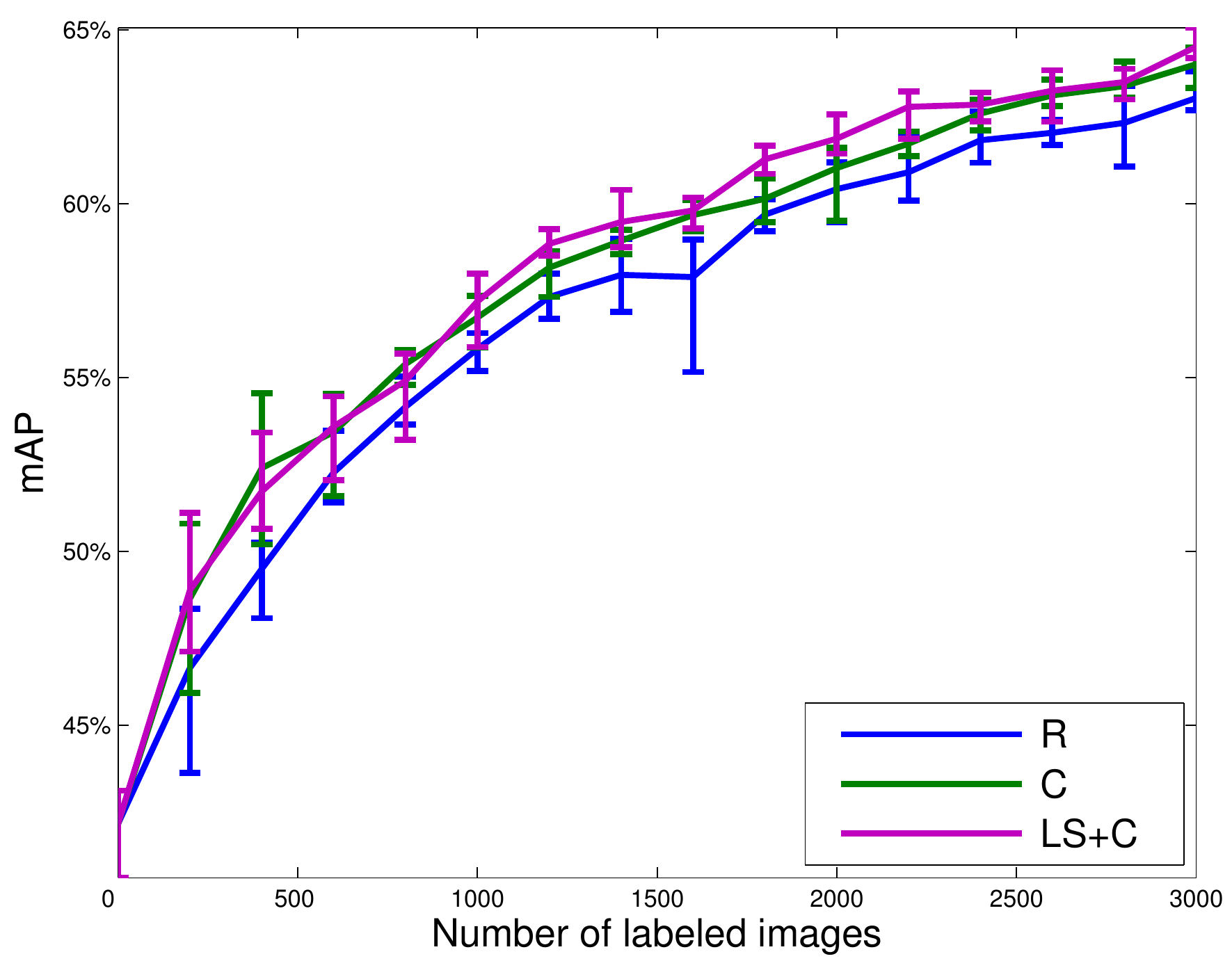}
\end{center}
   \caption{Mean average precision curve of different active learning methods with SSD on the PASCAL 2007 testing set. Each point in the plot is an average of 5 trials. The error bars represent the minimum and maximum values out of 5 trials at each point. This is a full version of Fig.~10a in the main paper.}
\label{fig:SSD_mAP_full}
\end{figure}

\paragraph{mAP Plots with Error Bars:}
In the original mAP plots of the FRCNN on the MS COCO dataset (Fig.~8a in the main paper) and the SSD on the PASCAL 2007 dataset (Fig.~9a in the main paper), only the average of multiple trials is plotted.
Here we add the error bars that represent the minimum and maximum values of multiple trials to the plot.
This shows the distribution of the result from different trials.
Fig.~\ref{fig:MSCOCO_mAP_full} and Fig.~\ref{fig:SSD_mAP_full} show the mAP curves of the FRCNN on the MS COCO dataset and the SSD on the PASCAL 2007 dataset.
Three methods (R, C, and LS+C) are tested in these two experiments.

\section*{D. Visualization of The Selection Process}
The most popular metric used for measuring the performance of an object detector is mAP.
We also use this metric to evaluate the performance of different active learning methods.
If one active learning method selects more informative images to label and add them into the training set, the detector trained on this set will have a higher mAP.
Besides this final numerical result, we are curious about what images are chosen in the selection process by different active learning methods, and how these chosen images are related to the average precision.

In order to visualize the selection process, we first visualize the PASCAL 2012 training set~\cite{PASCAL} by using t-Distributed Stochastic Neighbor Embedding (t-SNE)~\cite{t-SNE_Maaten2008}.
After knowing the distibution of the PASCAL 2012 training set, we further visualize the chosen images in the selection process by different active learning methods.

\paragraph{Visualization of the PASCAL 2012 Dataset:}
We first visualize the PASCAL 2012 training set (5,717 images) by using t-SNE with VGG16 model~\cite{VGG}.
t-SNE is a technique for dimensionality reduction that is tailored for visualizing high-dimensional datasets.
Features extracted from the conv5\_3 layer are used as the high-dimensional vector for each image in the PASCAL 2012 training set.
The visualization of the PASCAL 2012 training set by embedding each image to a point on the 2D plane is shown in Fig.~\ref{fig:full_training_set_with_imgs}.
Each data point in Fig.~\ref{fig:full_training_set_with_imgs} represents one image in the dataset.
Images with objects from only one class are represented by markers other than dots.
Note that there might be objects belong to different classes shown in one image.
Red dots (\textgreater{1cls}) are used for representing those images.
For each class, there is a certain region that images locate at. 
For example, images of aeroplanes (orange plus signs) are located at the top-right part, and images of cats (green squares) are located at the bottom-center part.

For those images have objects from muliple classes, we cannot tell what classes are included in each of them from Fig.~\ref{fig:full_training_set_with_imgs}.
Therefore, another visualization is shown in Fig.~\ref{fig:full_training_set_classwise} by considering whether one image has objects from a certain class or not.
For example, each orange plus sign in Fig.~\ref{fig:training_aero} represents an image which has at least one aeroplane in it, and each black dot represents an image that has no aeroplane in it.
Given Fig.~\ref{fig:full_training_set_with_imgs} and Fig.~\ref{fig:full_training_set_classwise}, we now have a better understanding about the distribution of the dataset, and the relationship between different classes.
For example, in the left part of the scatter plot in Fig.~\ref{fig:full_training_set_with_imgs}, we notice that there are many images that have objects belong to multiple classes (red dots).
From Fig.~\ref{fig:full_training_set_classwise}, we know that these images may contain people, chairs, tables, sofas, bottles, plants, and TVs.
Actually, these images are regular scenes in a living room, just like the 4 images shown in Fig.~\ref{fig:full_training_set_with_imgs}.
With these information, we can further analyze the selection process of different active learning methods.

\begin{figure}[t]
    \centering
    \includegraphics[width=\linewidth]{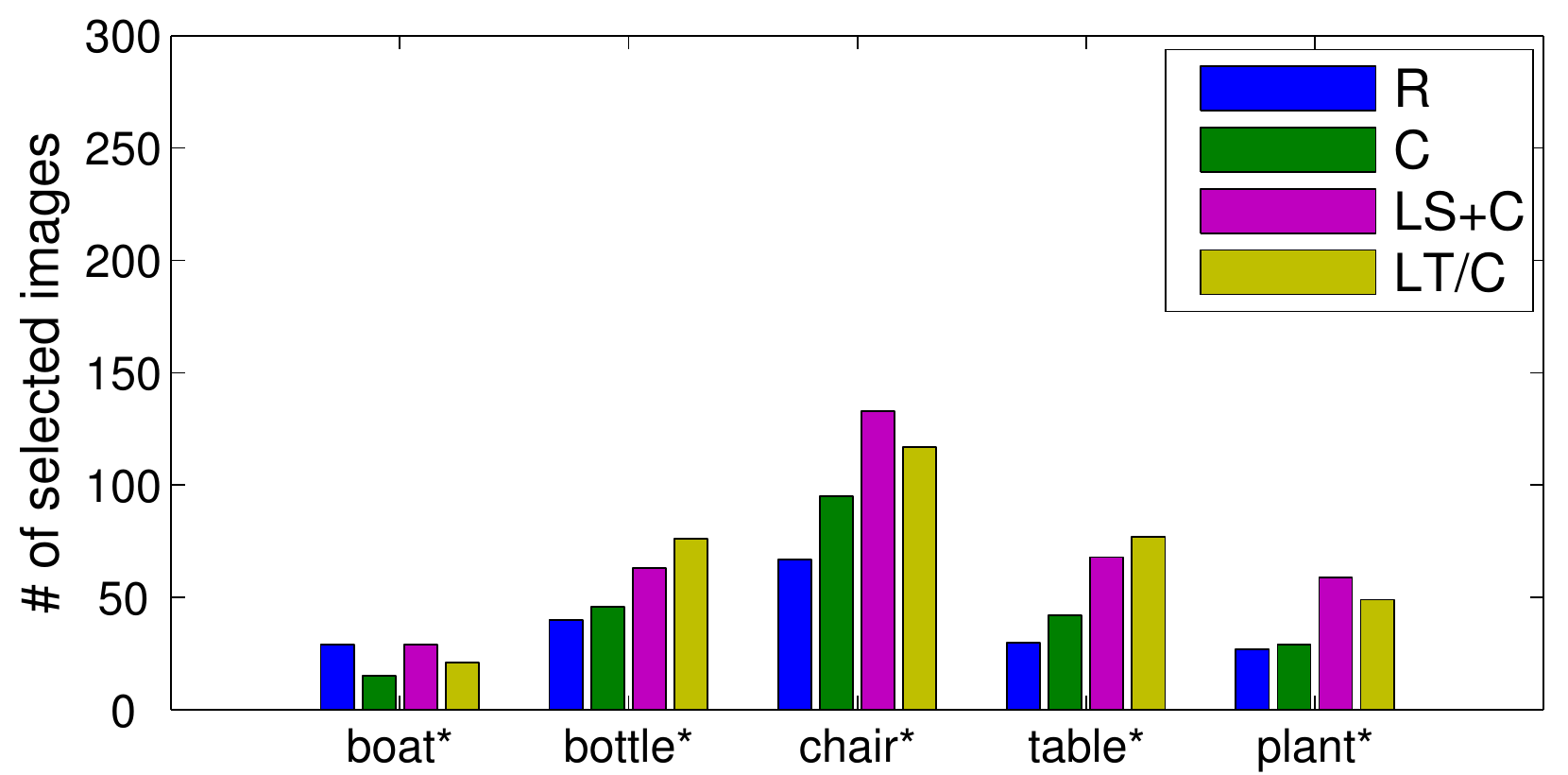}
    \caption{The number of selected images that contain objects belong to difficult classes by different active learning methods.}
\label{fig:selection_bar_chart_exp62}
\end{figure}

\begin{figure*}[t]
    \centering
    \includegraphics[width=\linewidth]{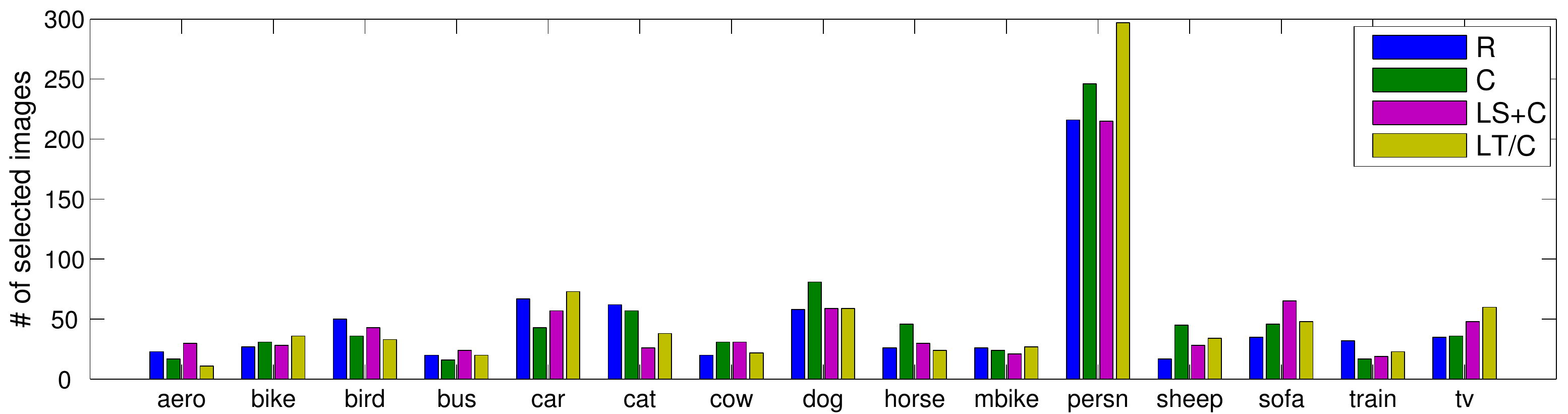}
    \caption{The number of selected images that contain objects belong to non-difficult classes by different active learning methods.}
\label{fig:selection_bar_chart_exp62_easy}
\end{figure*}

\begin{figure*}[t]
\begin{center}
   \includegraphics[width=0.95\linewidth]{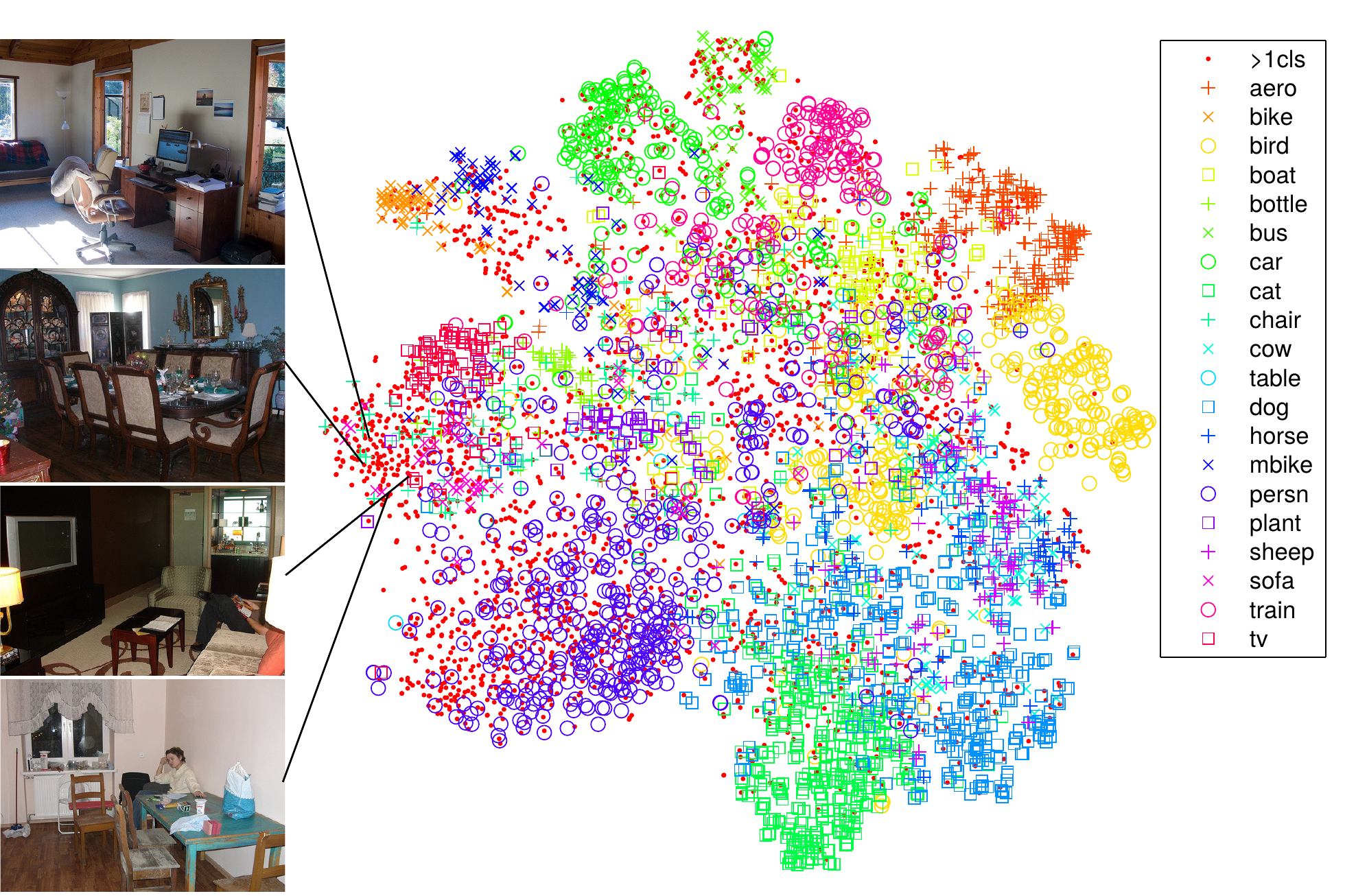}
\end{center}
   \caption{t-SNE embeddings of images on the PASCAL 2012 training set. VGG16 is used for generating high-dimensional vectors of images that used for the embedding. Each data point in the scatter plot is an image. ``\textgreater{1cls}'' represents an image that has objects belong to different classes. Images marked by only one class means that all the objects in the image belong to the same class. Images on the left are examples contain objects belong to difficult classes. As defined in Table~\ref{tab:PASCAL2012_1100imgs_exp62} ,the difficult classes are boat, bottle, chair, table, and plant.}
\label{fig:full_training_set_with_imgs}
\end{figure*}

\paragraph{Visualization of Different Active Learning Methods:}
We would like to visualize the selection process of different active learning methods. 
The experimental settings are the same with Sec.~4.1 in the main paper.
For the analysis and visualization in this section, we only use one trial instead of using the average of 5 trials for the easiness of reading.
The baseline FRCNN detector~\cite{Faster_RCNN_Ren2015} is trained on a training set of 500 labeled images, and then each active learning algorithm is executed for 3 rounds.
In each round, we select 200 images, add these images to the existing training set.
After 3 rounds, each method has selected 600 images for annotation, and a set with 1,100 labeled images is used to train the detector.

\begin{figure*}[t]
    \centering
    \begin{subfigure}[t]{0.49\textwidth} 
        \centering
        \includegraphics[width=\linewidth]{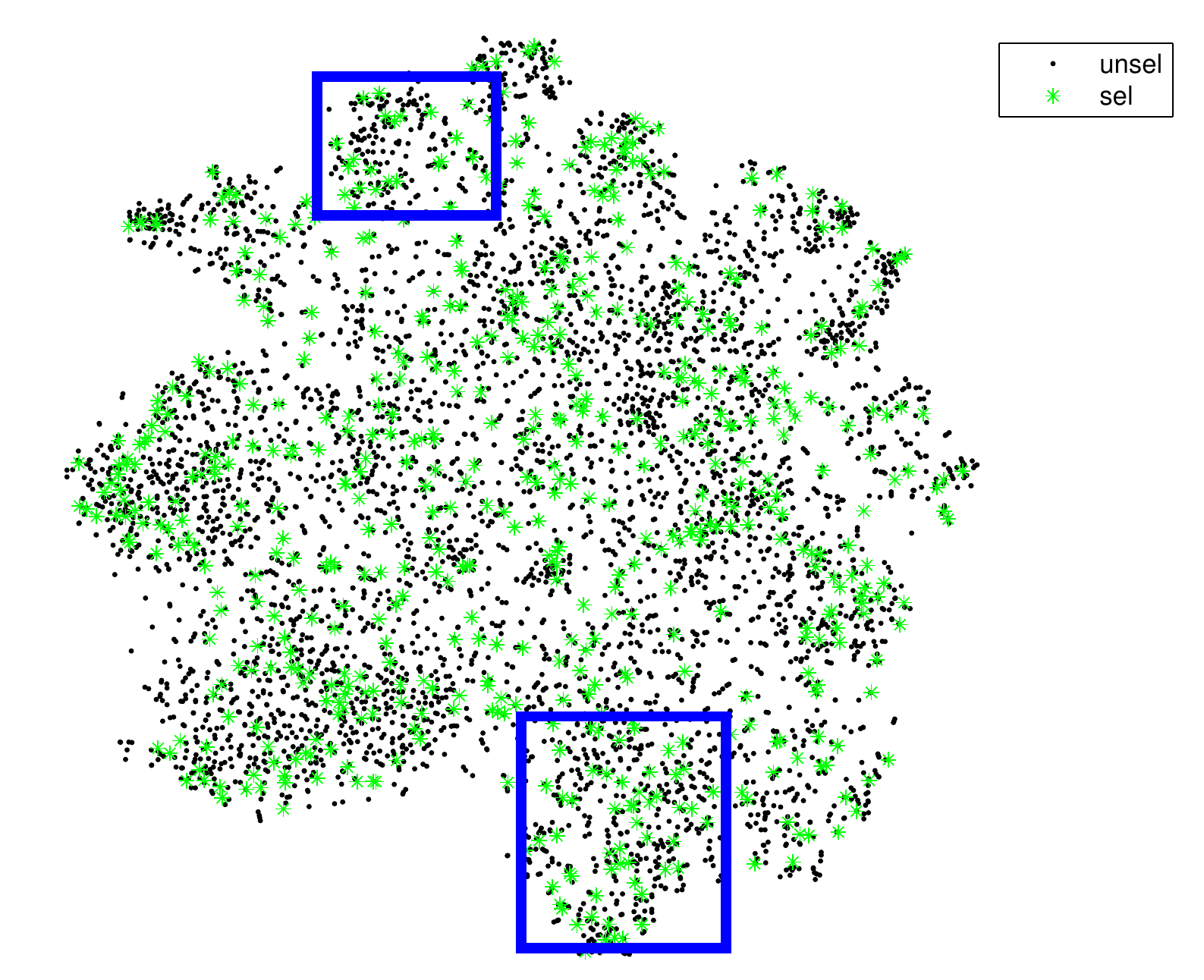}
        \caption{Random (R)}
        \label{fig:selection_R_binary}
    \end{subfigure}%
    ~
    \begin{subfigure}[t]{0.49\textwidth} 
        \centering
        \includegraphics[width=\linewidth]{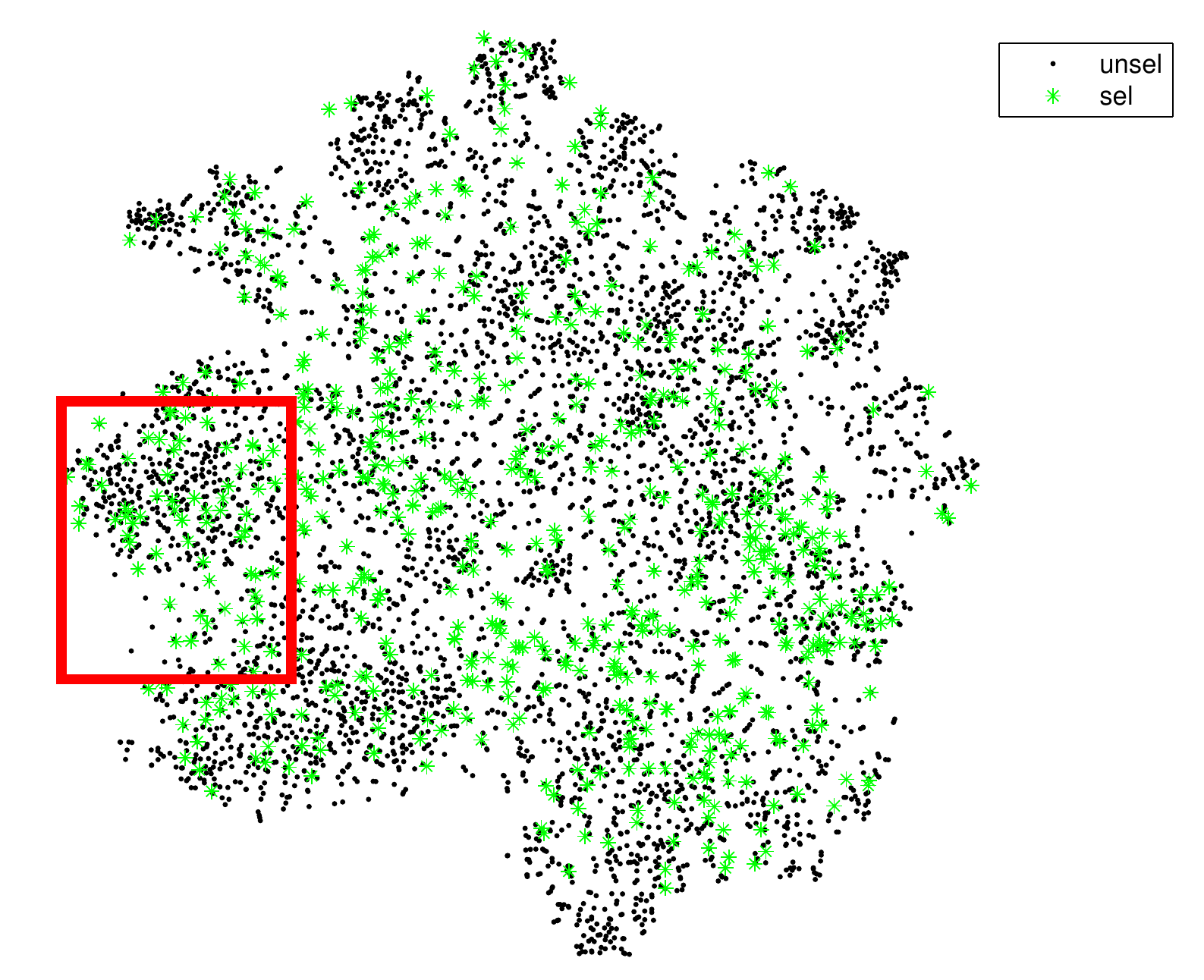}
        \caption{Classification (C)}
        \label{fig:selection_C_binary}
    \end{subfigure}
    \\
    \begin{subfigure}[t]{0.49\textwidth} 
        \centering
        \includegraphics[width=\linewidth]{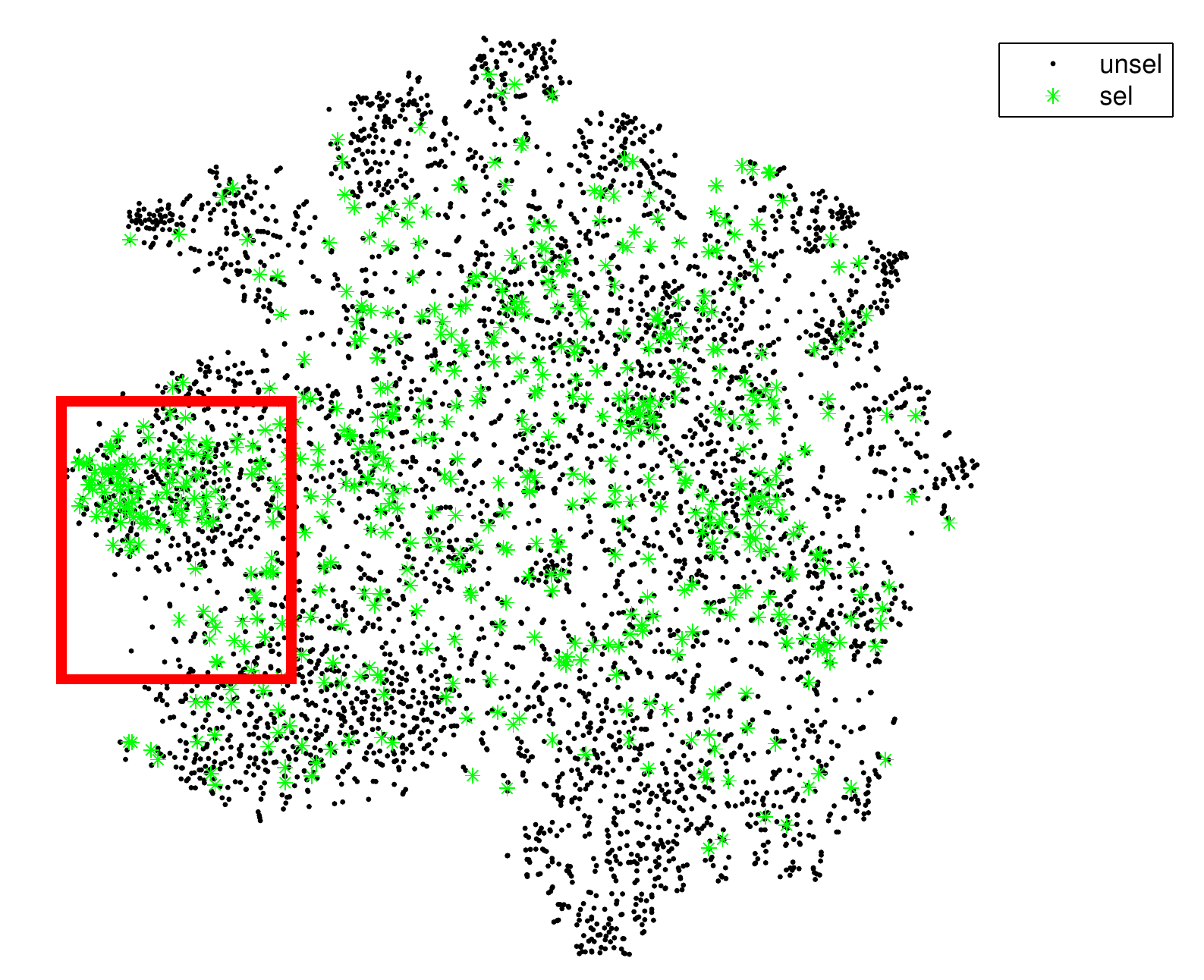}
        \caption{Localization stability + classification (LS+C)}
        \label{fig:selection_LS+C_binary}
    \end{subfigure}    
    ~
    \begin{subfigure}[t]{0.49\textwidth} 
        \centering
        \includegraphics[width=\linewidth]{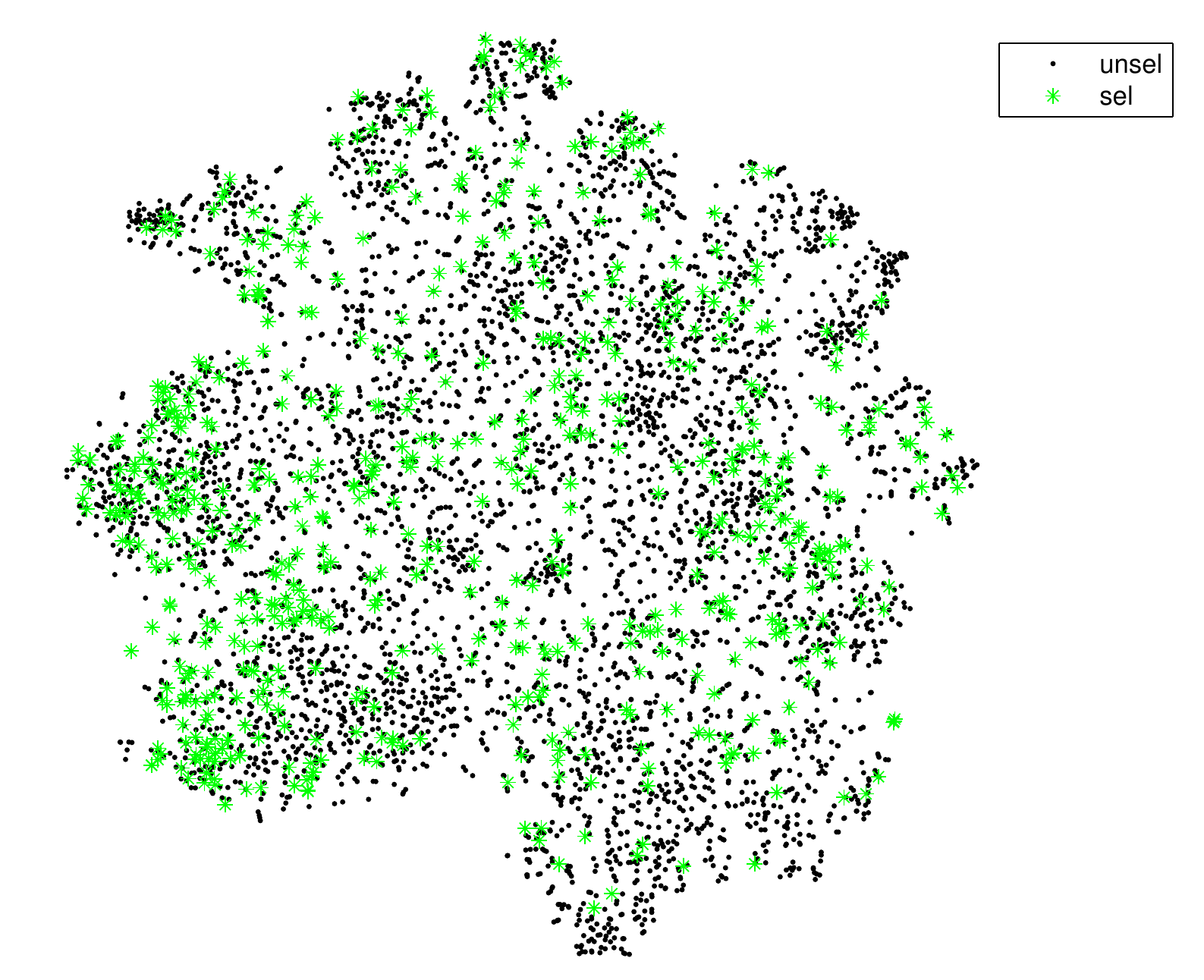}
        \caption{Localization tightness + classification (LT/C)}
        \label{fig:selection_LT+C_binary}
    \end{subfigure}
    \caption{The visualization of selection results by different active learning methods. Green asterisks (sel) are the images selected for annotation by each active learning method, and black dots (unsel) are the images that have not been selected. A detailed version of this graph with class-wise information is shown in Fig.~\ref{fig:selection_all_methods}.}
\label{fig:selection_all_methods_binary}
\end{figure*}

\begin{figure*}[t]
    \centering
    \begin{subfigure}[t]{0.24\textwidth} 
        \centering
        \includegraphics[width=\linewidth]{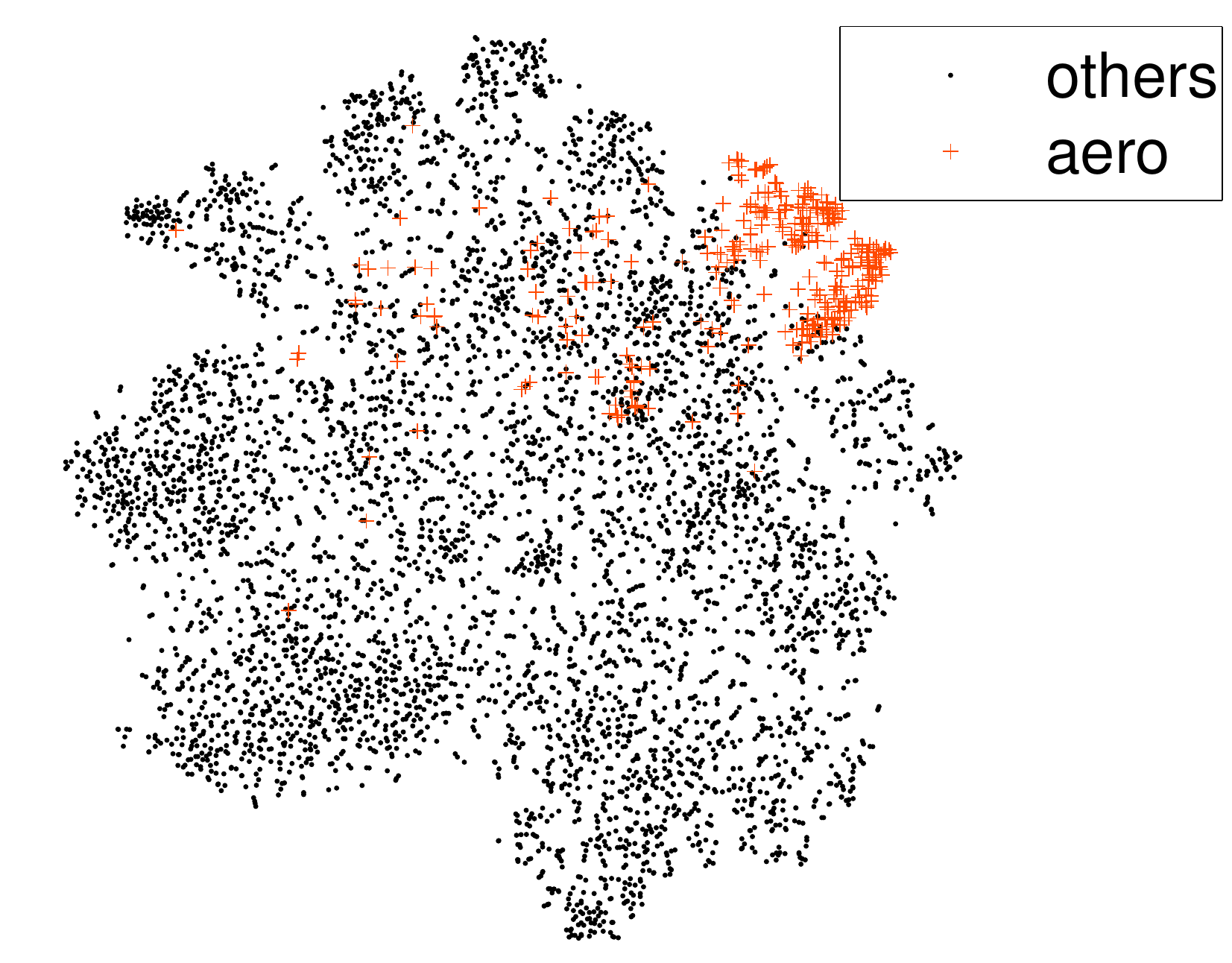}
        \caption{Aeroplane}
        \label{fig:training_aero}
    \end{subfigure}%
    ~
    \begin{subfigure}[t]{0.24\textwidth} 
        \centering
        \includegraphics[width=\linewidth]{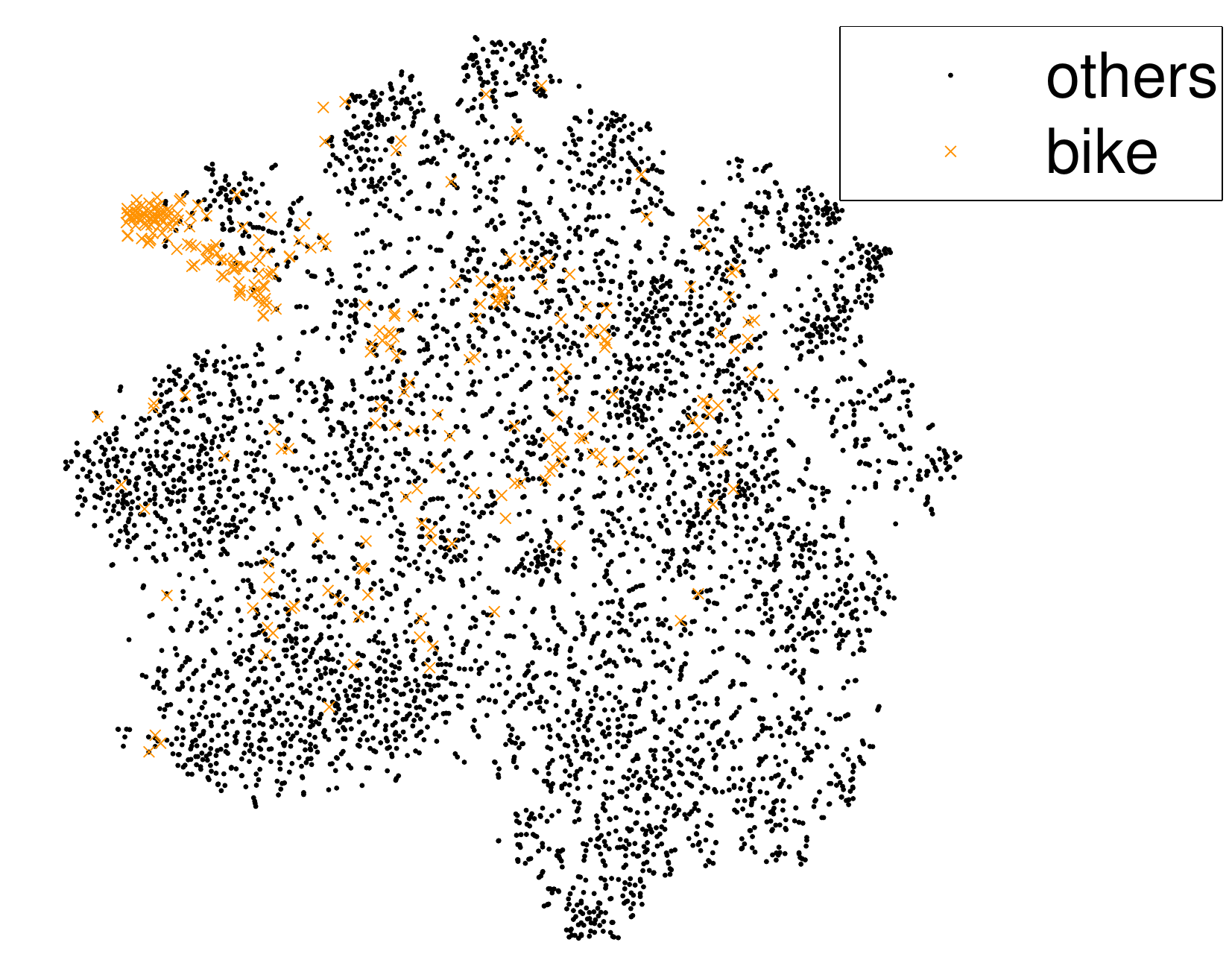}
        \caption{Bicycle}
        \label{fig:training_bike}
    \end{subfigure}
    ~
    \begin{subfigure}[t]{0.24\textwidth} 
        \centering
        \includegraphics[width=\linewidth]{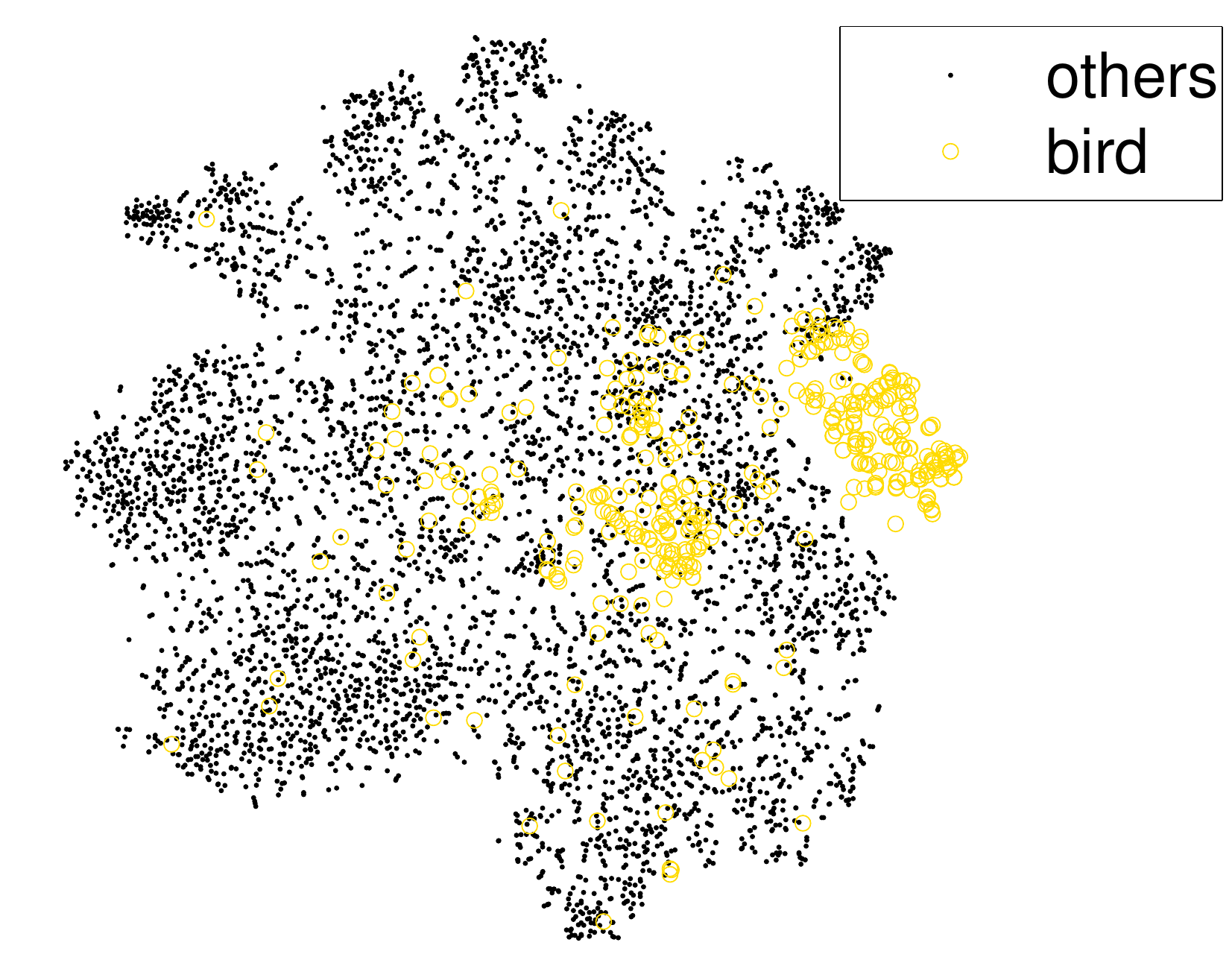}
        \caption{Bird}
        \label{fig:training_bird}
    \end{subfigure}
    ~
    \begin{subfigure}[t]{0.24\textwidth} 
        \centering
        \includegraphics[width=\linewidth]{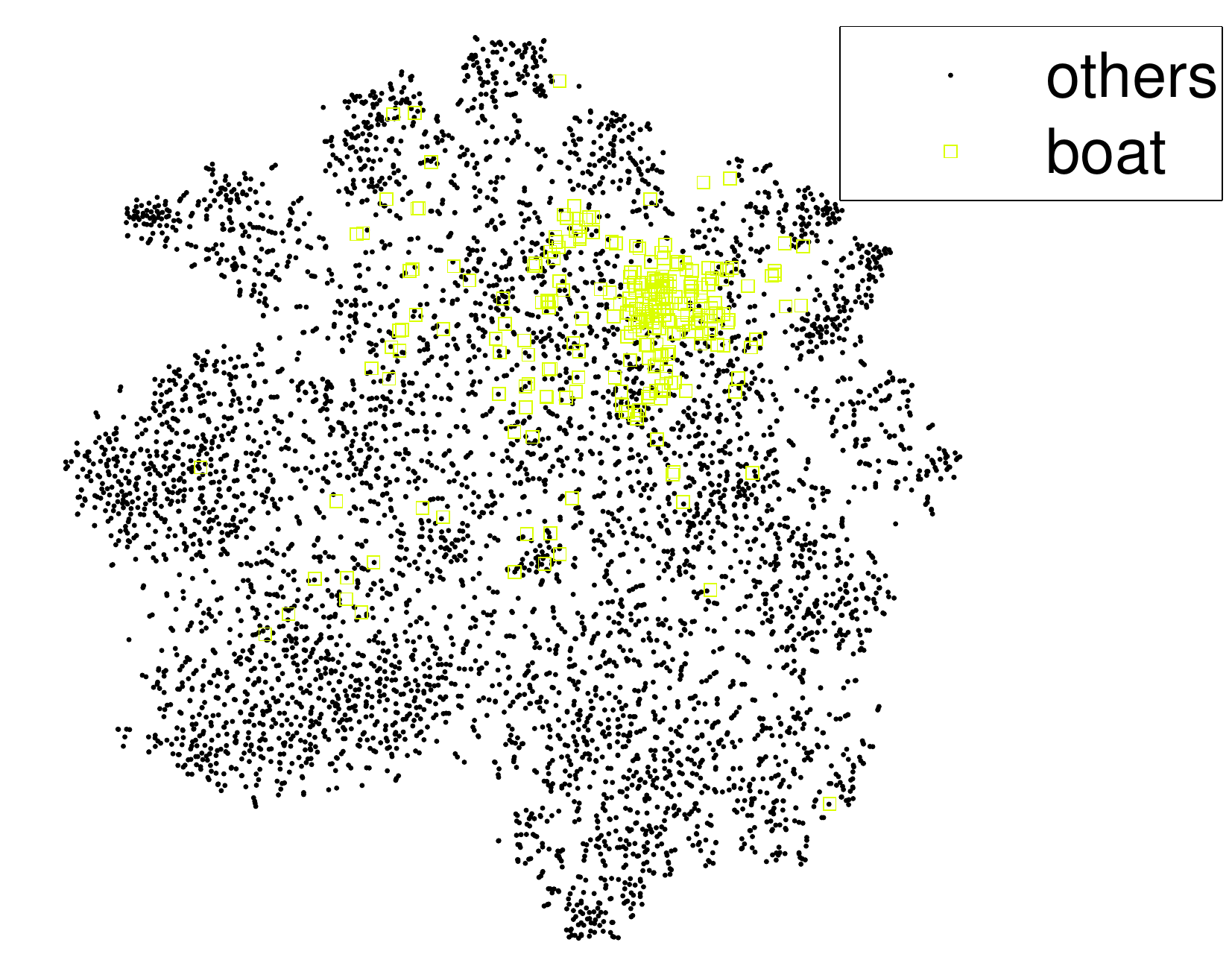}
        \caption{Boat}
        \label{fig:training_boat}
    \end{subfigure}
    \\        
        \begin{subfigure}[t]{0.24\textwidth} 
        \centering
        \includegraphics[width=\linewidth]{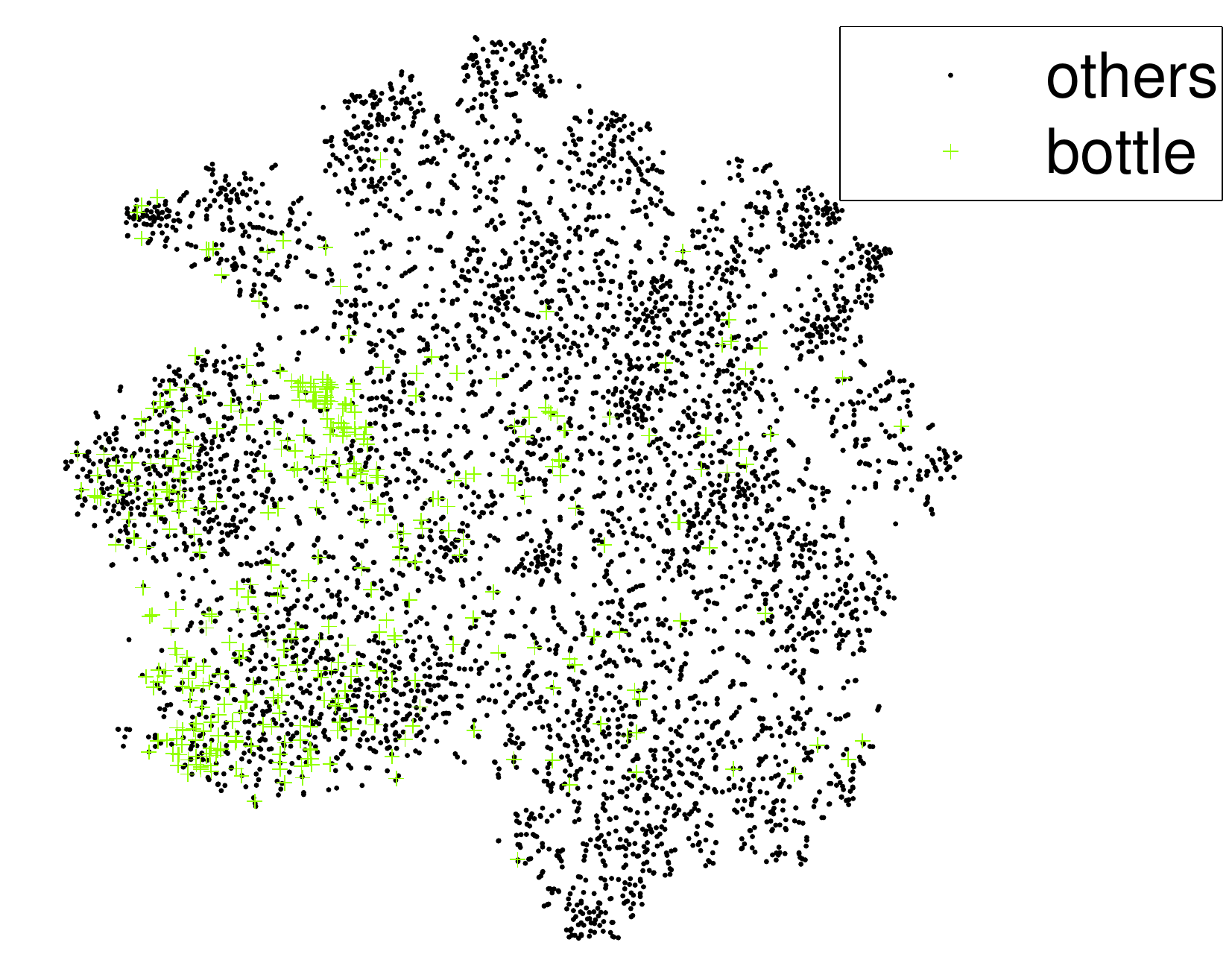}
        \caption{Bottle}
        \label{fig:training_bottle}
    \end{subfigure}%
    ~
    \begin{subfigure}[t]{0.24\textwidth} 
        \centering
        \includegraphics[width=\linewidth]{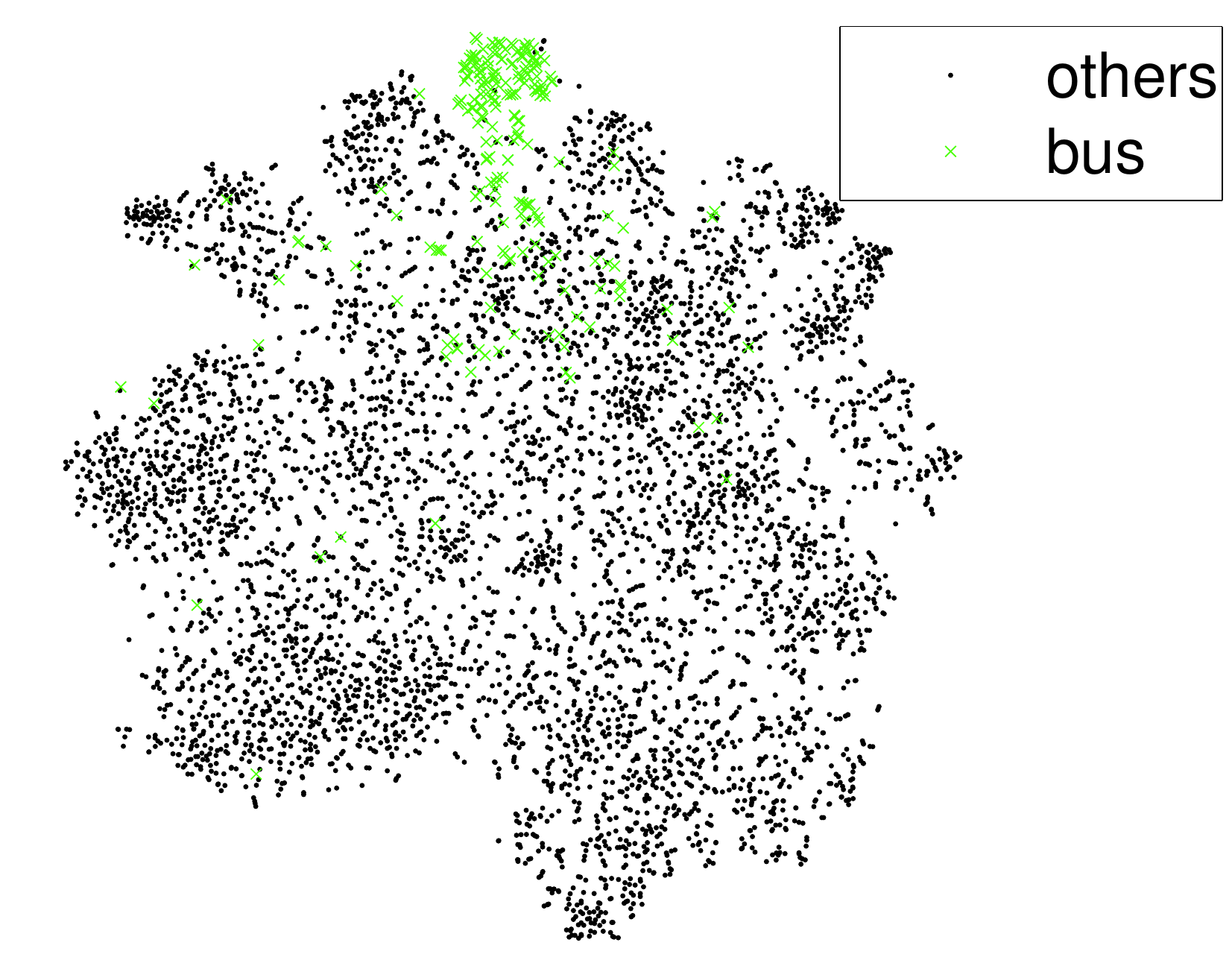}
        \caption{Bus}
        \label{fig:training_bus}
    \end{subfigure}
    ~
    \begin{subfigure}[t]{0.24\textwidth} 
        \centering
        \includegraphics[width=\linewidth]{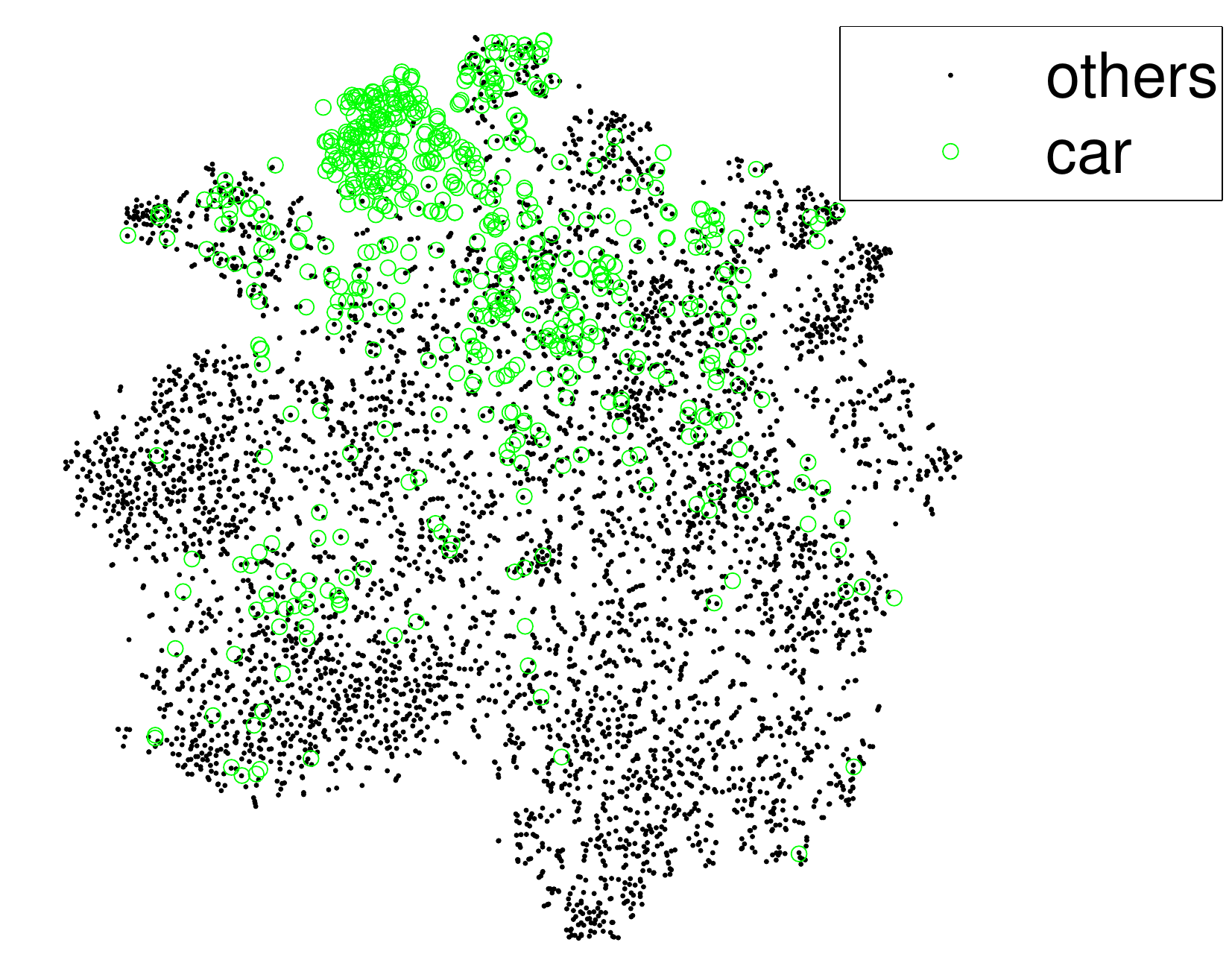}
        \caption{Car}
        \label{fig:training_car}
    \end{subfigure}
    ~
    \begin{subfigure}[t]{0.24\textwidth} 
        \centering
        \includegraphics[width=\linewidth]{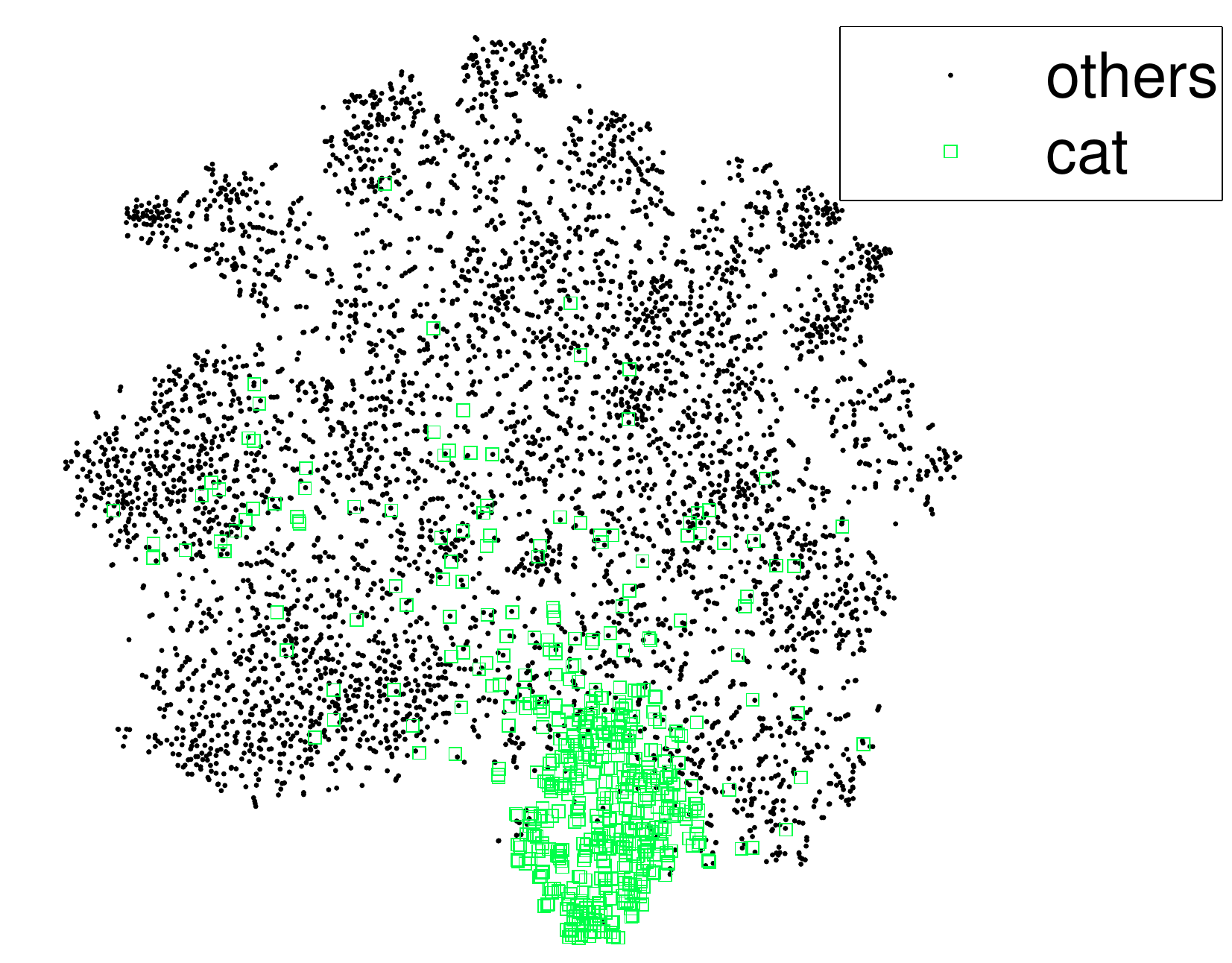}
        \caption{Cat}
        \label{fig:training_cat}
    \end{subfigure}
    \\        
        \begin{subfigure}[t]{0.24\textwidth} 
        \centering
        \includegraphics[width=\linewidth]{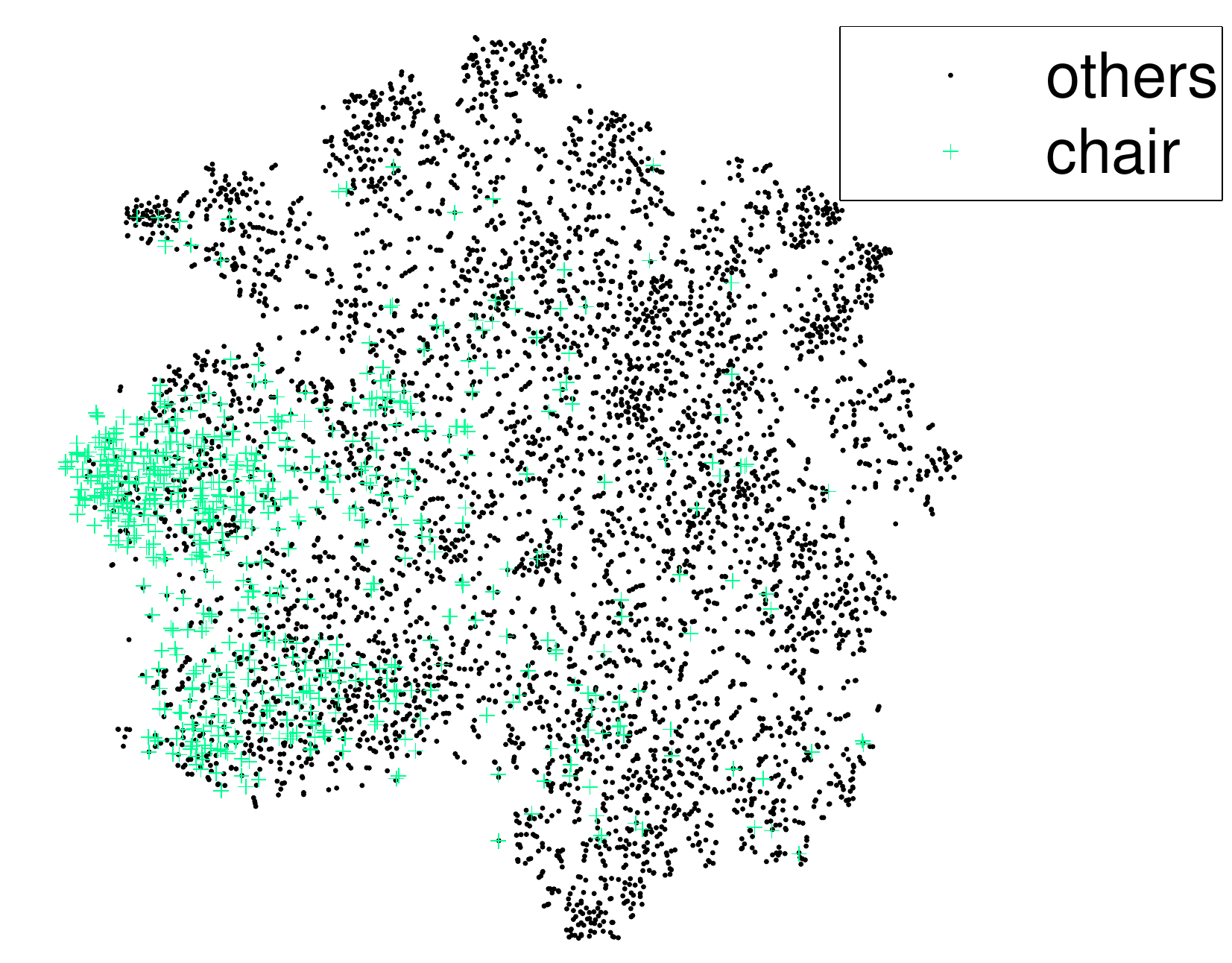}
        \caption{Chair}
        \label{fig:training_chair}
    \end{subfigure}%
    ~
    \begin{subfigure}[t]{0.24\textwidth} 
        \centering
        \includegraphics[width=\linewidth]{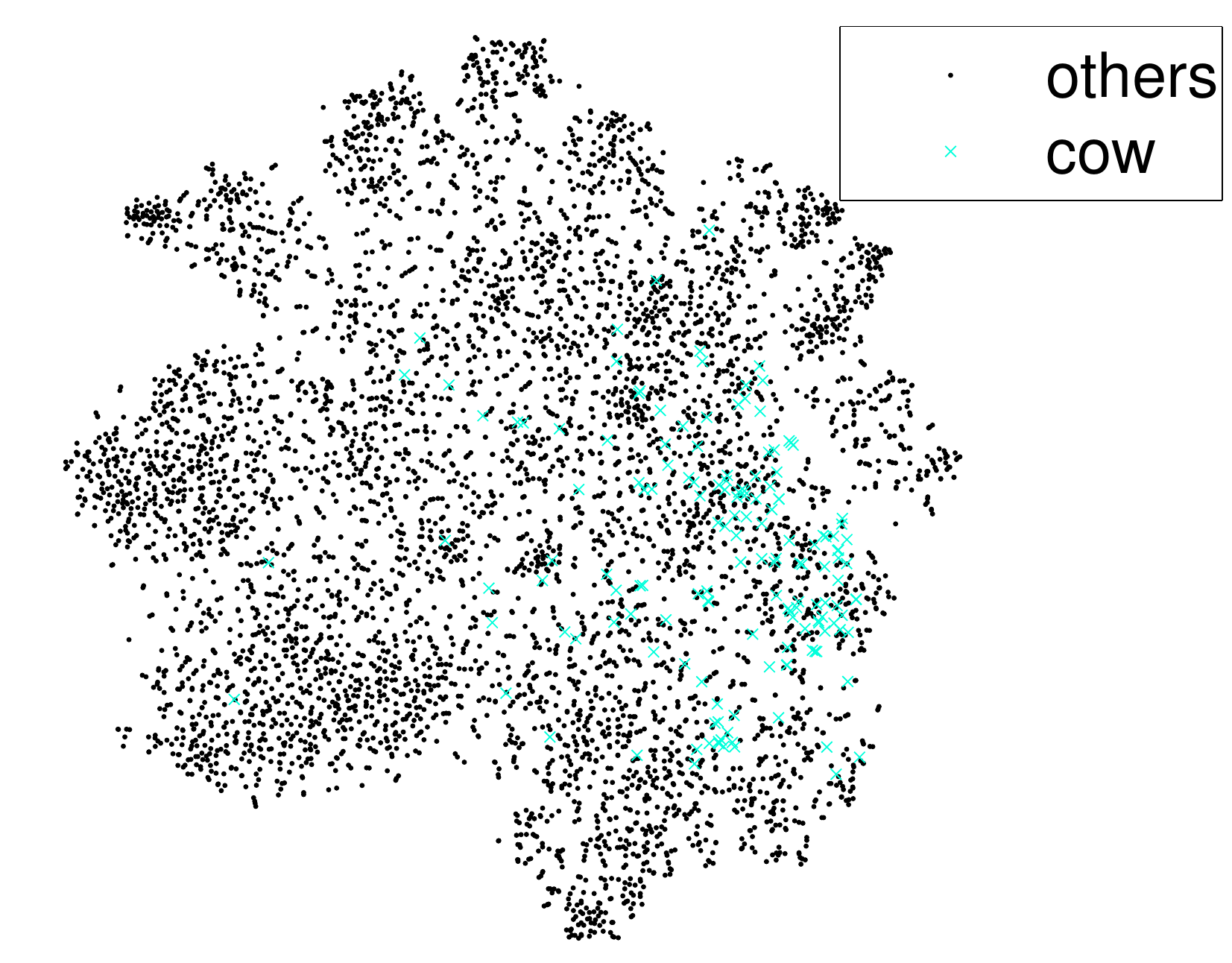}
        \caption{Cow}
        \label{fig:training_cow}
    \end{subfigure}
    ~
    \begin{subfigure}[t]{0.24\textwidth} 
        \centering
        \includegraphics[width=\linewidth]{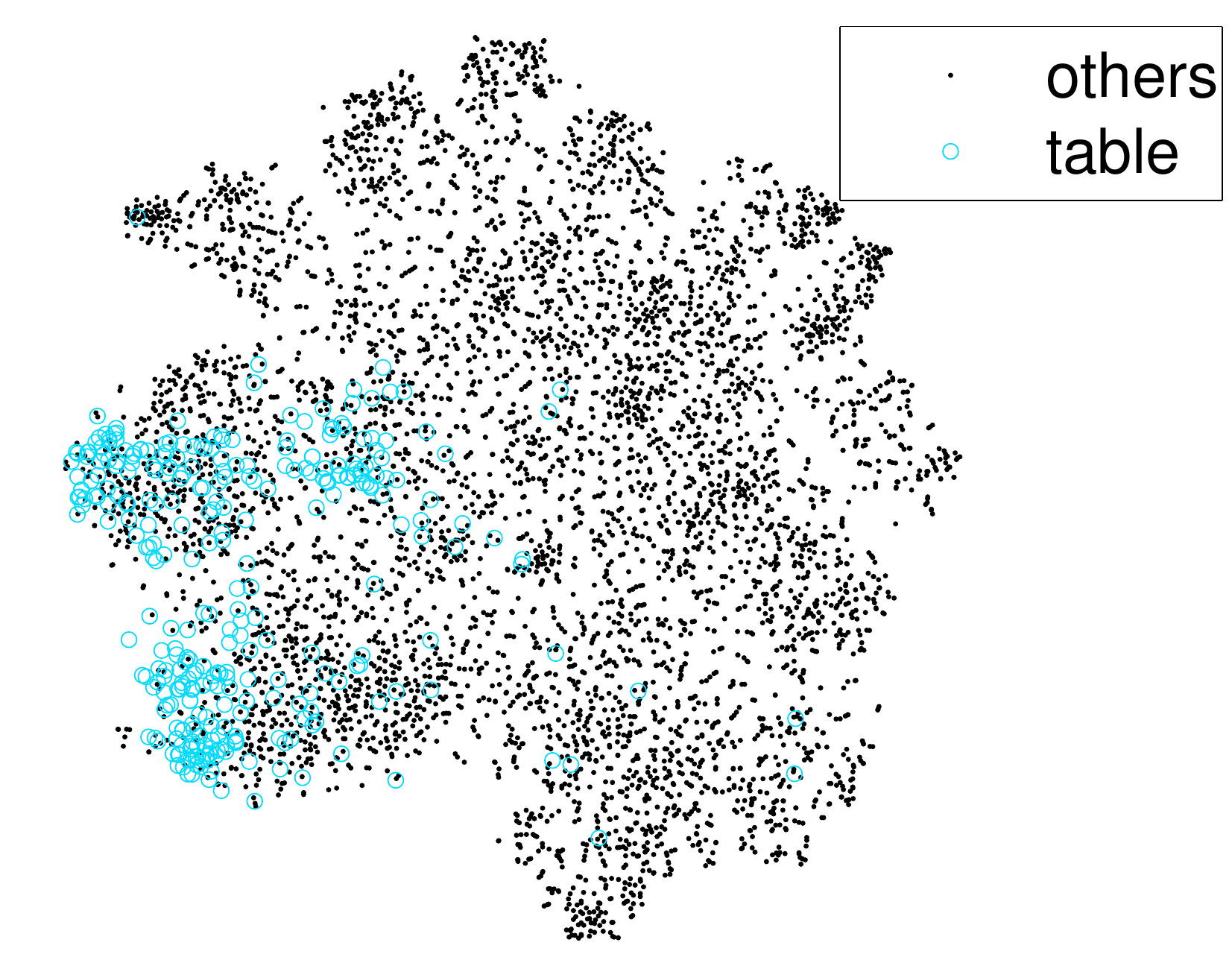}
        \caption{Diningtable}
        \label{fig:training_table}
    \end{subfigure}
    ~
    \begin{subfigure}[t]{0.24\textwidth} 
        \centering
        \includegraphics[width=\linewidth]{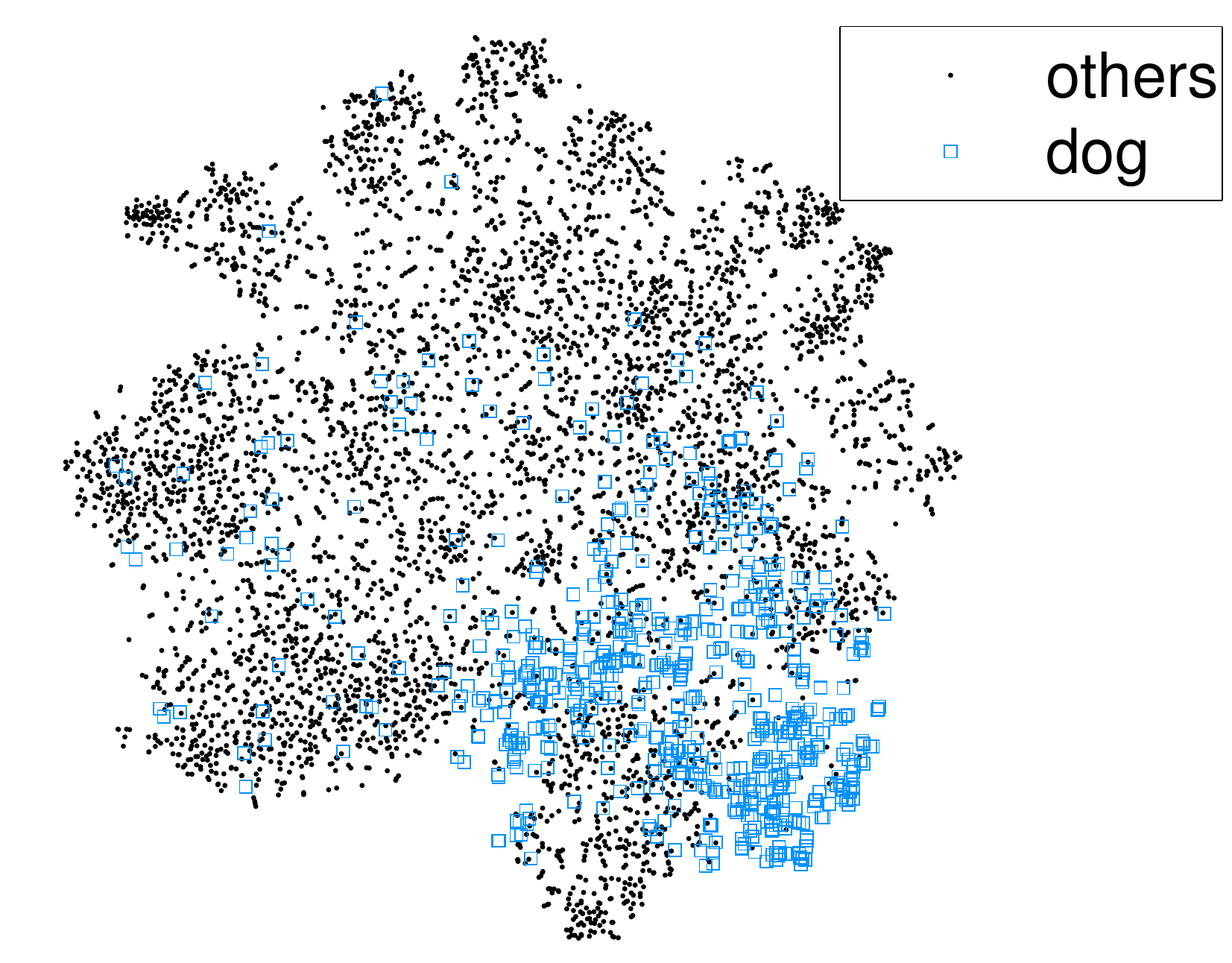}
        \caption{Dog}
        \label{fig:training_dog}
    \end{subfigure}
    \\        
        \begin{subfigure}[t]{0.24\textwidth} 
        \centering
        \includegraphics[width=\linewidth]{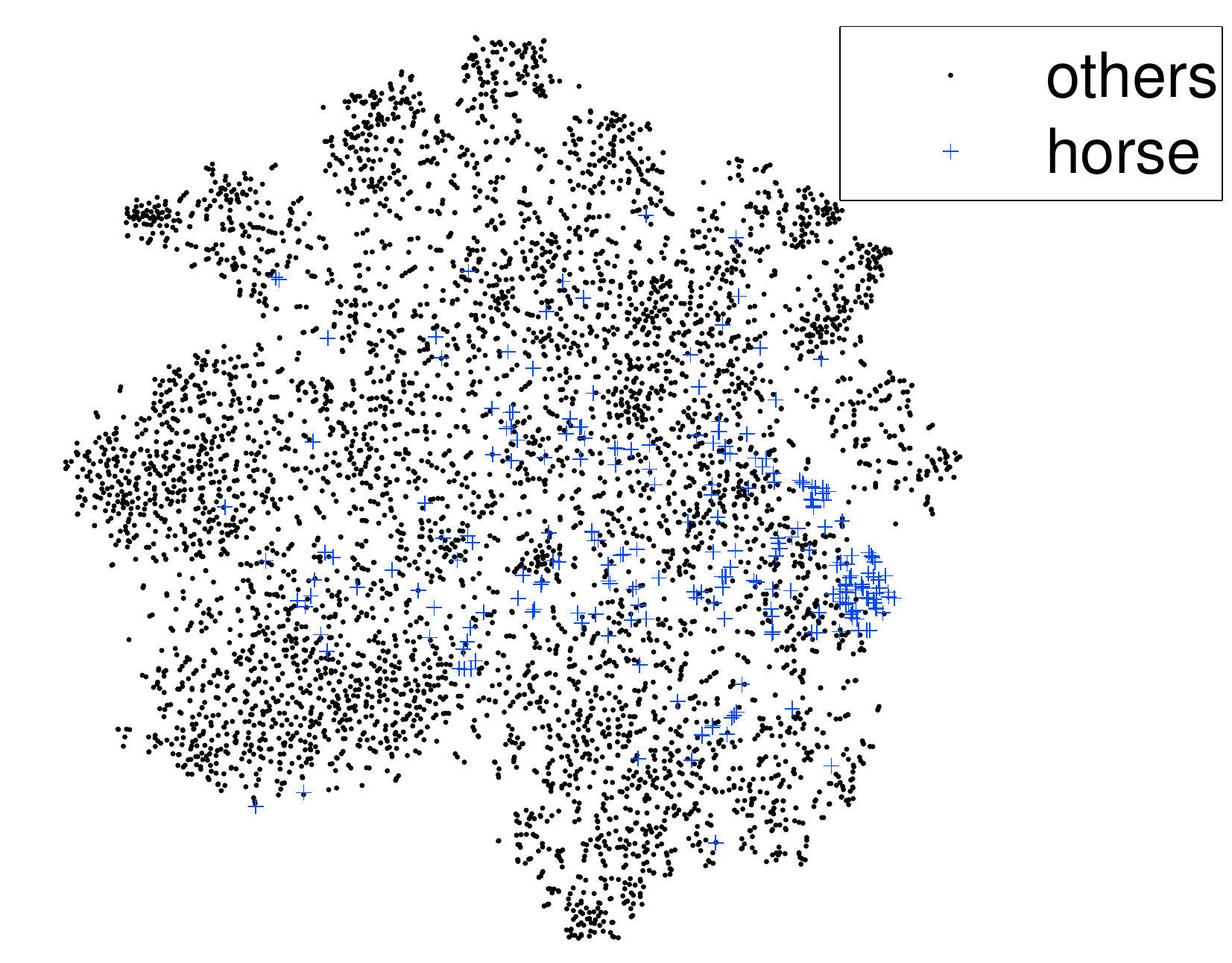}
        \caption{Horse}
        \label{fig:training_horse}
    \end{subfigure}%
    ~
    \begin{subfigure}[t]{0.24\textwidth} 
        \centering
        \includegraphics[width=\linewidth]{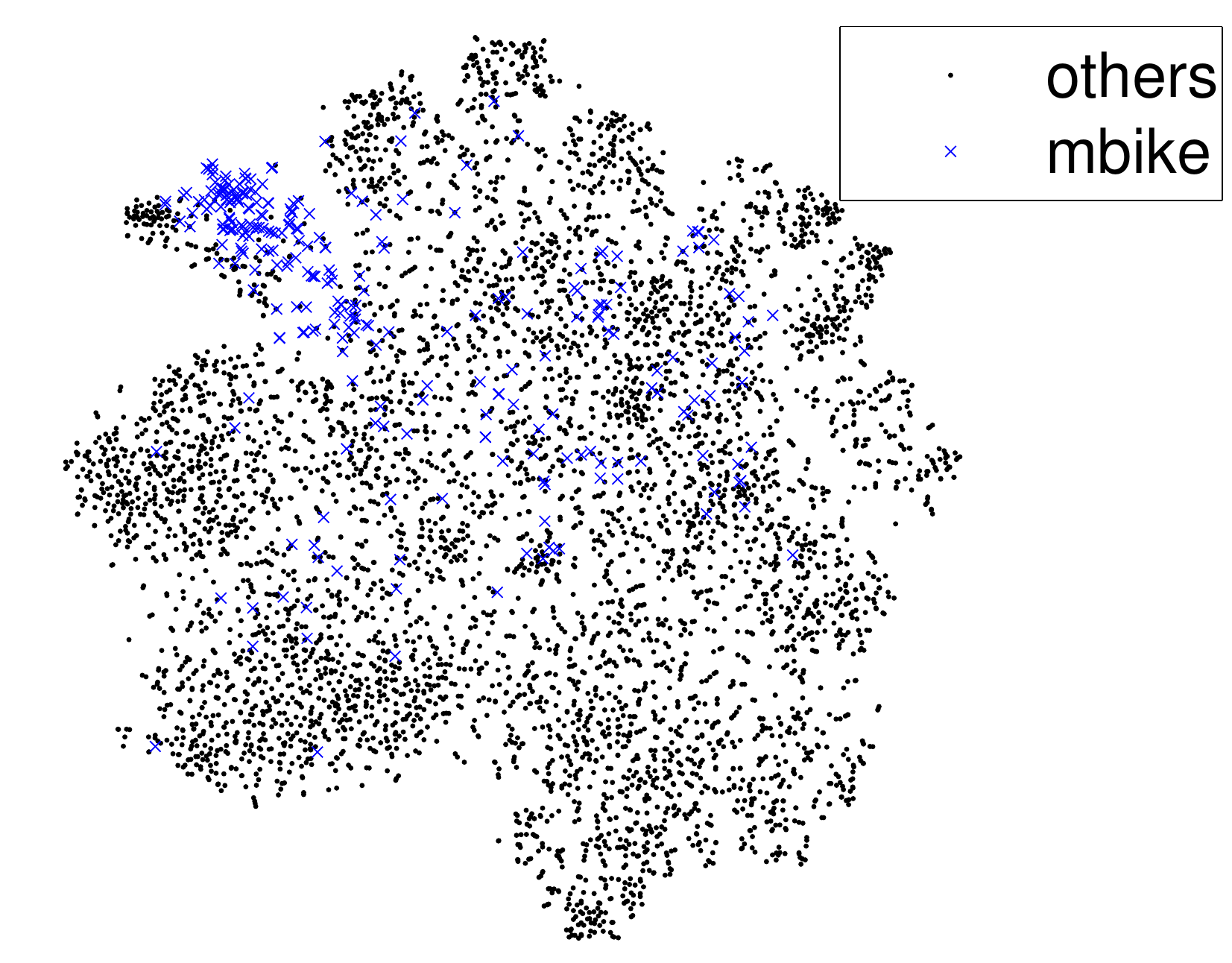}
        \caption{Motorbike}
        \label{fig:training_mbike}
    \end{subfigure}
    ~
    \begin{subfigure}[t]{0.24\textwidth} 
        \centering
        \includegraphics[width=\linewidth]{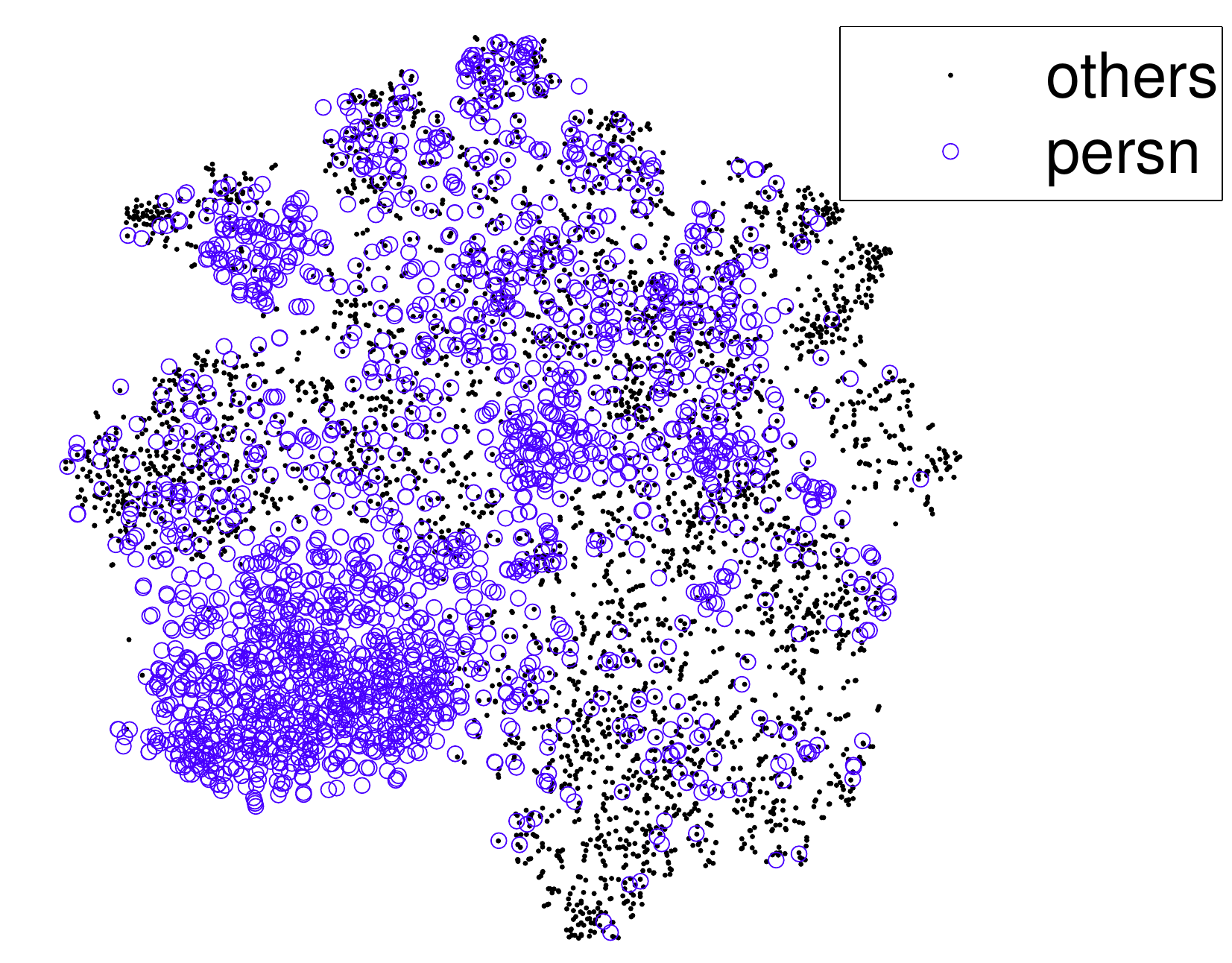}
        \caption{Person}
        \label{fig:training_persn}
    \end{subfigure}
    ~
    \begin{subfigure}[t]{0.24\textwidth} 
        \centering
        \includegraphics[width=\linewidth]{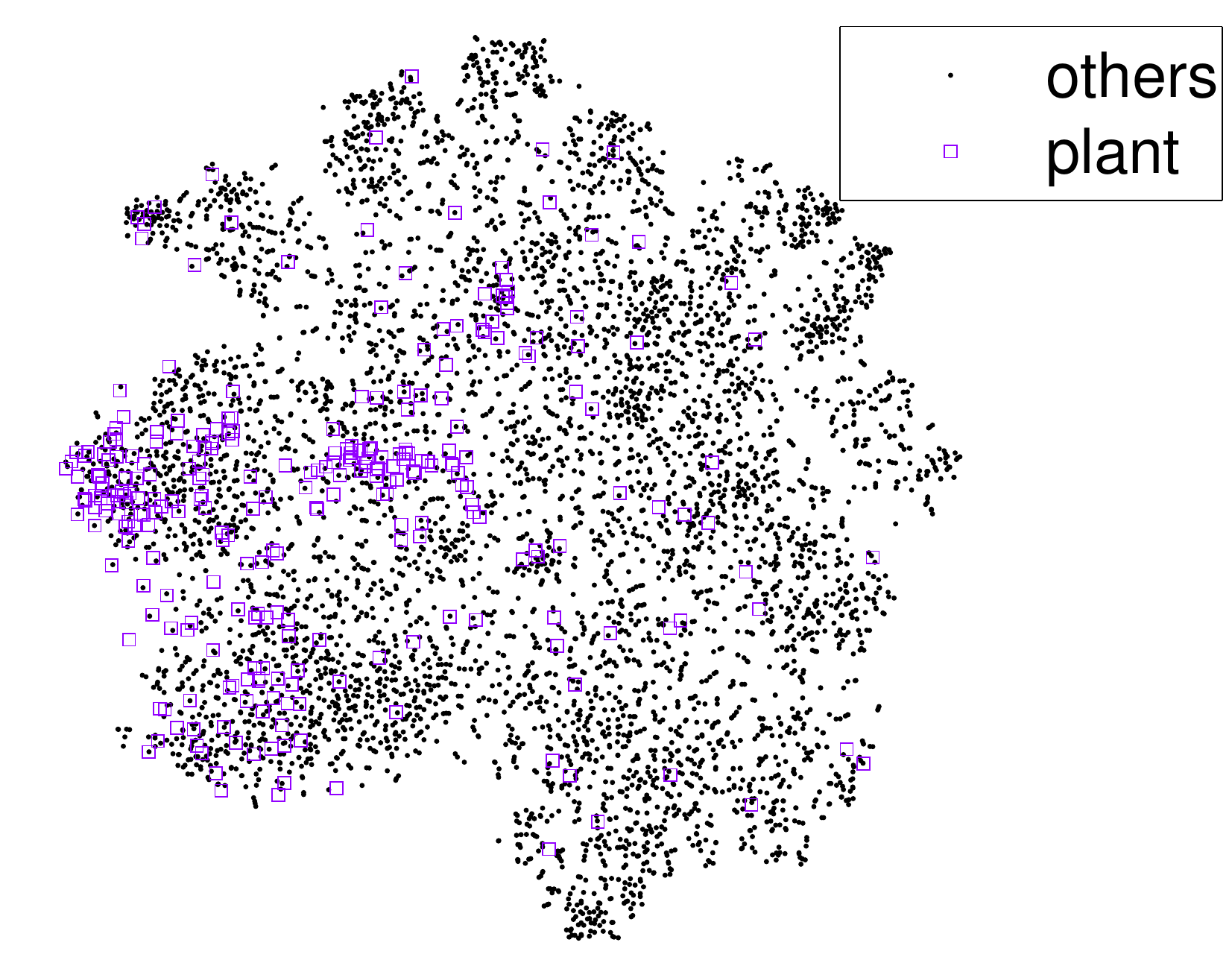}
        \caption{Pottedplant}
        \label{fig:training_pottedplant}
    \end{subfigure}
    \\        
        \begin{subfigure}[t]{0.24\textwidth} 
        \centering
        \includegraphics[width=\linewidth]{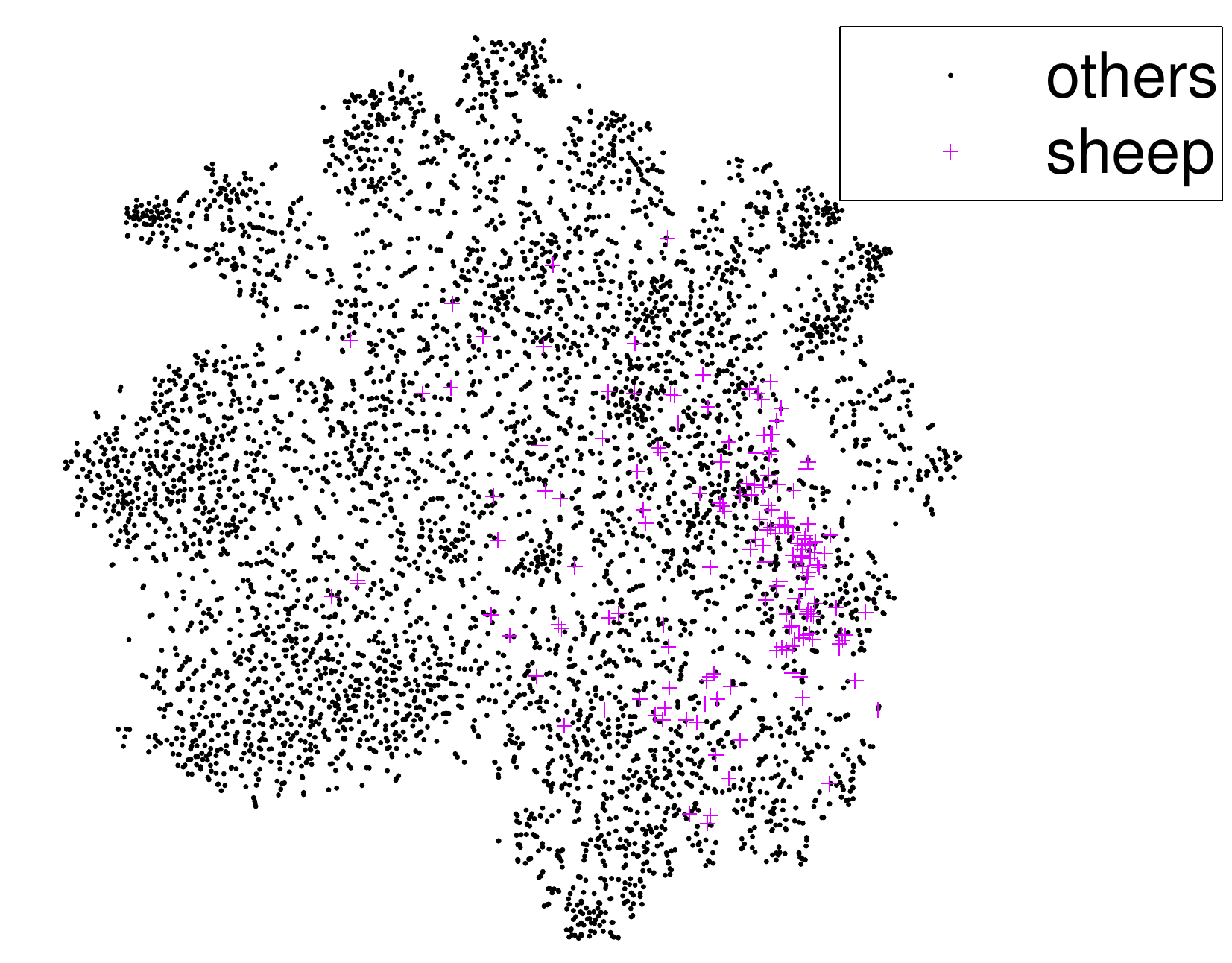}
        \caption{Sheep}
        \label{fig:training_sheep}
    \end{subfigure}%
    ~
    \begin{subfigure}[t]{0.24\textwidth} 
        \centering
        \includegraphics[width=\linewidth]{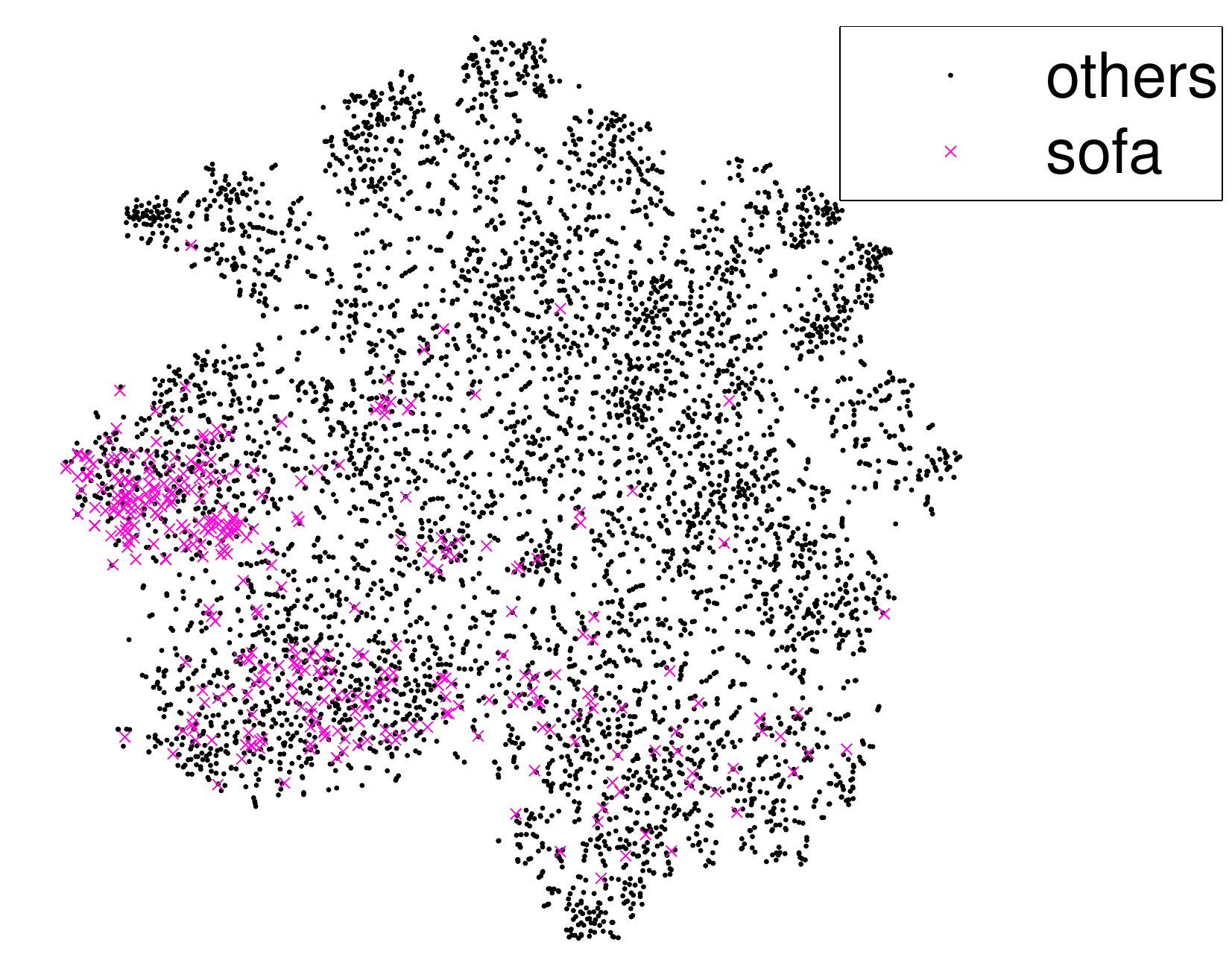}
        \caption{Sofa}
        \label{fig:training_sofa}
    \end{subfigure}
    ~
    \begin{subfigure}[t]{0.24\textwidth} 
        \centering
        \includegraphics[width=\linewidth]{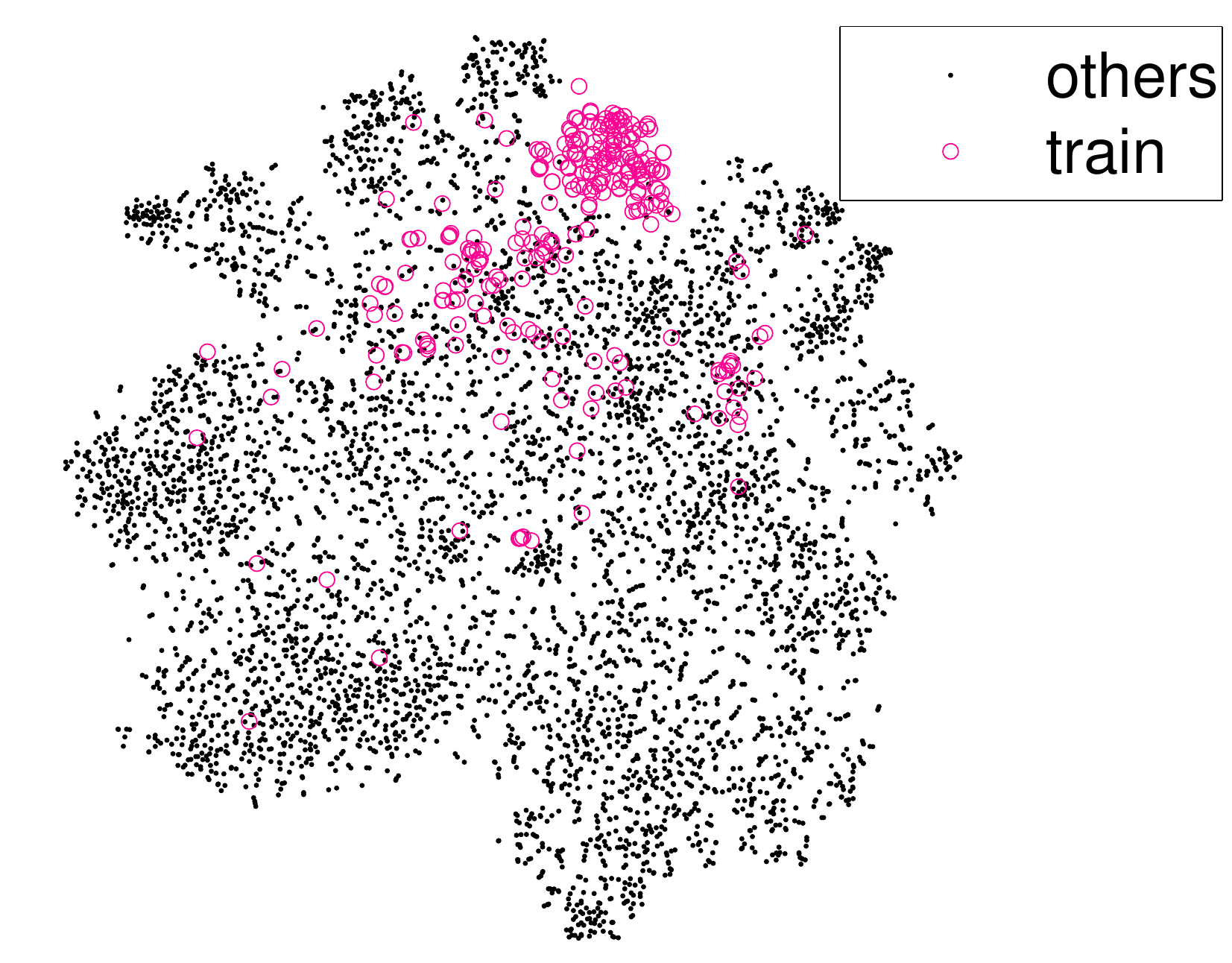}
        \caption{Train}
        \label{fig:training_train}
    \end{subfigure}
    ~
    \begin{subfigure}[t]{0.24\textwidth} 
        \centering
        \includegraphics[width=\linewidth]{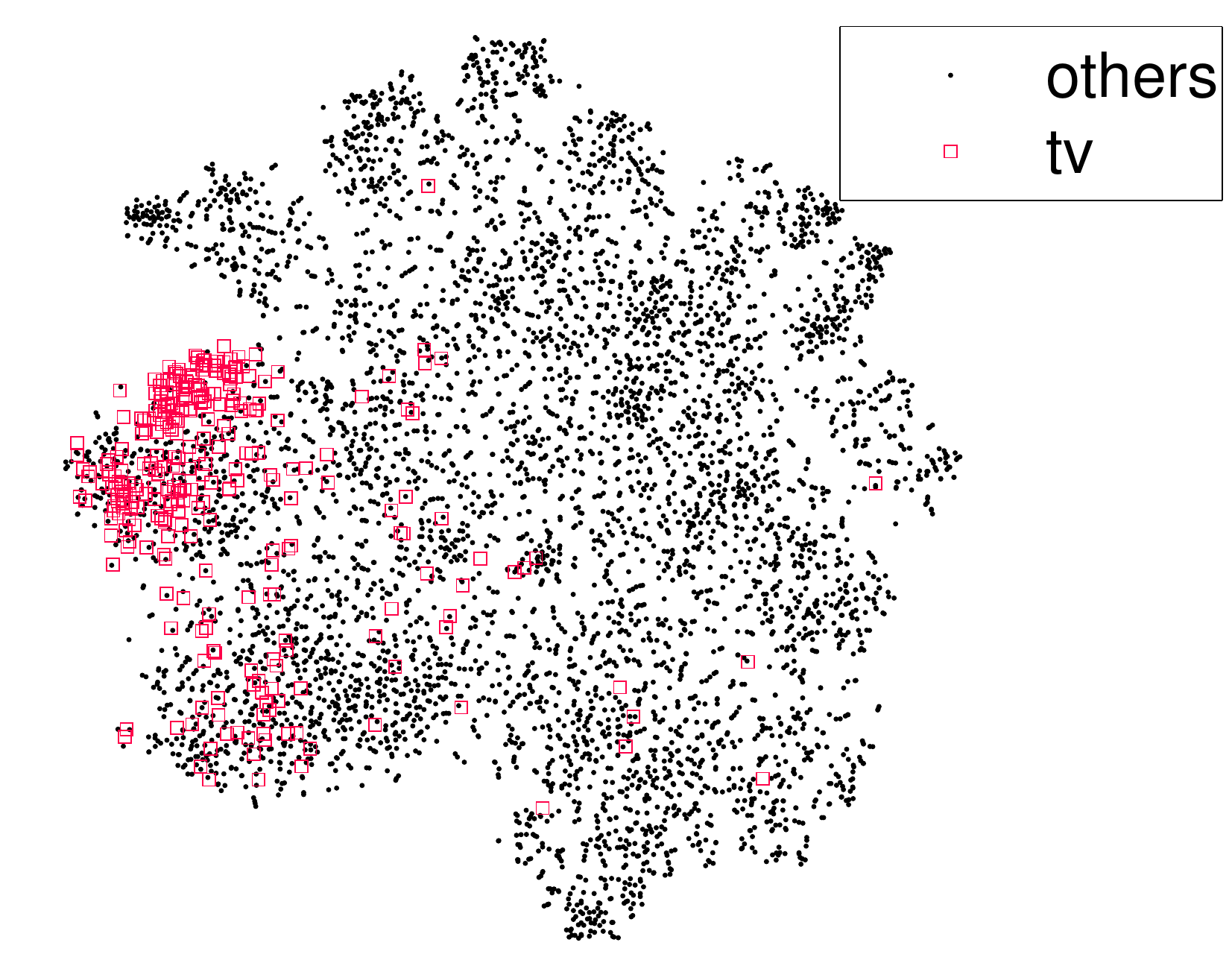}
        \caption{TV monitor}
        \label{fig:training_tvmonitor}
    \end{subfigure}
    \caption{t-SNE embeddings of images for each category on the PASCAL 2012 training set. Different from Fig.~\ref{fig:full_training_set_with_imgs}, each colored point in the graphs represents an image that includes at least one object belongs to the target class. For example, each orange plus sign in (a) represents an image which has at least one aeroplane in it.
    }
\label{fig:full_training_set_classwise}
\end{figure*}

Table~\ref{tab:PASCAL2012_1100imgs_exp62} shows the average precision for each method on the PSACAL 2012 validation set after 3 rounds of active learning.
As defined in the main paper, catergories with AP lower than 40\% in passive learning (R) are defined as difficult categories.
These difficult classes are marked by an asterisk in Table~\ref{tab:PASCAL2012_1100imgs_exp62}.
We further analyze the selection result of different methods by a visualization as shown in Fig.~\ref{fig:selection_all_methods_binary}.
There are total 5,217 images (500 images in the initial training set of this trial are not included) in each graph.
600 images selected for annotation by each active learning method are represented by green asterisks, and the rest 4,617 images that have not been chosen are represented by black dots.

\begin{figure*}[t]
    \centering
    \begin{subfigure}[t]{0.49\textwidth} 
        \centering
        \includegraphics[width=\linewidth]{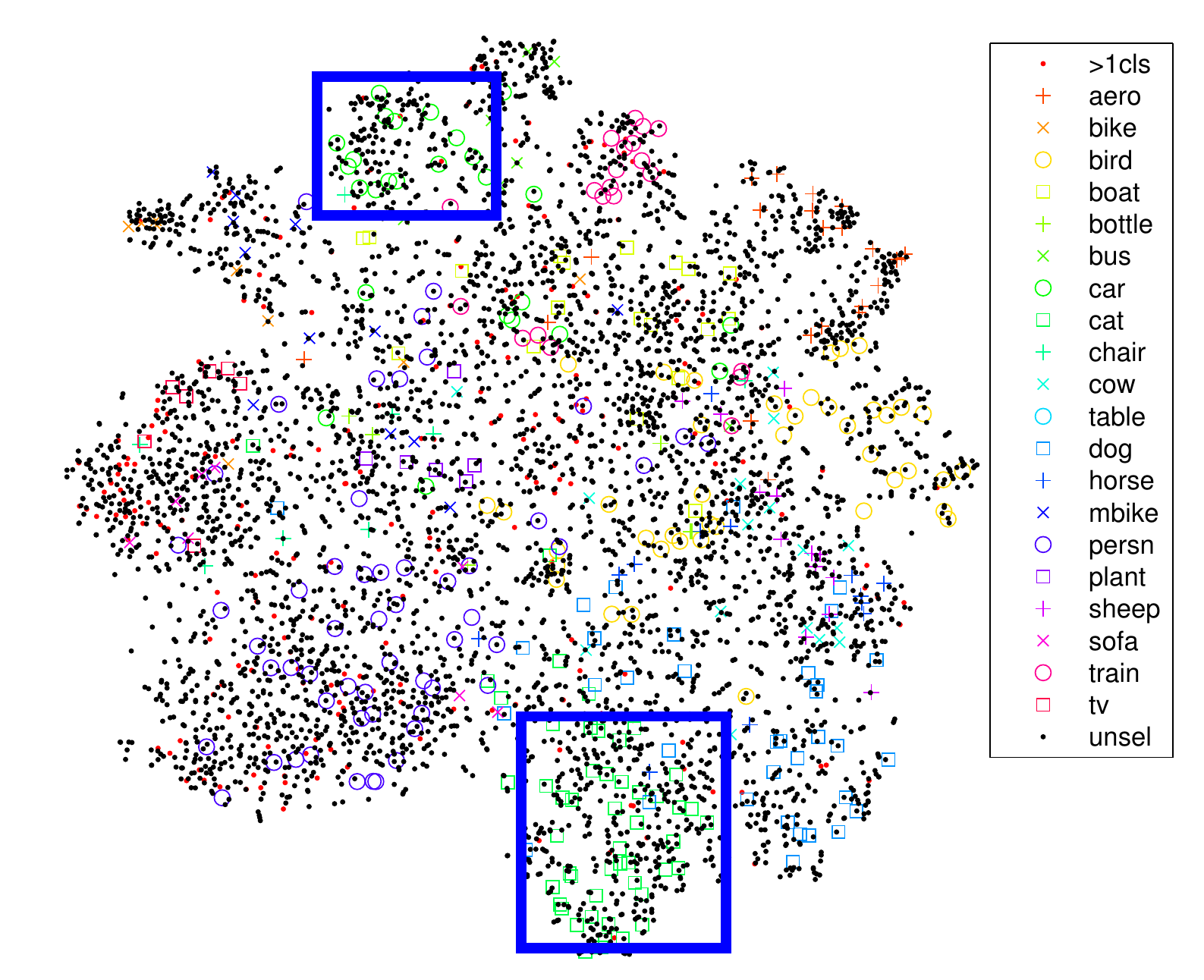}
        \caption{Random (R)}
        \label{fig:selection_R}
    \end{subfigure}%
    ~
    \begin{subfigure}[t]{0.49\textwidth} 
        \centering
        \includegraphics[width=\linewidth]{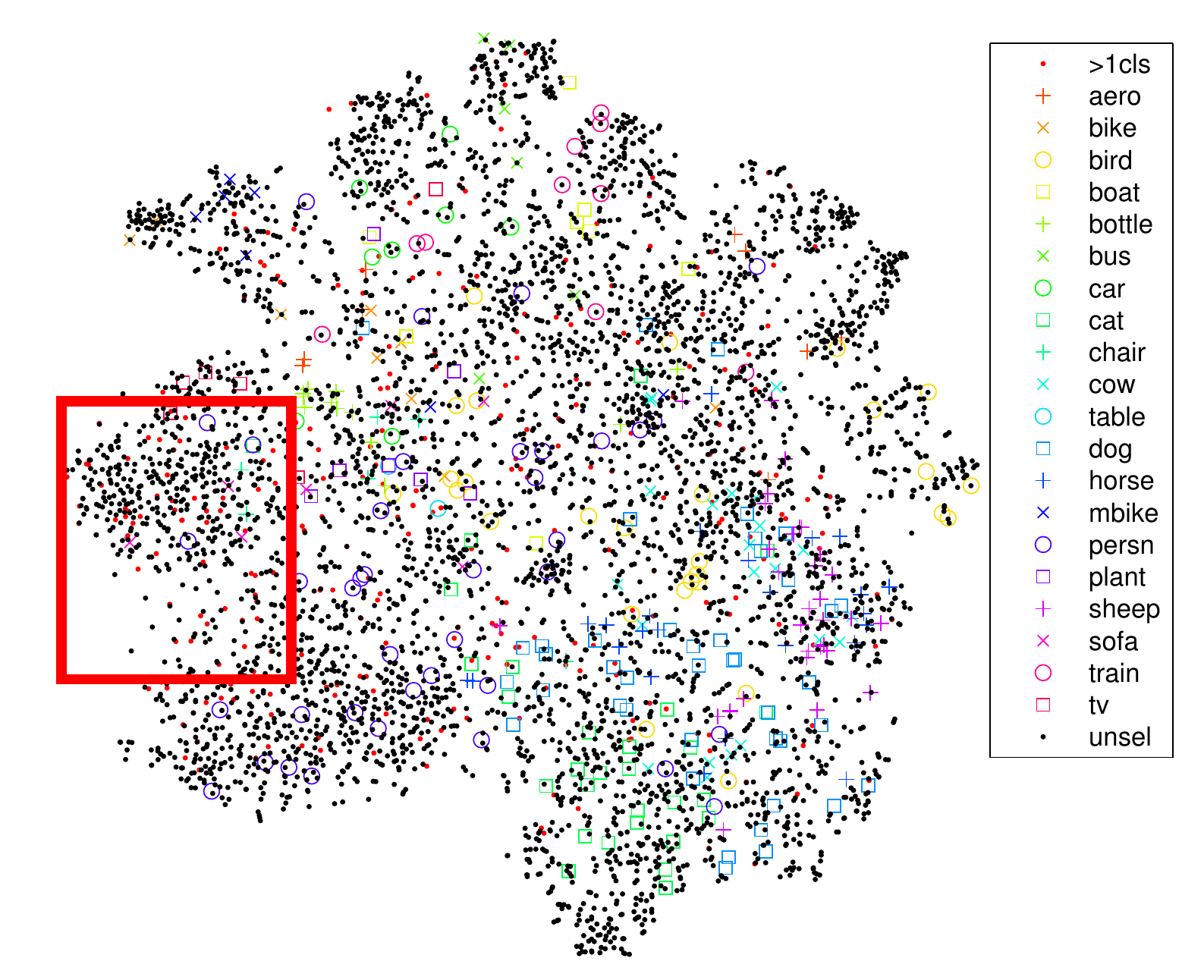}
        \caption{Classification (C)}
        \label{fig:selection_C}
    \end{subfigure}
    \\
    \begin{subfigure}[t]{0.49\textwidth} 
        \centering
        \includegraphics[width=\linewidth]{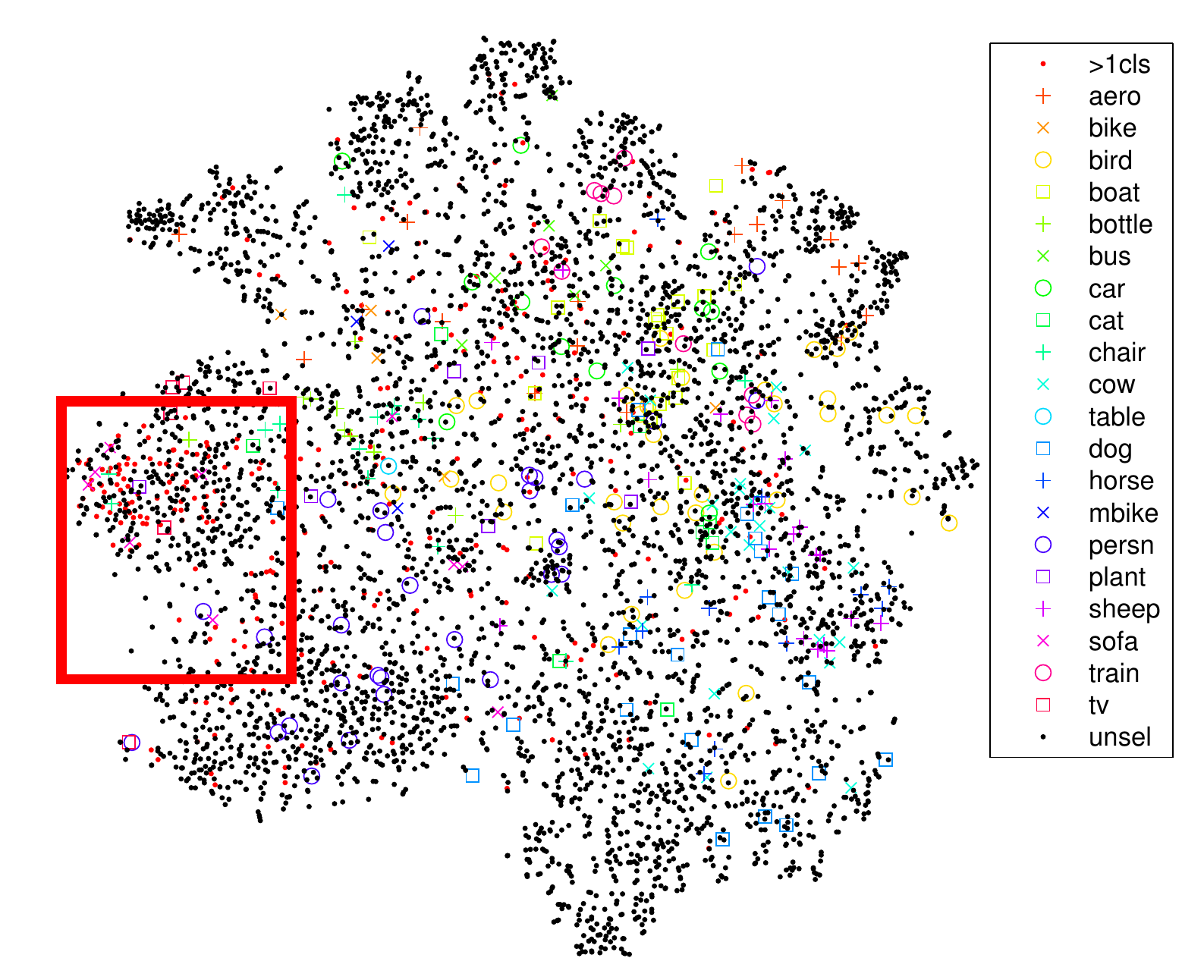}
        \caption{Localization stability + classification (LS+C)}
        \label{fig:selection_LS+C}
    \end{subfigure}    
    ~
    \begin{subfigure}[t]{0.49\textwidth} 
        \centering
        \includegraphics[width=\linewidth]{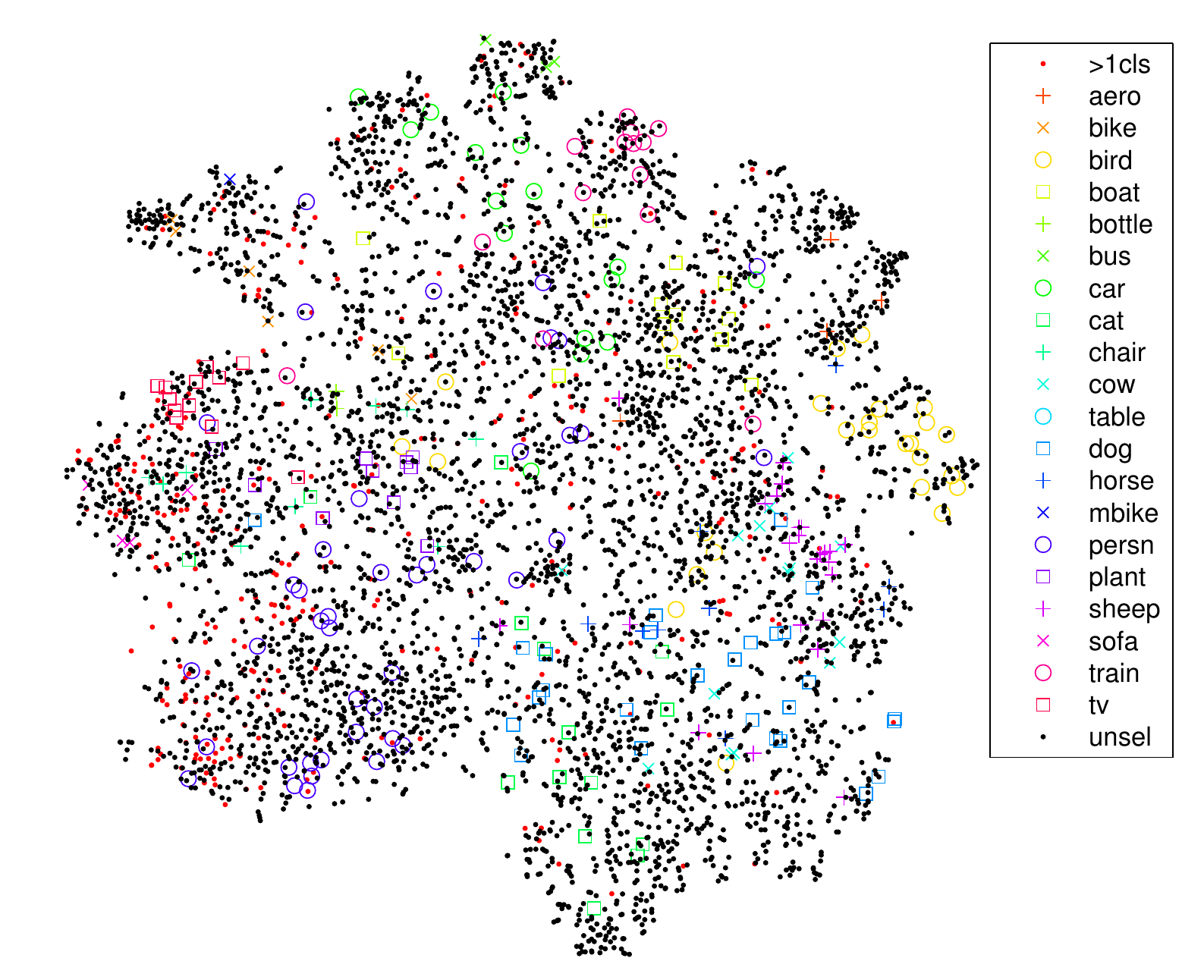}
        \caption{Localization tightness + classification (LT/C)}
        \label{fig:selection_LT+C}
    \end{subfigure}
    \caption{The visualization of selection results by different active learning methods. Different from Fig.~\ref{fig:selection_all_methods_binary}, each colored marker not only represents a selected image, but also indicates the class that objects contained in the image belong to.}
\label{fig:selection_all_methods}
\end{figure*}

We have two major observations from the visualzation results on the PASCAL 2012 dataset.
First, the random sampling (R) method selects images for annotation across all categories, no matter it is a difficult class or an easy class.
Compared to the other methods, lots of images of cats and cars are selected by R (blue rectangles in Fig.~\ref{fig:selection_R_binary} and Fig.~\ref{fig:selection_R}).
However, these classes are relatively easy so the room for improvements is not that large.
Also, the selected images are not informative so that even many images are selected in these classes, there is no large improvement over the other methods.

\begin{table*}[htbp]
\center
\begin{scriptsize}
\tabcolsep=0.11cm\begin{tabular*}{0.93\linewidth}{l|*{20}{c}|c}
method &aero&bike&bird&boat*&bottle*&bus&car&cat&chair*&cow&table*&dog&horse&mbike&persn&plant*&sheep&sofa&train&tv&mAP\\
\hline
R&68.3 &61.5 &\underline{54.2} &27.8 &30.4 &\underline{68.2} &\underline{58.2} &76.3 &28.4 &44.8 &31.1 &\underline{73.7} &64.1 &\underline{67.9} &66.7 &21.9 &52.4 &\underline{41.7} &\textbf{64.8} &\underline{55.5} &52.9 \\
C&\underline{72.8} &\underline{66.6} &50.8 &28.5 &34.8 &64.3 &54.3 &\underline{77.5} &27.2 &\textbf{53.2} &36.3 &\textbf{79.0} &\textbf{70.4} &66.5 &\underline{69.0} &21.9 &\underline{59.6} &38.8 &60.6 &54.5 &54.3 \\
LS+C&68.1 &\textbf{68.0} &52.0 &\textbf{34.2} &\underline{34.9} &\textbf{70.0} &\textbf{59.9} &74.4 &\underline{30.3} &44.2 &\textbf{42.1} &73.6 &63.3 &\textbf{69.7} &\textbf{71.7} &\textbf{28.5} &\textbf{60.2} &40.6 &\underline{64.4} &\textbf{59.0} &\underline{55.5} \\
LT/C&\textbf{74.8} &64.8 &\textbf{60.1} &\underline{28.7} &\textbf{36.4} &63.9 &58.1 &\textbf{79.7} &\textbf{31.0} &\underline{51.1} &\underline{38.1} &72.9 &\underline{66.0} &66.9 &67.2 &\underline{23.7} &56.4 &\textbf{50.4} &64.3 &54.6 &\textbf{55.5} \\
\end{tabular*}
\end{scriptsize}
\caption{Average precision for each method on the PASCAL 2012 validation set after 3 rounds of active learning (the number of labeled images in the training set is 1,100). Each number shown in the table is the result of one trial (different from Table 1 in the main paper which shows the average over 5 trials) and displayed in percentage. Numbers in bold are the best results per column, and underlined numbers are the second best results. Catergories with AP lower than 40\% in passive learning (R) are defined as difficult categories and marked by an asterisk.}
\label{tab:PASCAL2012_1100imgs_exp62}
\end{table*}

Second, as mentioned in Sec.~4.1 in the main paper, the proposed method LS+C outperforms the baseline method C especially in the difficult categories.
There is a 10$\times$ difference between difficult and non-difficult categories in the improvement of LS+C over C as shown in Fig.~6a in the main paper.
These 5 difficult categories are: boat, bottle, chair, table, and plant.
Fig.~\ref{fig:full_training_set_classwise} shows that all difficult categories but boat locate at the left part of the 2D plane.
These categories also are the ones show in scenes of a living room (Fig.~\ref{fig:full_training_set_with_imgs}), as mentioned in the previous section. 
By visual inspection, the red rectangles in Fig.~\ref{fig:selection_LS+C_binary} and Fig.~\ref{fig:selection_C_binary} show that the proposed LS+C tends to select more images for annotation in these difficult classes than the baseline method C.
Quantitative results are shown in Fig.~\ref{fig:selection_bar_chart_exp62}. 
The proposed LS+C selects images that contain objects belong to difficult classes much more than the baseline method C.
By selecting more images for annoation, the proposed LS+C gets more improvement in these difficult classes.
In contrast, for easy classes (catergories with AP higher than 70\% in passive learning) like cat and dog, the baseline method C selects more images than the proposed LS+C as shown in Fig.~\ref{fig:selection_bar_chart_exp62_easy}.
These observations indicate that C focuses on non-difficult categories to get an overall improvement in mAP, but does not perform well in difficult categories.

\end{document}